\documentclass[preprint]{elsart}
\usepackage{ifpdf}
\newif\ifpdf\ifx\pdfoutput\undefined\pdffalse\else\pdfoutput=1\pdftrue\fi
\ifpdf\fi

\usepackage{graphicx,amssymb}

\usepackage{times}
\usepackage{url}
\usepackage{epsfig}
\usepackage{graphicx}
\usepackage{amssymb}
\usepackage[cmex10]{amsmath}
\usepackage{subfigure}
\usepackage{cite}
\usepackage{comment}
\usepackage{textcomp}
\usepackage{color}
\usepackage{multicol}
\usepackage{multirow}

\DeclareMathOperator*{\argmax}{arg\,max}

\def\TReg{\textsuperscript{\textregistered}}

\def\TTra{\textsuperscript{\texttrademark}}

\excludecomment{simplepseudocode}
\excludecomment{fullpseudocode}

\graphicspath{{./}}
\DeclareGraphicsExtensions{.pdf}

\begin{document}

\begin{frontmatter}

\title{Technical Report \\Automatic Contour Extraction \\from $2D$ Neuron Images}

\author[imeusp]{J. J. G. Leandro\corauthref{cor}}\ead{jleandro@vision.ime.usp.br}, 
\author[imeusp]{R. M. Cesar-Jr}\ead{cesar@vision.ime.usp.br}, \author[ifscusp]{L. da F. Costa}\ead{luciano@ifsc.usp.br}

\corauth[cor]{Corresponding author.}

\address[imeusp]{Institute of Mathematics and Statistics - USP, Department of Computer Science\\
 Rua do Mat\~{a}o, 1010 -  S\~{a}o Paulo - SP, 05508-900 Brazil} 
\address[ifscusp]{Instituto de F\'{\i}sica de S\~{a}o Carlos - USP, 
 Department of Physics and Informatics\\ Av. Trabalhador S\~{a}ocarlense, 400 - S\~{a}o Carlos - SP, 13560-970 Brazil}

\begin{abstract} 
  This work describes a novel methodology for automatic contour
  extraction from 2D images of 3D neurons (e.g. camera lucida images
  and other types of 2D microscopy).
  Most contour-based shape analysis methods can not be used to
  characterize such cells because of overlaps between neuronal
  processes. The proposed framework is specifically aimed at the
  problem of contour following even in presence of multiple overlaps.
  First, the input image is preprocessed in order to obtain an
  $8$-connected skeleton with one-pixel-wide branches, as well as a
  set of \emph{critical regions} (i.e., bifurcations and crossings).  
  Next, for each subtree, the tracking
  stage iteratively labels all valid pixel of \emph{branches}, up
  to a \emph{critical region}, where it determines the
  suitable direction to proceed. Finally, the labeled \emph{skeleton
  segments} are followed in order to yield the \emph{parametric contour} 
  of the neuronal shape under analysis.  
  The reported system was successfully tested with respect to several images
  and the results from a set of three neuron images are presented here,
  each pertaining to a different class, i.e. alpha, delta and
  epsilon ganglion cells, containing a total of $34$ crossings. The
  algorithms successfully got across all these overlaps. The
  method has also been found to exhibit robustness even for images with
  close parallel segments. The proposed method is robust and may be
  implemented in an efficient manner. The introduction of this
  approach should pave the way for more systematic application of 
  contour-based shape analysis methods in neuronal morphology.
\end{abstract}


\end{frontmatter}

\section{Introduction}

\begin{figure*}[htb]
\centering
 \centerline{\subfigure[]{\includegraphics[width=80mm]{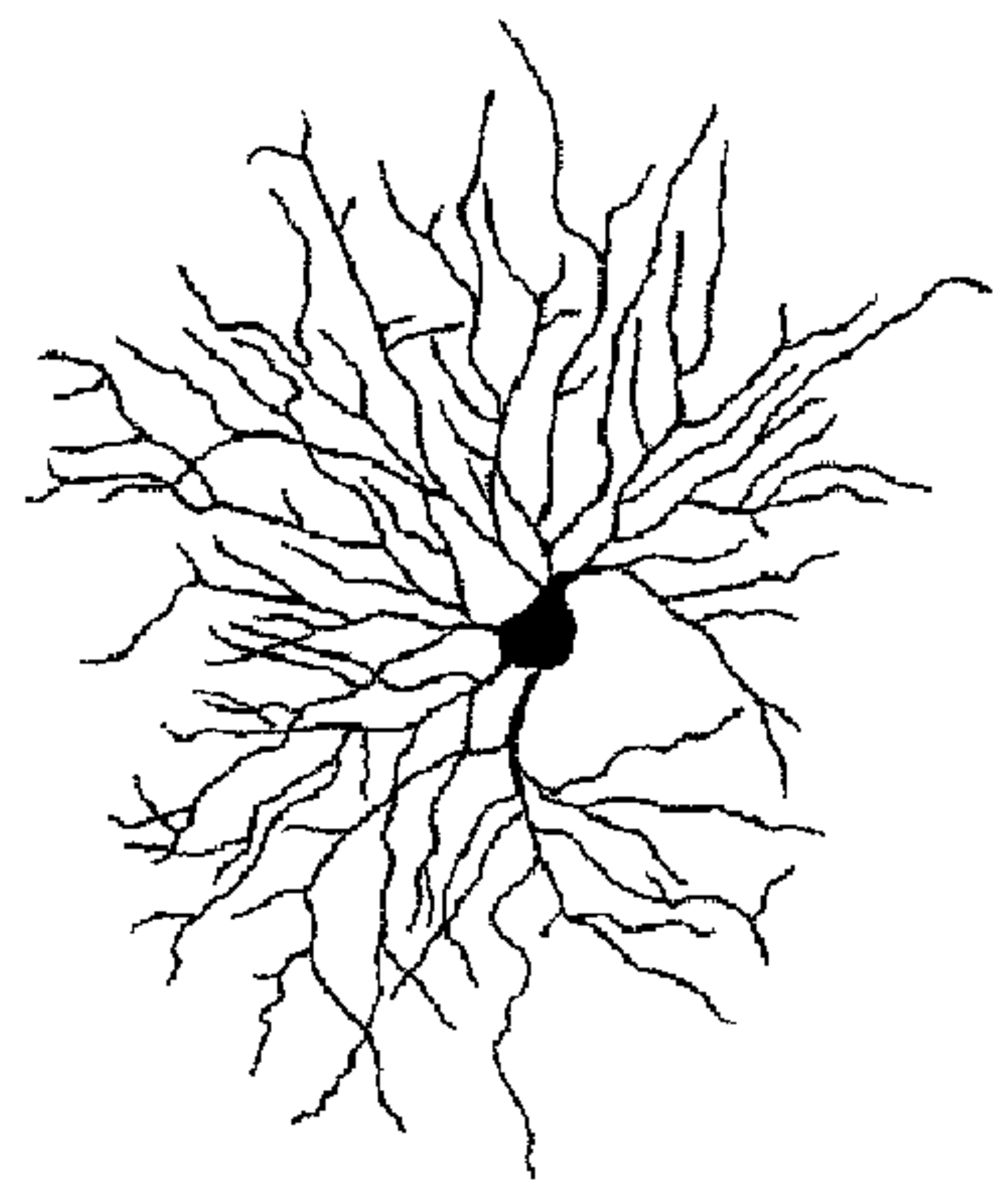}
\label{fig:neuron}}
\hfil
\subfigure[]{\includegraphics[width=80mm]{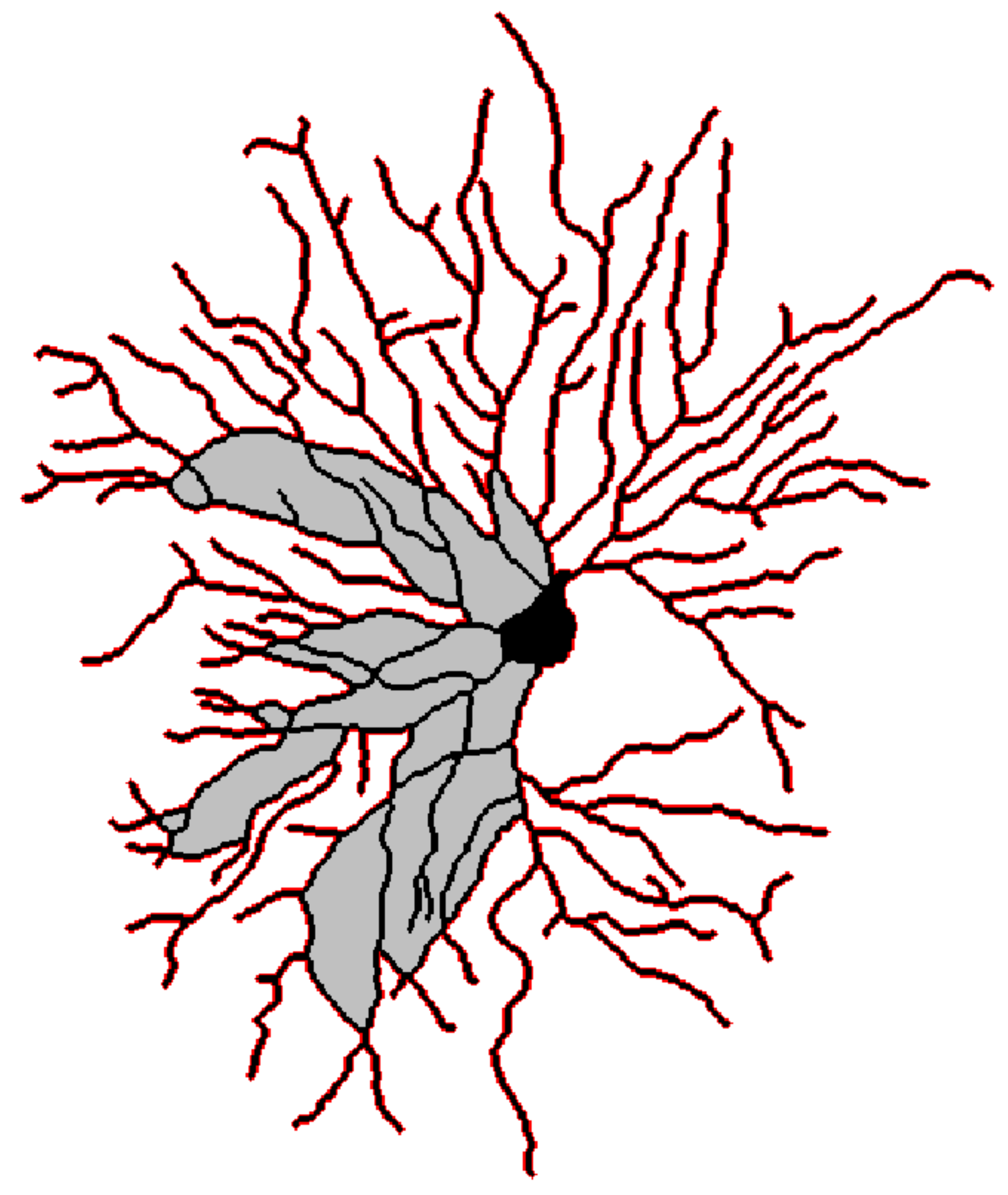}
\label{fig:unreachable}}}
  \caption{(a) Example of $2D$ neuron image considered in this work.
  (b) Neuron image (black) and respective contour (red) as provided by
  a traditional contour following algorithm. The light gray shaded
  areas represent the innermost regions that remain inaccessible for
  such algorithms.}
\label{fig:problems-with-traditional}
\end{figure*}

Neurons can be understood as cells specialized in interconnections,
which are implemented through synapses extending from axonal to dendritic
arborizations.  Though the connectivity of mature neuronal systems may
seem to be stable, it is actually subjected to continuing
re-organizations influenced by stimuli presentation and biological
changes.  The number of connections which a neuron may receive is to a
large extent defined by the shape of its dendritic tree, which serves
as a target for growing axons.  As a consequence, the full
understanding of the functionality of neuronal circuits requires the
proper characterization of the neuronal morphology
(e.g.~\cite{surv_condmat, costa2002,rocchi2007}).  Among the several
approaches based on the characterization of the geometry and
connectivity of neuronal cells~\cite{caserta90, costa2002,
jelinekfernandez98, morigiwa89, rocchi2007}, a particularly important 
and broad set of shape analysis algorithms relies on a parametric 
representation of the neuronal shape~\cite{costabook01}, i.e. in the 
form $c(t)=(x(t),y(t))$.\footnote{For instance, a circle may be represented as
$c(t)=(x(t),y(t))=(cos(t),sin(t))$}. The proper contour extraction of
$2D$ neuron images yields parametric signals, from which features
can be calculated and used to characterize differences between
neuronal shapes. Despite the availability of algorithms to extract 
parametric contours from digital images, they can not be applied directly 
in neuroscience because of the intense overlap (crossings) which is 
frequently observed among the neuronal processes. In order to better 
appreciate this limitation, please refer to 
Fig.~\ref{fig:problems-with-traditional}.

Often, the 3D neurons are projected into the 2D space (e.g. camera
lucida and several types of 2D microscopy), so that the contour of the
cells can, in principle, be represented as 1D parametric
curves~\cite{herman1998} (Fig. \ref{fig:neuron}). Important
information, such as the normal and/or tangent orientation fields
along such contours, as well as the arc length of each segment,
can then be obtained, allowing the estimation of
important geometrical properties such as the contour curvature and 
wavelets (e.g.~\cite{cesarcost:1998,costa2002}), which are known to 
provide particularly valuable information about the shape of the neuron,
including its bending and concavity.  Another possible application of
contours in neuroscience is as a means to automatically obtain
neuronal dendrograms~\cite{cesar99}. However, such an approach is
often complicated by the presence of crossings between the neuronal
processes in the $2D$ image, implying some regions of the cell to
become inaccessible for traditional contour extraction algorithms (see
Fig.\ref{fig:unreachable}).  Henceforth, we refer to such
inaccessible contour portions as the \emph{innermost regions} of the
shape. In brief, most contour following algorithms work as
follows. Firstly, the algorithm detects an initial contour pixel.
Then, the algorithm searches the next contour pixel, by probing the
current contour pixel vicinity. The algorithm travels around the whole
object, until revisiting the first pixel, once the task 
has been completed. Further details can be found in the full description of
such methods in \cite{costabook01}.

In contour following algorithms, it is usually impossible to traverse
the innermost regions delimited by crossings (due to the 3D to 2D
projection).  Consequently, only the outer contour of the cell is
represented, thus missing the innermost structures. This fact is
illustrated in Fig. \ref{fig:unreachable}, where the light gray shaded
innermost regions represent areas inaccessible to traditional contour
following algorithms, thus yielding just the red curve as the
respective contour.

It should be observed that, in principle, 3D imaging is naturally
better than 2D imaging because it contains more information about the
acquired structure.  However, 3D imaging intrinsically demands
additional computational resources, especially for the enhancement and
interpretation of the structures.  In addition, in cases where the
original neurons are mostly planar, such as the ganglion cells in the
retina, the 3D capture often contributes little to the separation
between thin crossing segments, which are mostly contained within a
narrow range of z-coordinates.  Another reason why 2D imaging remains
valuable in neuroscience concerns the fact that not all types of
microscopy can be performed in 3D.  Finally, many 2D neuron imaging
systems and images are already available (e.g. camera lucida images
from previous experiments and published literature), so that
research using such images would also benefit from the introduction
of the methodology proposed in the present paper.

Despite the importance of the problem of contour extraction from $2D$
neuron images, we were not able to find any related approaches in the
literature. To the best of our knowledge, the only similar work in the
literature was described in \cite{Rothaus:2007,Rothaus:2008}, which
presents a semi-automatic method to separate veins and arteries in the
vascular trees of fundus images. In order to solve such a problem, it
is necessary to label the vessels, which are also branching
structures. Firstly, a skeleton image is obtained from the vessel
trees.  The skeleton is then represented as a vascular graph named
$G$, which comprises all the information regarding the connectivity
among critical regions and branches segments. Subsequently, that
method propagates the labels manually assigned to other vessels
throughout the vascular graph $G$.  In this case, it is well-known
that crossings only take place between a vein and an artery, thus such
a priori knowledge on the vessels structure is used to simplify the
approach, determining that opposite branches segments in a crossing
should be necessarily equally labeled.  Such assumptions make this
approach specific to the vein/artery tracing problem.  Alternatively,
neighbouring branches segments in a crossing should be labeled with
distinct tags.  Although there are some similarities between the
approaches reported in~\cite{Rothaus:2007,Rothaus:2008} and in the
current article (mainly concerning the use of skeletons and case
analysis of bifurcations and crossings), the latter methodology adopts
a different and potentially more general approach. Since the
aforementioned system has been specially designed for vascular trees,
the number of labels available to be assigned is bounded to two: veins
and arteries. Conversely, our algorithms yield assignments of a
distinct label for each existing branch and also a distinct label for
each dendritic tree, regardless of the number of existing dendritic
trees in the neuron. Also, by segmenting each branch within the image,
our method allows the branches to be counted as well as the lengths of
the segments to be measured. In contrast 
to the Rothaus' system, ours solely carries out local assessment of the image 
topology in a sequential-like fashion, without graph representations, 
thus avoiding the backtracking step. Furthermore, our system provides the
parametric contour of the whole structure, being thus possible to
extract several geometrical features to be fed to a classifier.

Also, the results reported in our work can also be useful for the 
unsolved $3D$ cases by confocal microscopy. In addition, there are more important aspects 
regarding the importance and applicability of our contribution, and these are as follows. 
First, there are dozens of other microscopic techniques 
which cannot yield $3D$, but only $2D$ images, necessarily implying tangling of 
neuronal branches which can be treated by our method. Such microscopy techniques are 
often required instead of confocal microscopy because they can reveal specific properties 
of the analyzed tissues and structures which cannot be imaged by confocal methodology. 

Moreover, as already mentioned, the proposed methodology is quite general and may 
be applied to other branching structures.

The rationale of the present work is to deal with neuronal overlaps by
incorporating several criteria such as the use of similarities along
the tangent orientation as a means to identify the proper continuation
of the neuronal processes at crossing-points.  The proposed
methodology is composed by the subsequent application of the following
three algorithms:

\begin{enumerate}
\item Preprocessing
\item \emph{Branches Tracking Algorithm (BTA)}
\item \emph{Branching Structures Contour Extraction Algorithm (BSCEA)}
\end{enumerate}

In short, the \emph{BTA} is an algorithm aimed at the
segmentation of each distinct branch within a $2D$ neuron image other
than the soma and intercepting regions. The
\emph{BSCEA} is an algorithm intended to the extraction of the
parametric contour from a $2D$ neuron image, based on the \emph{BTA}.

For clarity's sake, this paper is presented in increasing levels of detail, hence 
developing as follows. Section~\ref{sec-preambule} contains an
overview of the proposed framework, which is further detailed in
Section~\ref{sec-general}. Experimental results
considering real neuronal cells are presented in
Section~\ref{sec-results}. The paper concludes in Section~\ref{sec-concluding},
by identifying the main contributions, as well as possibilities for future works. 
Low level descriptions has been left to the Appendices~\ref{sec-bta-cr-class-rules} 
and~\ref{sec-bta-bfs-examples}.

\section{Preambule}
\label{sec-preambule}

\begin{table}
\renewcommand{\arraystretch}{1.3}
\caption{Summary of concetps.}
  
\label{tab:table20}
\centering
\begin{tabular*}{1\textwidth}{@{\extracolsep{\fill}}ll|l}
\hline
\multicolumn{2}{c|}{TERM} & \multicolumn{1}{c}{DESCRIPTION}\\
\hline\hline
\multicolumn{3}{c}{Points} \\
\hline
\multirow{2}{*}{seed} & \multicolumn{1}{|l|}{primary} & origin of a segment stemming from the soma\\
\cline{2-3}
& \multicolumn{1}{|l|}{secondary} & origin of a segment stemming from a critical region\\
\hline
\multicolumn{2}{c|}{termination}   & end point of a branch\\
\hline\hline
\multicolumn{3}{c}{Lines}\\
\hline
\multicolumn{2}{l|}{segment} & line of pixels delimited by other structures \\ 
\hline
\multicolumn{2}{l|}{inward segment}  & incoming segment at a critical region\\ 
\hline
\multicolumn{2}{l|}{outward segment}  & outgoing segment from a critical region\\
\hline
\multicolumn{2}{l|}{branch}  & string of segments\\ 

\hline\hline
\multicolumn{3}{c}{Critical Regions (\emph{CR})}\\
\hline
\multicolumn{2}{l|}{\multirow{2}{*}{bifurcation}} & cluster of pixels where an inward segment splits into\\
												  & &  two; one of which in other direction\\
\hline
\multicolumn{2}{l|}{\multirow{2}{*}{crossing}} & cluster of pixels where an inward segment splits into\\
											   & &  three; two of which in other direction\\
\hline
\multicolumn{2}{l|}{\multirow{2}{*}{superposition}} & cluster of bifurcations where an inward segment splits\\
                                               & &  into three; two of which in other direction\\
\hline\hline
\multicolumn{3}{c}{Collections}\\
\hline
\multicolumn{2}{l|}{dendritic arbour} & collection of branches growing out of soma\\
\hline
\multicolumn{2}{l|}{periphery} & collection of dendritic arbours (excluding soma)\\
\hline
\multicolumn{2}{l|}{skeleton} & one-pixel wide skeleton from the periphery\\
\hline
\end{tabular*}
\end{table}

Usually, an optical acquisition device yields an image as output,
corresponding to a summary and incomplete representation of the
information originally present in the original object~\cite{castleman79}. 
As a result, images are normally devoid
of some information, such as related to depth, a problem arising from
the supression of the third dimension in the $3D$ original object as
implied by its object projection onto the $2D$ plane.  In the context
of complex shape images, like neurons, depth information is of
extreme importance to properly discern the structures in the
image. The current work approaches this problem, more especifically
the extraction of contours of neuronal cells imaged onto $2D$ frames. In particular, 
the $2D$ neuron images used herein have been obtained through a camera lucida device.

\begin{figure*}[ht]
\centering
  \centerline{\includegraphics[width=135mm]{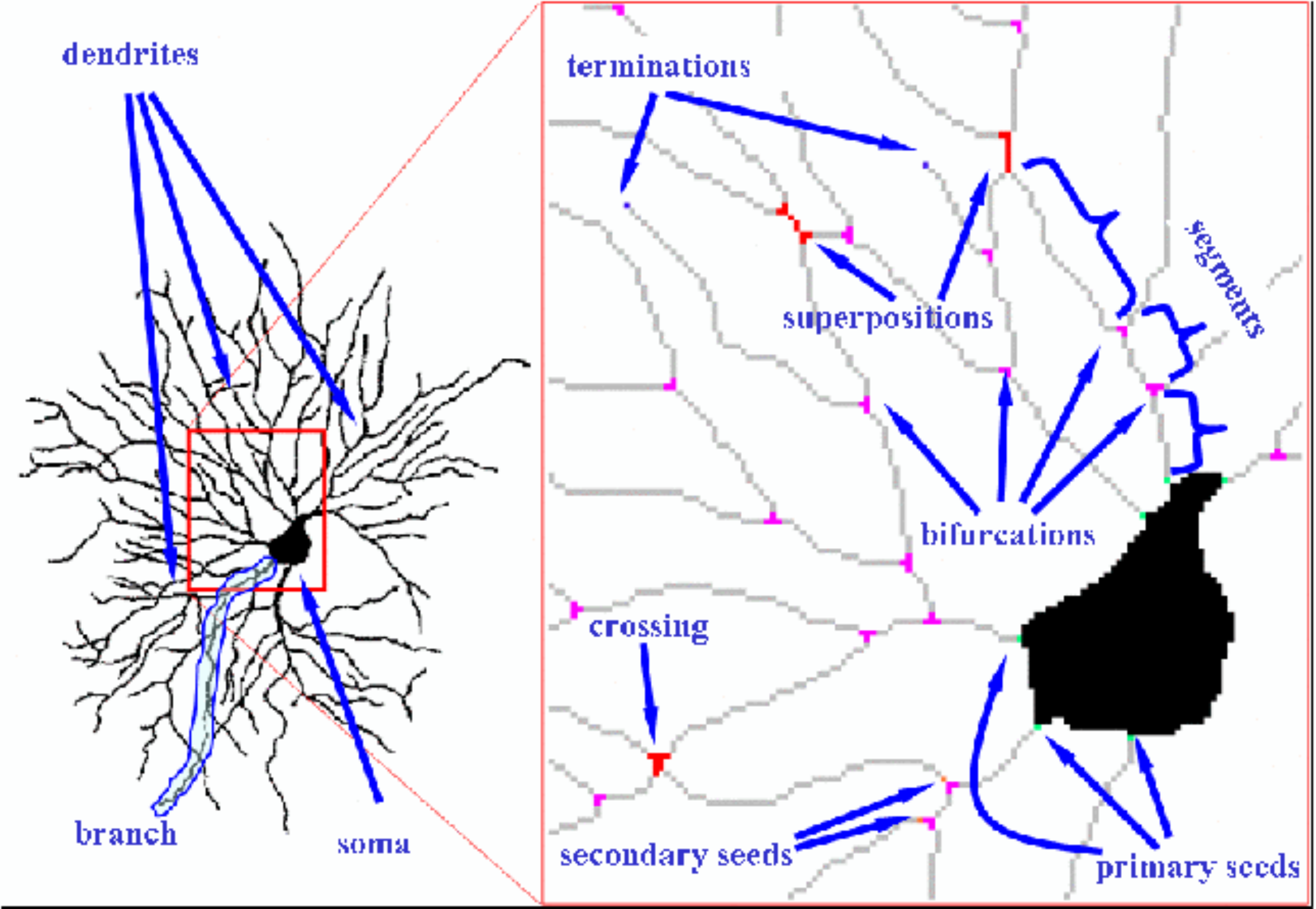}}
\caption{Extended terminology adopted in this work:
dendrites, soma, branches, segments, seeds, terminations and critical regions.}
\label{fig:terminology}
\end{figure*}

\subsection{Terminology}
\label{sec-terminology}

Initially, our approach considered the existence of only two types of
structures among branches, namely bifurcations and crossings. However
the number of adjacent segments at each critical region proved not to
be enough to properly classify them, leading to misclassifications.
Only through the incorporation of additional information, namely the
identification of several geometrical features along the neuronal
shape, it has been possible to achieve correct classification of the
critical regions.  The neuronal shape can be described as the union of
its constituent parts, i.e. soma, dendrites or dendritic arbours and
axon.  In order to elaborate on the explanation of our methods, we
extend such a terminology.  Notice that it does not refer to
additional functional parts in neurons, but rather to morphological
building blocks which compose those functional parts in their
respective $2D$ images.  As a matter of fact, some structures such as
crossings and superpositions do not even occur in a real $3D$ neuron,
being just an immediate consequence of the projection from the $3D$
space onto the plane or their close proximity even in 3D
spaces. Hence, the arbours present in $2D$ skeletons obtained from
neuron images are subdivided into their morphological constituent
parts as follows (see Fig.~\ref{fig:terminology} and
Tab.~\ref{tab:table20}):

\begin{itemize}
\item \emph{Points}
\item \emph{Lines}
\item \emph{Critical Regions}
\item \emph{Collections}
\end{itemize}

The aforementioned categories of structures encompass the typical
structures which usually appear in $2D$ neuron images.  Such
structures must be distinguished so as to provide additional
information about the original neuronal shape, therefore allowing more
general and effective performance.

The category \emph{Points} comprises three classes of extremity
points: primary seeds, secondary seeds and terminations. Each
extremity point is classified regarding its location, i.e. a
\emph{primary seed} corresponds to a junction point between a dendritic
tree and the soma, while a \emph{secondary seed} refers to a junction
point between a critical region and a dendritic subtree. Basically,
the difference between a primary seed and a secondary seed is that
a primary seed is necessarily adjacent to the soma, while a secondary
seed is not. \emph{Terminations} are end points of branches.  The
reason for distinguishing between points is that the tracking starts
from the primary seeds and finishes at terminations, occasionally
repeating itself in a recursive-like fashion from secondary seeds.

The category \emph{Lines} encompasses two cases: segments and
branches.  Each line is classified with respect to its extremity
points, i.e.  a segment may grow out from either a primary or a
secondary seed, but does not necessarily end at a
termination. \emph{Segments} are lines of pixels delimited by a pair
of minor structures, for instance a seed and a critical region, or two
critical regions, or a critical region and a termination. Conversely,
a branch may stem from either a primary or a secondary seed, ending
necessarily at a termination. It follows from such a definition that a
\emph{branch} is a ramification made up of a string of segments,
growing out of a seed up to a termination, as shown in
Fig.\ref{fig:terminology}. In addition, segments may be further
subclassified depending on their relationship to an adjacent critical
region. By analyzing a neuron shape from inside out, that is, from the
soma towards its terminations, an incoming adjacent segment to a
critical region is said an \emph{inward segment}, while an outgoing
adjacent segment to a critical region is said an
\emph{outward segment}.  The reason for distinguishing between lines 
is the need to recognize the constituent parts (segments) of branches
every time the tracking algorithm reaches a critical region. Adjacent
segments which present tangent similarity should be regarded as part
of the same branch. Segments and branches play different roles in the
tracking algorithm, thus deserving distinct names.

\begin{figure*}[htbp]
\centering
 \centerline{\subfigure[]{\includegraphics[width=40mm]{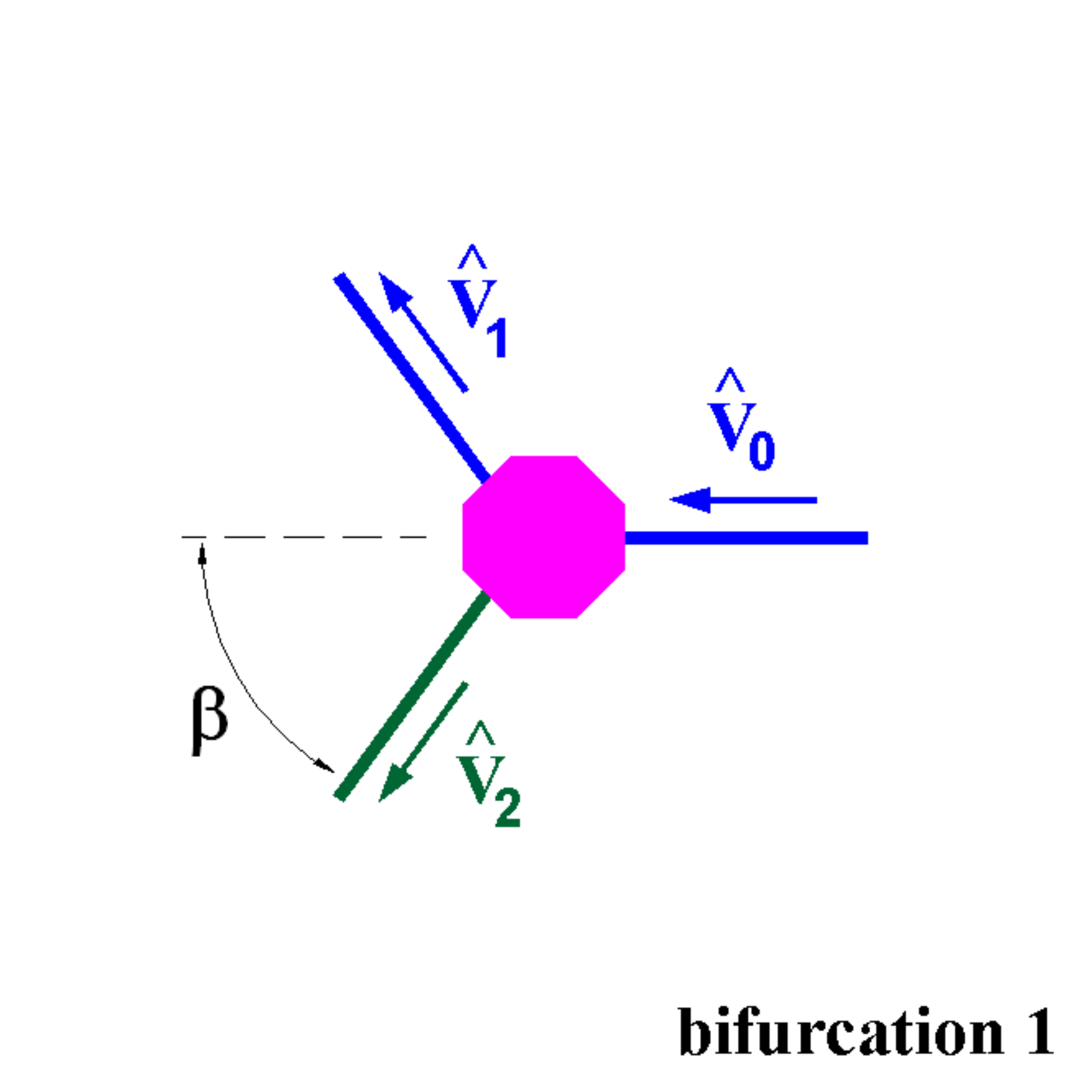}
\label{fig:crc_a}}
\hfil
\subfigure[]{\includegraphics[width=40mm]{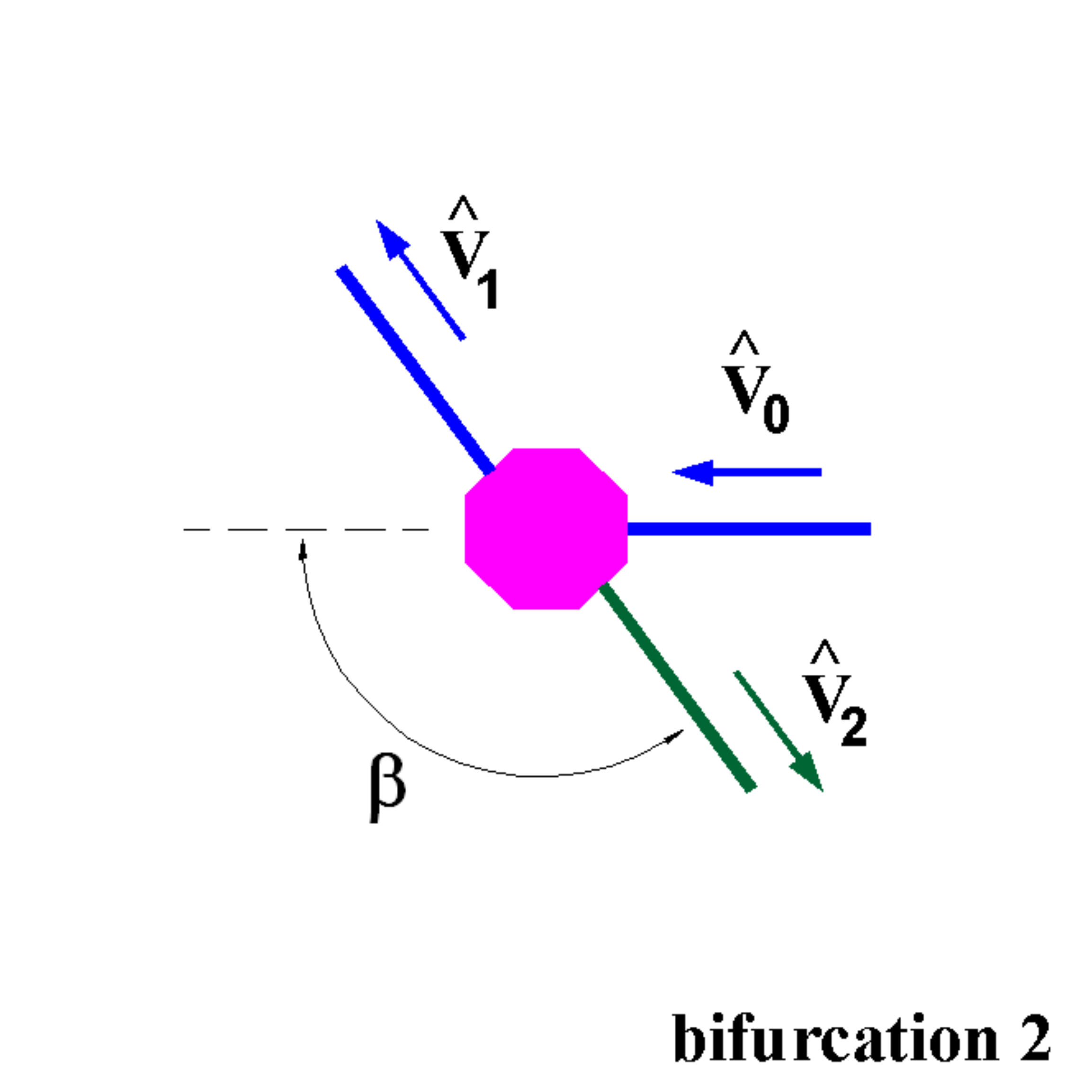}
\label{fig:crc_b}}
\hfil
\subfigure[]{\includegraphics[width=40mm]{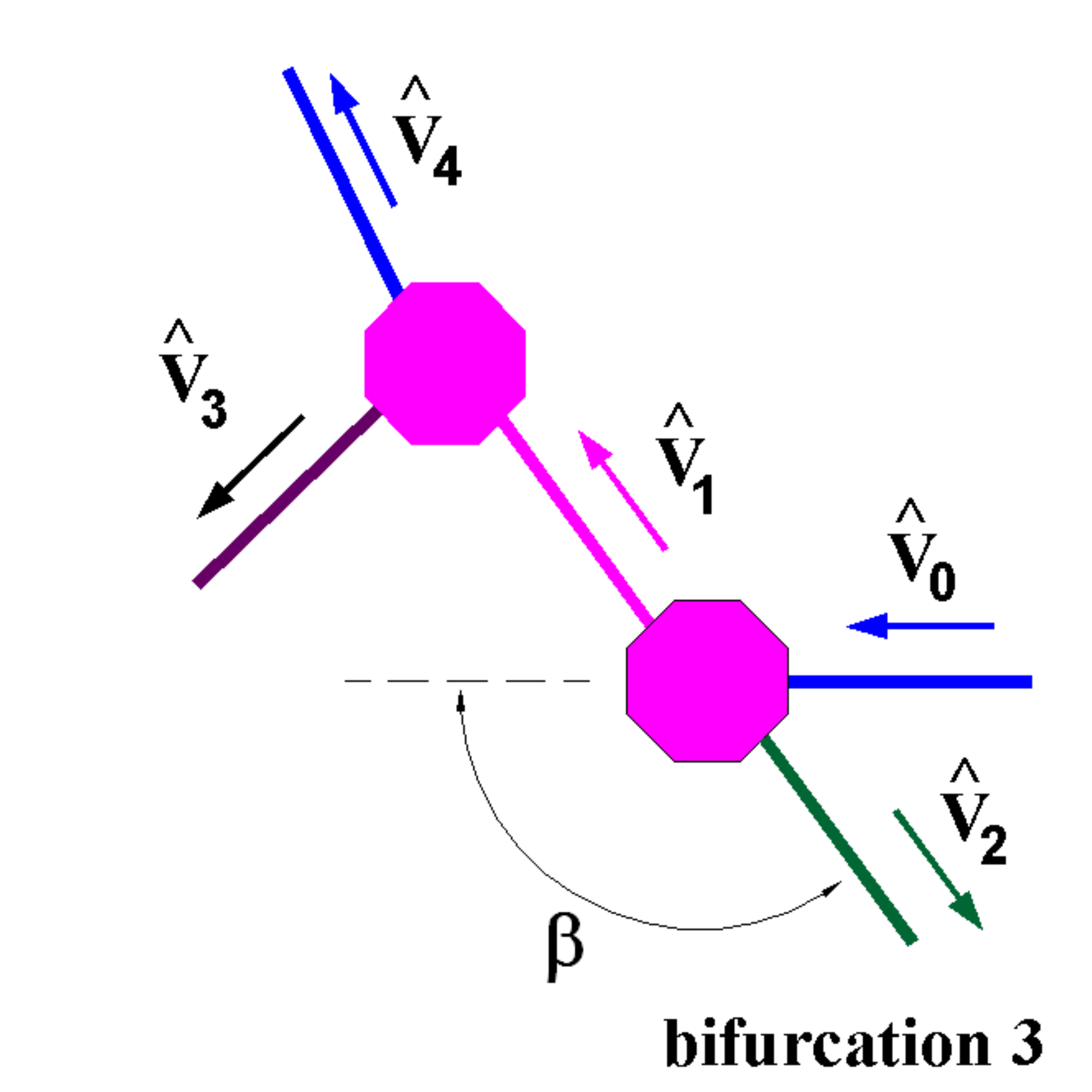}
\label{fig:crc_c}}}

\centerline{\subfigure[]{\includegraphics[width=40mm]{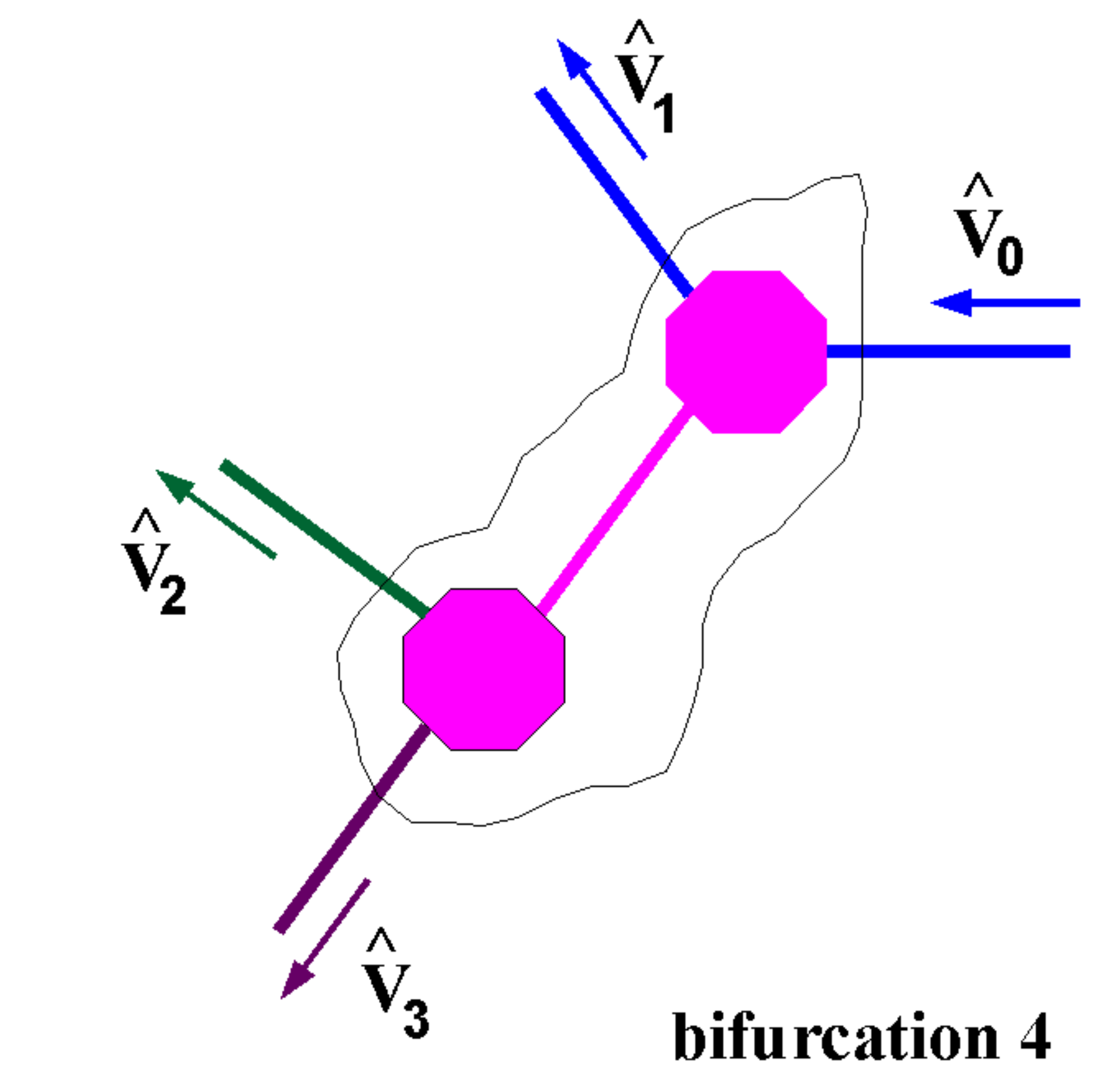}
\label{fig:crc_d}}
\hfil
\subfigure[]{\includegraphics[width=40mm]{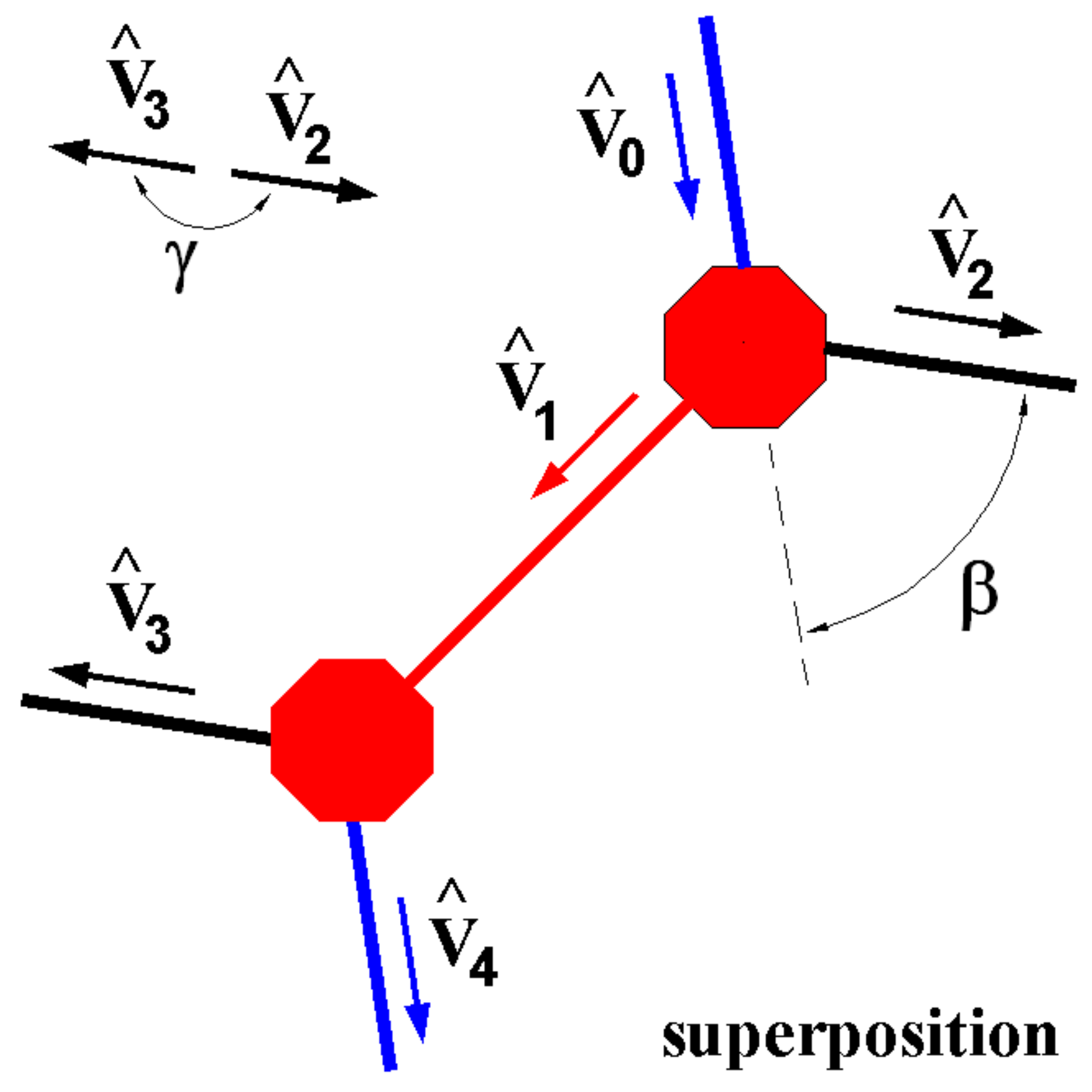}
\label{fig:crc_e}}
\hfil
\subfigure[]{\includegraphics[width=40mm]{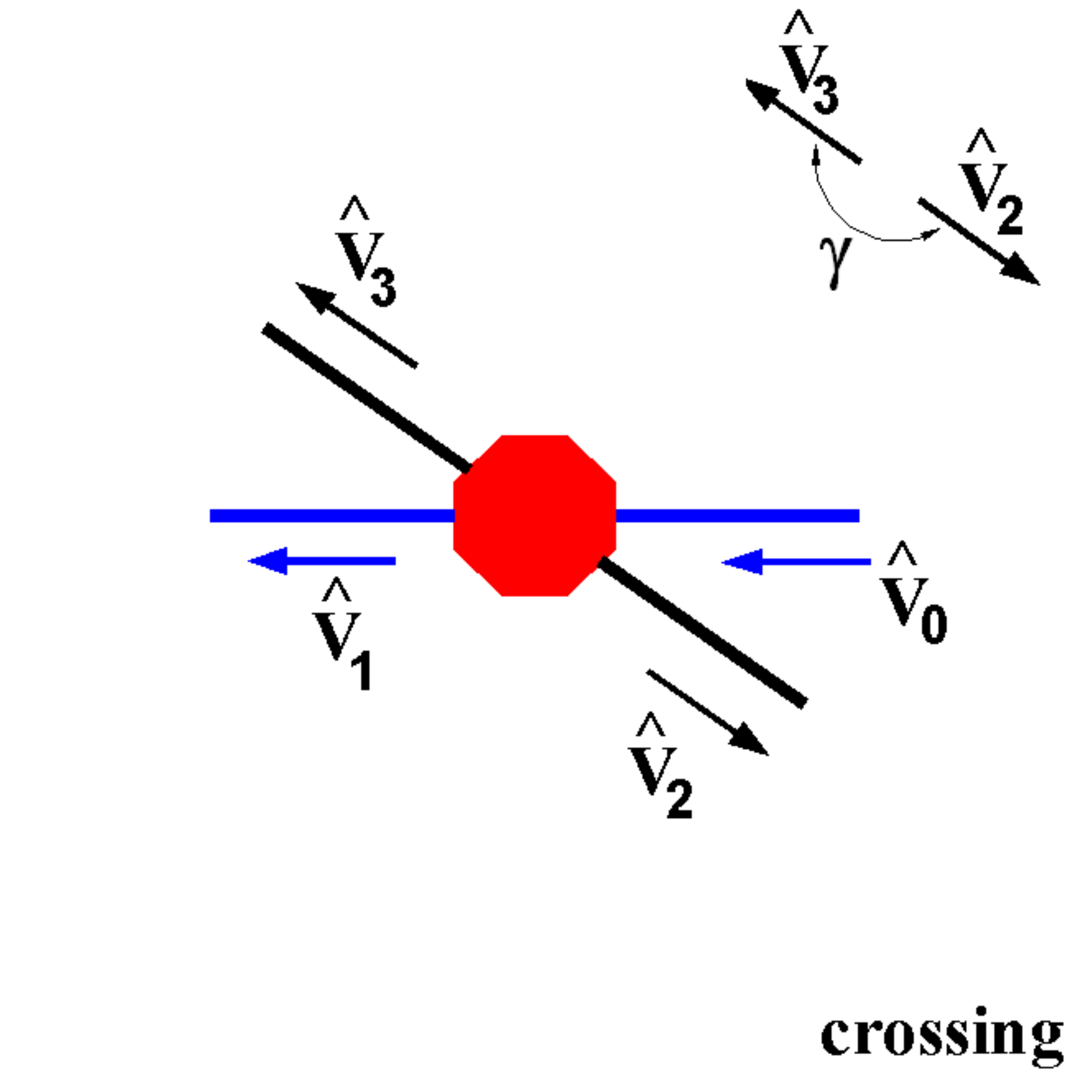}
\label{fig:crc_f}}}
     \caption{The critical regions classification rules take into
     account the angle $\beta$ between the inwards direction vector -
     $\hat{v}_0$ - and the outwards direction vector - $\hat{v}_2$,
     the angle $\gamma$ between any pair of outwards direction vectors
     and the cardinality $|E_1|$ of the set of outwards direction
     vectors $E_1$ related to the current critical region $s_1$.  (a)
     If a critical region presents $|E_1| = 2$ and $\beta <
     90^{\circ}$ it is immediately classified as a bifurcation $1$;
     (b) if $\beta \geq 90^{\circ}$ it may be a bifurcation $2$ or (c)
     even a bifurcation $3$ as long as there is another critical
     region nearby. (d) A bifurcation $4$ presents $|E_1|=3$, but
     there is no $\gamma \approx 180^{\circ}$.  (e) A superposition
     appears with $|E_1|=2$ and another critical region $s_2$ nearby,
     with $\gamma \approx 180^{\circ}$. (f) Finaly, a crossing has
     $E_1=3$ and $\gamma \approx 180^{\circ}$.  Notice that
     bifurcation $1$ could be mistaken for either bifurcation $4$ or
     superposition and bifurcation $2$ could be mistaken for
     bifurcation $3$ in case the adopted rules had not been considered.}
     \label{fig:crclassification}
\end{figure*}

\emph{Critical Regions} are clusters of pixels where branches 
meet each other. This category includes three classes of regions:
bifurcations, crossings and superpositions.  Each Critical Region is
classified by considering its shape, the number of segments adjacent
to it and their mutual orientation relationship, as well as the
proximity relationship between the current critical region and other
regions nearby.  On a \emph{bifurcation}, an inward segment often
separates into two outward segments with different orientations, as
depicted in Fig.\ref{fig:crclassification}-(a-b).  Occasionally,
bifurcations may occur very close one another.  Viewing such regions
from a larger scale would suggest just one critical region, where an
inward segment splits into three outward segments, as shown in
Fig.\ref{fig:crclassification}-(c-d).  Similarly, on a
\emph{superposition}, an inward segment splits into three outward
segments, two of them in normally distinct and opposite orientations,
as can be seen in Fig.~\ref{fig:crclassification}-(e).  If
superpositions are not considered, they could be locally misunderstood
as two very close bifurcations attached by a tiny segment.  Finally, a
\emph{Crossing Region} is a cluster of pixels where an inward segment
splits into three outward segments, two of them in quite distinct and
necessarily opposite orientations, as shown in
Fig.\ref{fig:crclassification}-(f).

Though all critical regions share the property of being formed by
pixels with neighborhood greater than two, their shape structure are
quite different. The reason for distinguishing between critical
regions is to assure that both the tracking and the contour extraction
algorithms behave as expected whenever such structures are found. The
algorithms undergo different processings for each kind of critical
region.

\begin{figure}[!htb]
\centering
  \centerline{\includegraphics[width=8cm]{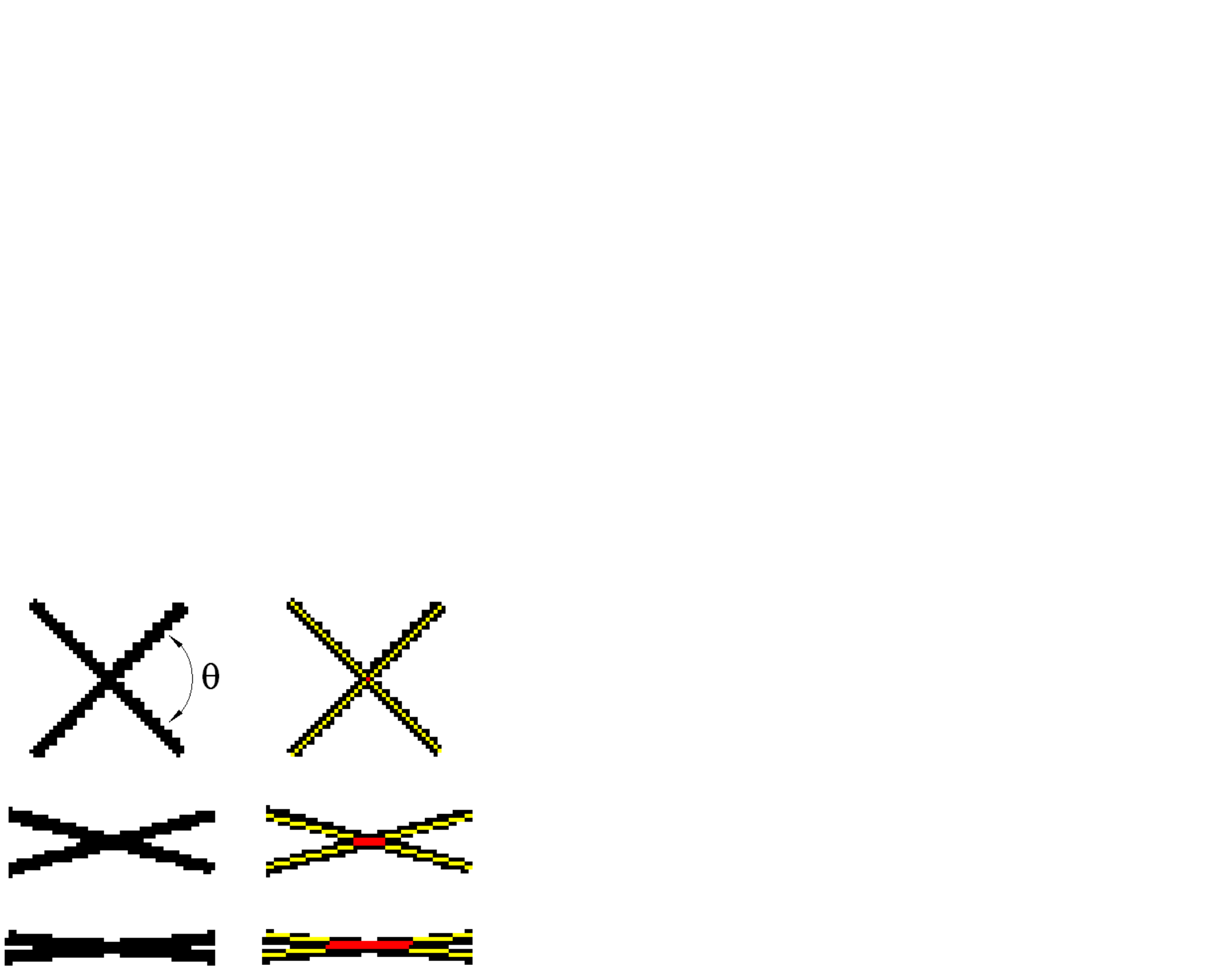}}
     \caption{The dependence between the angle $\theta$ of inclination 
     between intercepting branches (yellow) and the shape of the respective 
     critical region (red). Observe that as $\theta$ decreases, more 
     elongated the critical region becomes.}   
     \label{fig:cross-angles}
\end{figure}

At this point, it is worth emphasizing the difference between a
crossing and a superposition: although both share the property of
having an inward segment splitting into three outward segments, their
shapes are slightly different. Notice that a crossing appears as just
a cluster of pixels, while a superposition is apparently made up of
two clusters of pixels (bifurcations) attached by a short line. In
spite of the fact that both structures have been originated from
overlapping processes, the angle of inclination between these
processes plays a central role, in that the steeper the slope between
them, the greater the chance of obtaining a crossing, while the
smoother the slope between them, the greater the chance of obtaining a
superposition, as illustrated in Figure~\ref{fig:cross-angles}.

The category \emph{Collections} simply represents groups of the
aforedefined objects. A \emph{Dendritic Arbour} is a collection of
branches having roots in the soma. Hencerforth the collection of
Dendritic Arbours, that is, the neuron without the soma, is simply
referred as the \emph{Periphery}.
These concepts are summarized in the Table.~\ref{tab:table20}.

\subsection{System overview}
\label{sec-overview}

Briefly, the proposed approach for the parametric contour extraction of 
branching structures involves the following three steps:

\begin{itemize} 
\item \textbf{I}nput \textbf{P}reprocessing \textbf{A}lgorithm. In 
summary, this algorithm is aimed at preprocessing the input $2D$
image, unfolding it into additional images containing its required
structural building blocks for the subsequent steps.  The input image
is preprocessed by means of mathematical morphology
operations~\cite{morphlotufo, soille99}, yielding its separate
components, namely:

\begin{itemize}
\item periphery skeleton image, henceforth referred to as skeleton
\item critical regions image
\item terminations image
\item soma image
\item queue of primary seeds
\end{itemize}

\item \textbf{B}ranch \textbf{T}racking \textbf{A}lgorithm. The 
\textbf{BTA} has two main goals: to label each branch and to
classify each critical region. It is applied for every primary seed
present in the queue. The labelling procedure starts at the segment adjacent to the primary
seed.  After reaching a critical region, the current segment will have
been entirely labeled, so a decision concerning the next segment to
continue with the tracking must be taken.  In addition to finding the
optimal segment to move ahead, the algorithm also identifies the
current critical region as either a bifurcation, a superposition or a
crossing.  If the current critical region is a bifurcation, the
\emph{BTA} stores the related secondary seed in an auxiliary queue,
otherwise the \emph{BTA} stores the addresses of the current segment
end point and the next segment starting point.  By doing so, the
\emph{BTA} labels all the segments comprising each dendritic branch in
a recursive-like fashion, until reaching a termination.

\item \textbf{B}ranching \textbf{S}tructure \textbf{C}ontour 
\textbf{E}xtraction \textbf{A}lgorithm. 
The \textbf{BSCEA} main role is to extract the parametric contour
$c(t)=(x(t),y(t))$ along the segments comprising a $2D$ neuron image
by using the labeled branches and classified critical regions obtained
in the previous step.  Basically, the \emph{BSCEA} follows the
segments defining branching structures (resulting from the union
between the labeled skeleton and the soma) by entering all the shape
innermost regions. During the contouring process, whenever a branching
region is found, the \emph{BSCEA} contours the shape outwards, as the
traditional algorithm would. On the other hand, whenever a crossing or
a superposition is found, the \emph{BSCEA} contours the shape inwards,
by traversing the current critical region through the addresses stored
in pointers by the \emph{BTA}. Finally the \emph{BTA} gives as a
result the contour parametric functions $x(t)$ and $y(t)$ as well as a
contour image (Fig.\ref{fig:comp_alfa_BSCEA_cont}).

\end{itemize}

These procedures are detailed in Sections \ref{sec-preproc},
\ref{sec-bta} and \ref{sec-contourextraction}, respectively.

\section{General Framework}
\label{sec-general}

\begin{figure}[!htb]
\centering
  \centerline{\subfigure[]{\includegraphics[width=15mm]{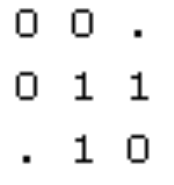}
\label{fig:figstrucelem_a}}
\hfil
\subfigure[]{\includegraphics[width=40mm]{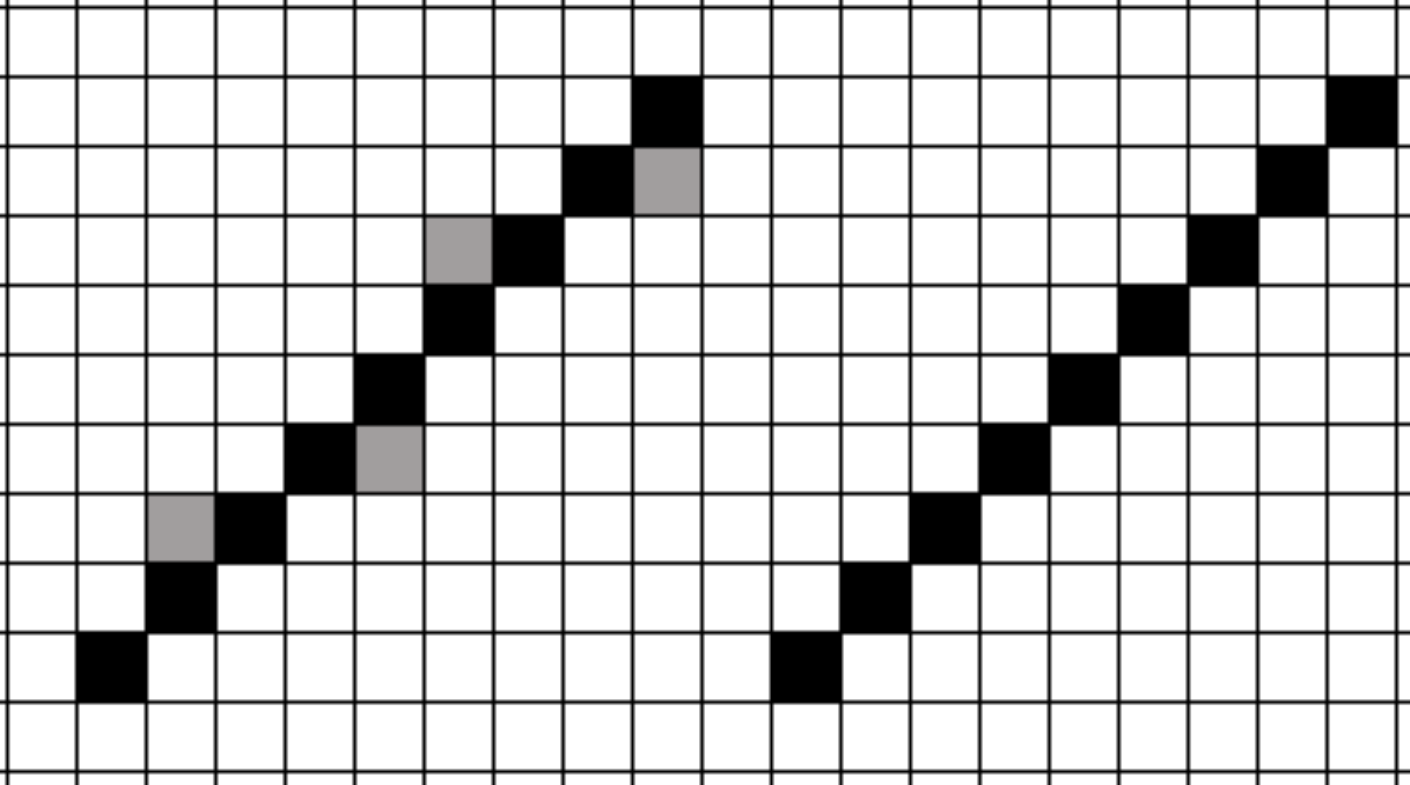}
\label{fig:figstrucelem_b}}}
     \caption{(a) Hit-or-Miss template used to filter the pruned skeleton,
     resulting in an $8$-connected skeleton with one-pixel wide branches. (b) The light 
     shaded pixels at the left-hand side must be removed yielding
     the essential structure at the right-hand side.}
      \label{fig:strucelem} 
\end{figure}

\subsection{Preprocessing} 
\label{sec-preproc}

Some important shape parts are detected by taking into account
specific features, such as the number of each pixel's neighbors and
the size of the shape. For example, pixels of branches are expected to
have only $2$ neighbors each, while critical regions and the soma have
more. Moreover, the soma area is greater than the areas of the
critical regions.

\begin{figure}[!htb]
\centering
  \centerline{\includegraphics[width=12cm]{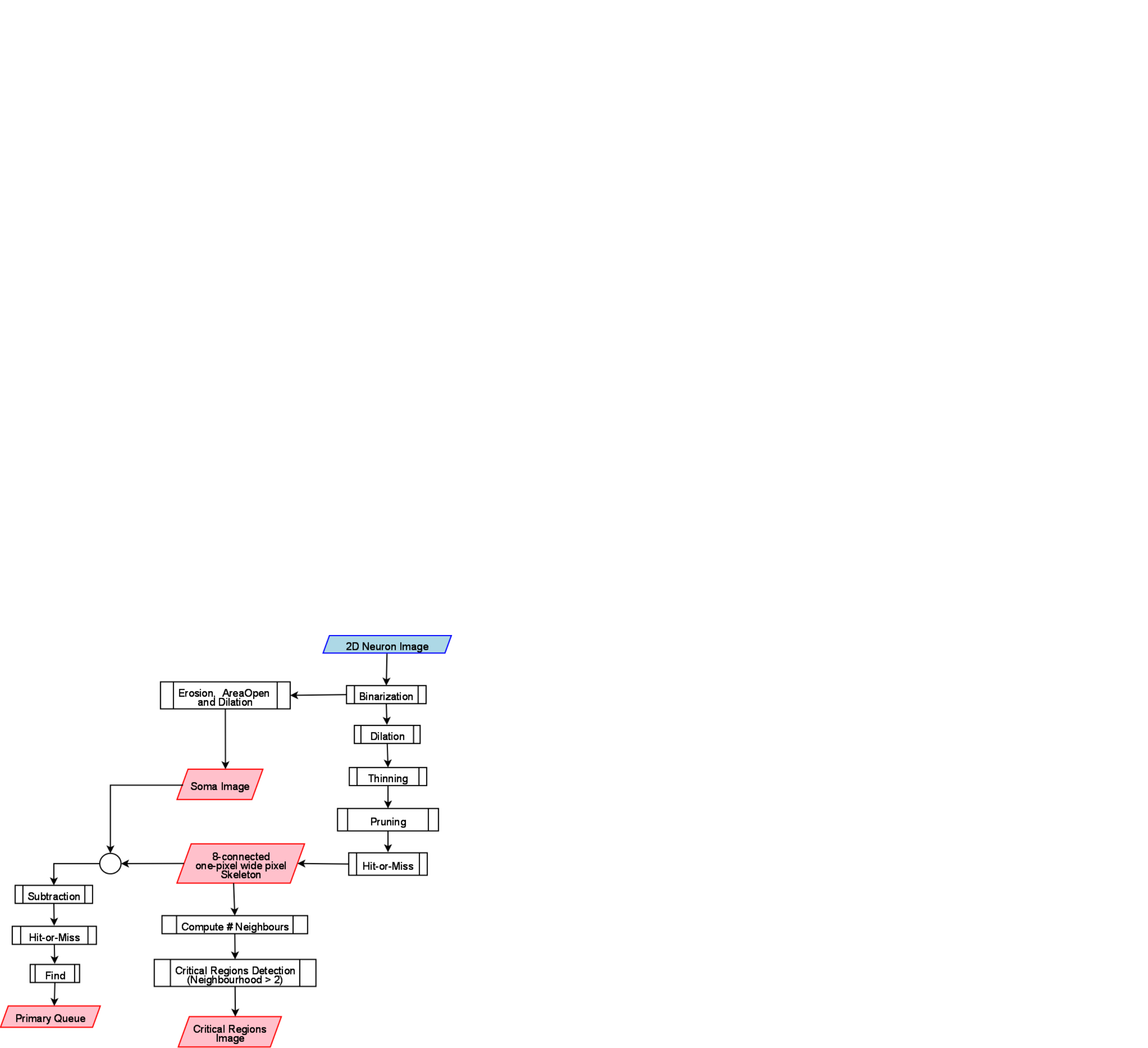}}
     \caption{Flowchart of the preprocessing pipeline. The red
     polygons represent the outcomes.}  \label{fig:preprocessing-eng-dia}
\end{figure}

Initially, a preprocessing pipeline involving mathematical morphology
transformations~\footnote{The reader is referred to \protect
\cite{morphlotufo,soille99} for details on the mathematical morphology
operations.} is carried out on the input image, so as to obtain the
separate components of the neuron image, that is the skeleton
comprised of $8$-connected one-pixel-wide branches, the critical
regions, the terminations, the soma and the queue of primary seeds.
The referred separate components on different images are obtained as
described in the flowchart diagram depicted in the
Fig.~\ref{fig:preprocessing-eng-dia}.

All the used structuring elements are flat and centered at their origins, i.e. their centroids.

In order to isolate the soma, the image is eroded by a disk with
radius $3$, followed by a dilation with a disk with radius $1$. It is
known that soma shapes do not follow a clear pattern, making their
segmentation critical. Herein, the soma segmentation is attained
through erosion, noise filtering by area opening, followed by a
dilation. Casual noisy pixels surrounding the soma image are wiped out
through the skeleton area opening. Then, additional processing is
applied in order to obtain an $8$-connected skeleton with one-pixel
wide branches~\cite{leandro:2008}(\ref{fig:preprocresults}).

The most critical and perhaps difficult template to define would be that 
portrayed in Fig.~\ref{fig:strucelem} for the Hit-or-Miss operation. The Hit-or-Miss is 
a mathematical morphology operation~\cite{morphlotufo}, being a sort of loose template matching, 
because the template itself is an interval, instead of a specific shape. 
Whenever certain small structure present on the image fits inside this interval, it is marked. 
Herein, the Hit-or-Miss operation is applied using the template depicted in Fig.~\ref{fig:figstrucelem_a} 
to detect redundant skeleton pixels which should be ruled out, as shown in Fig.~\ref{fig:figstrucelem_b}.

\begin{figure*}[htbp]
\centering  
  \centerline{\subfigure[]{\includegraphics[width=40mm]{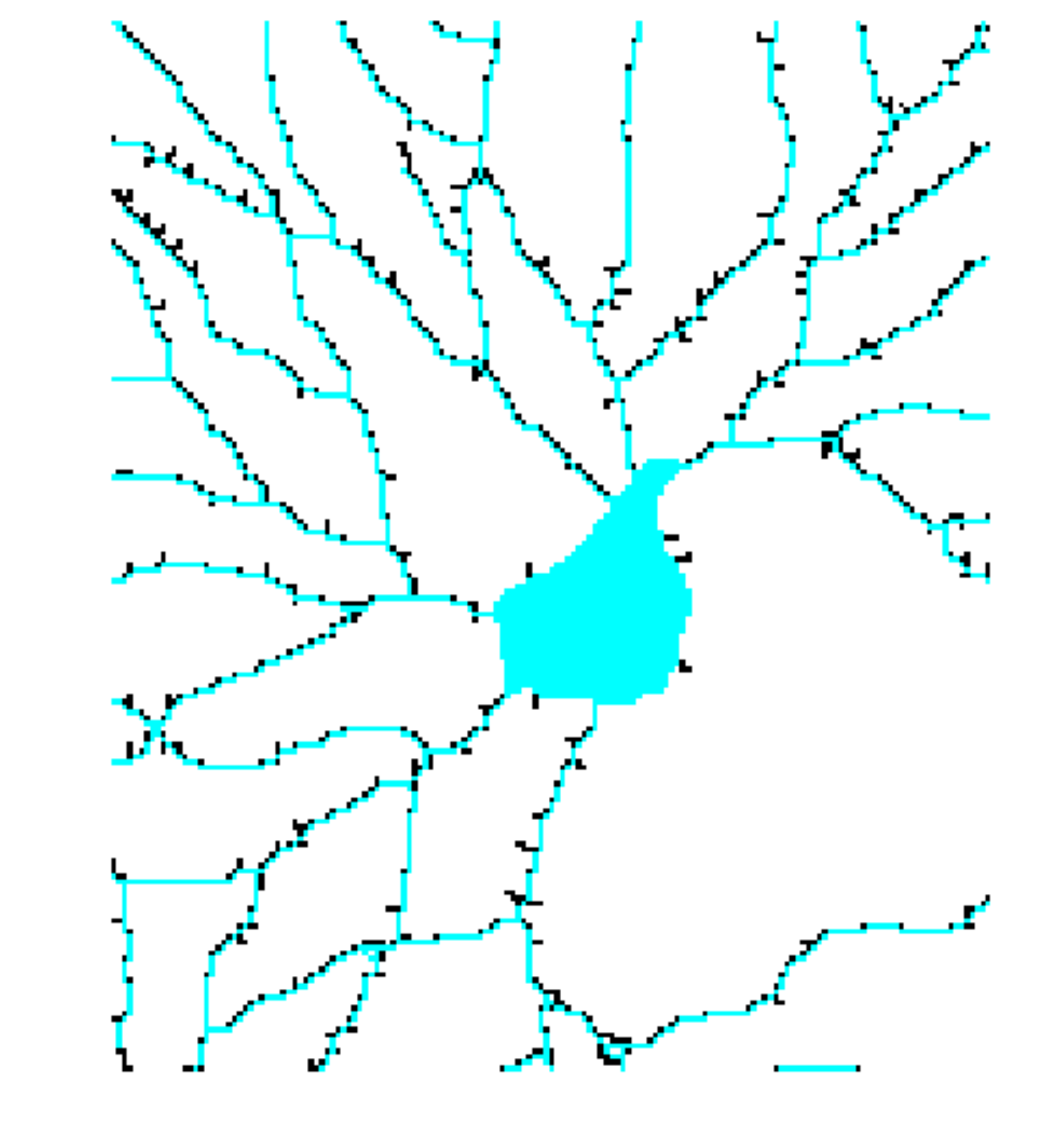}
\label{figpreproc_a}}
\hfil
\subfigure[]{\includegraphics[width=40mm]{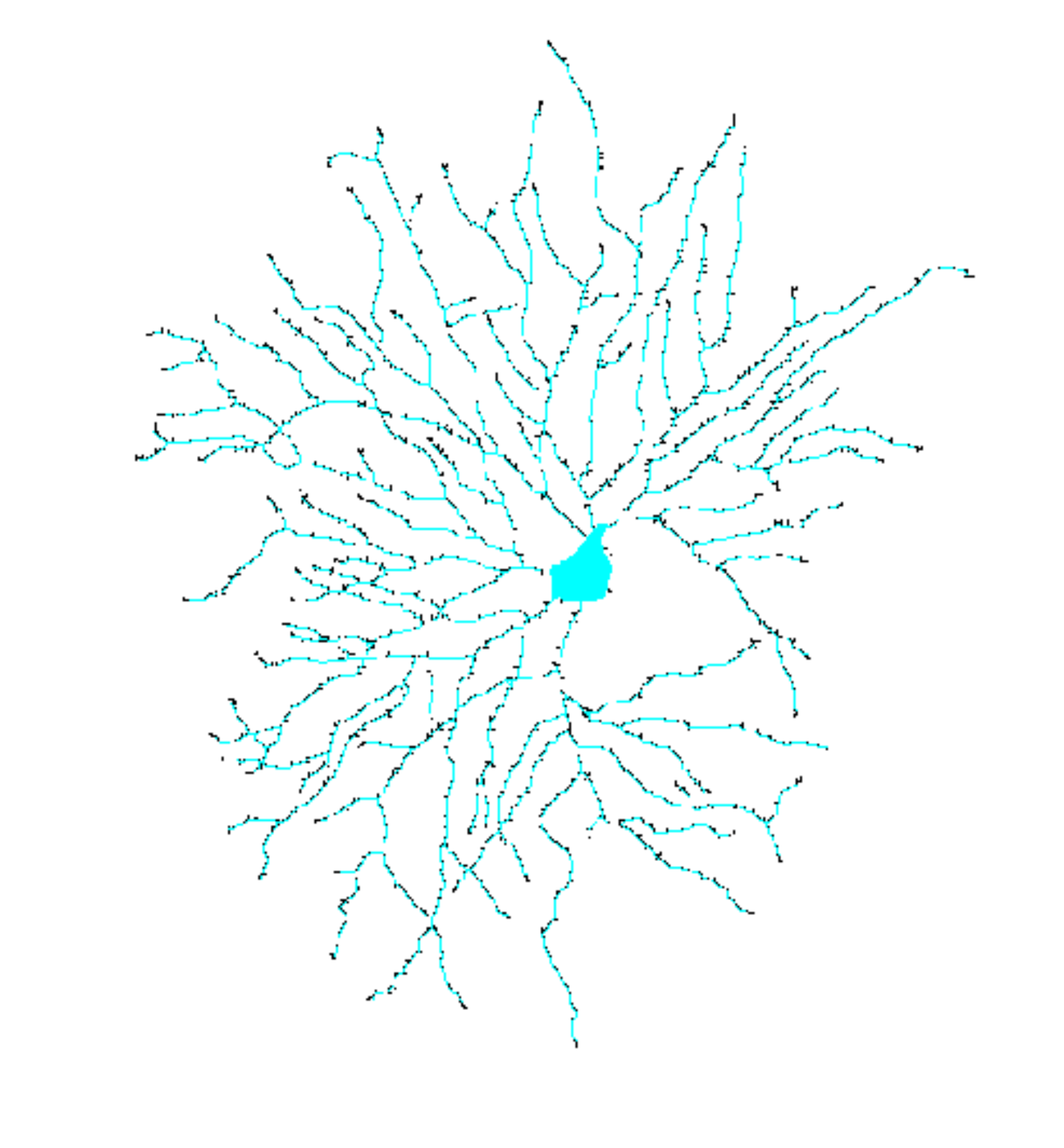}
\label{figpreproc_b}}}

 \centerline{\subfigure[]{\includegraphics[width=40mm]{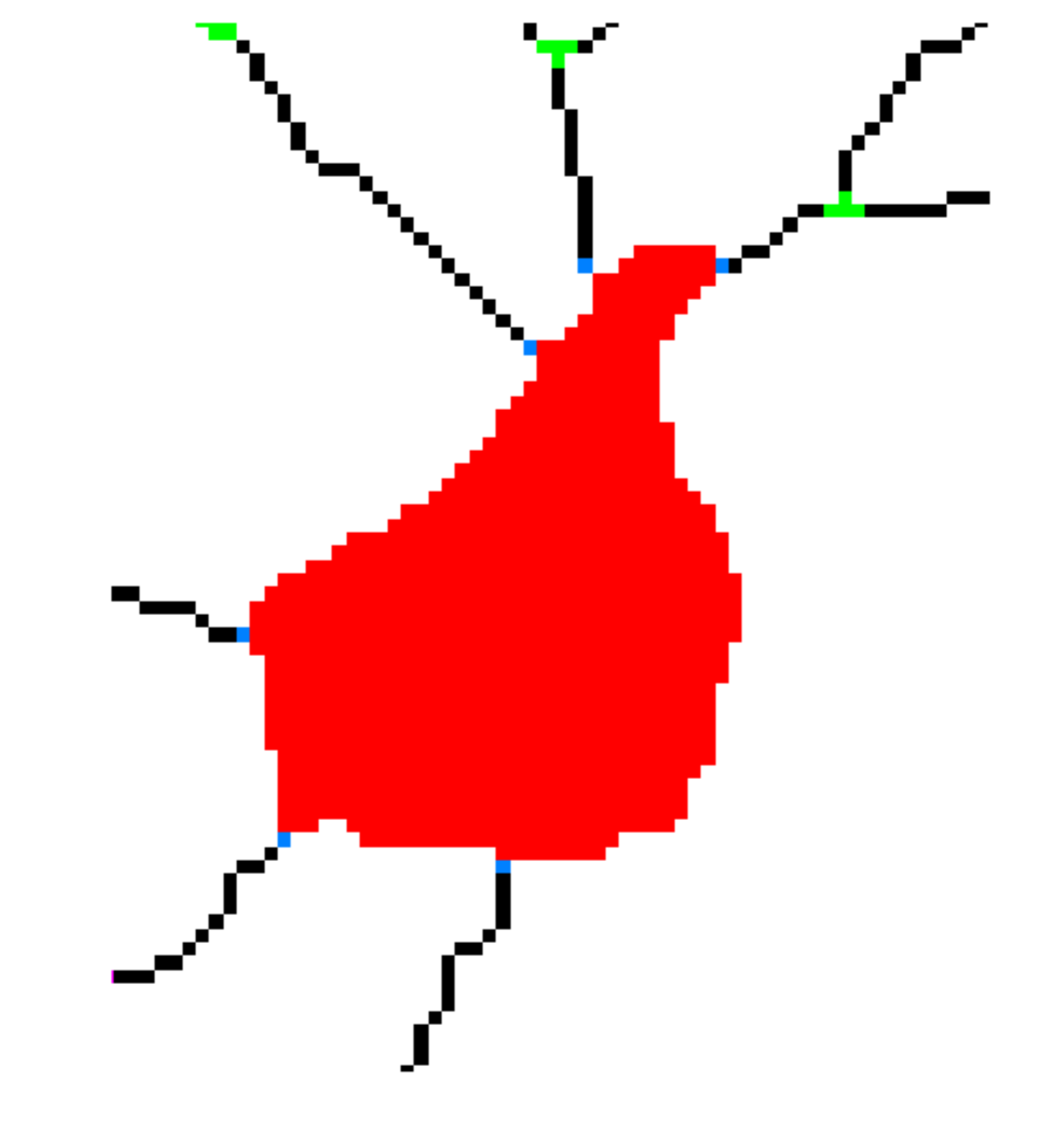}
 \label{figpreproc_c}}
\hfil
\subfigure[]{\includegraphics[width=40mm]{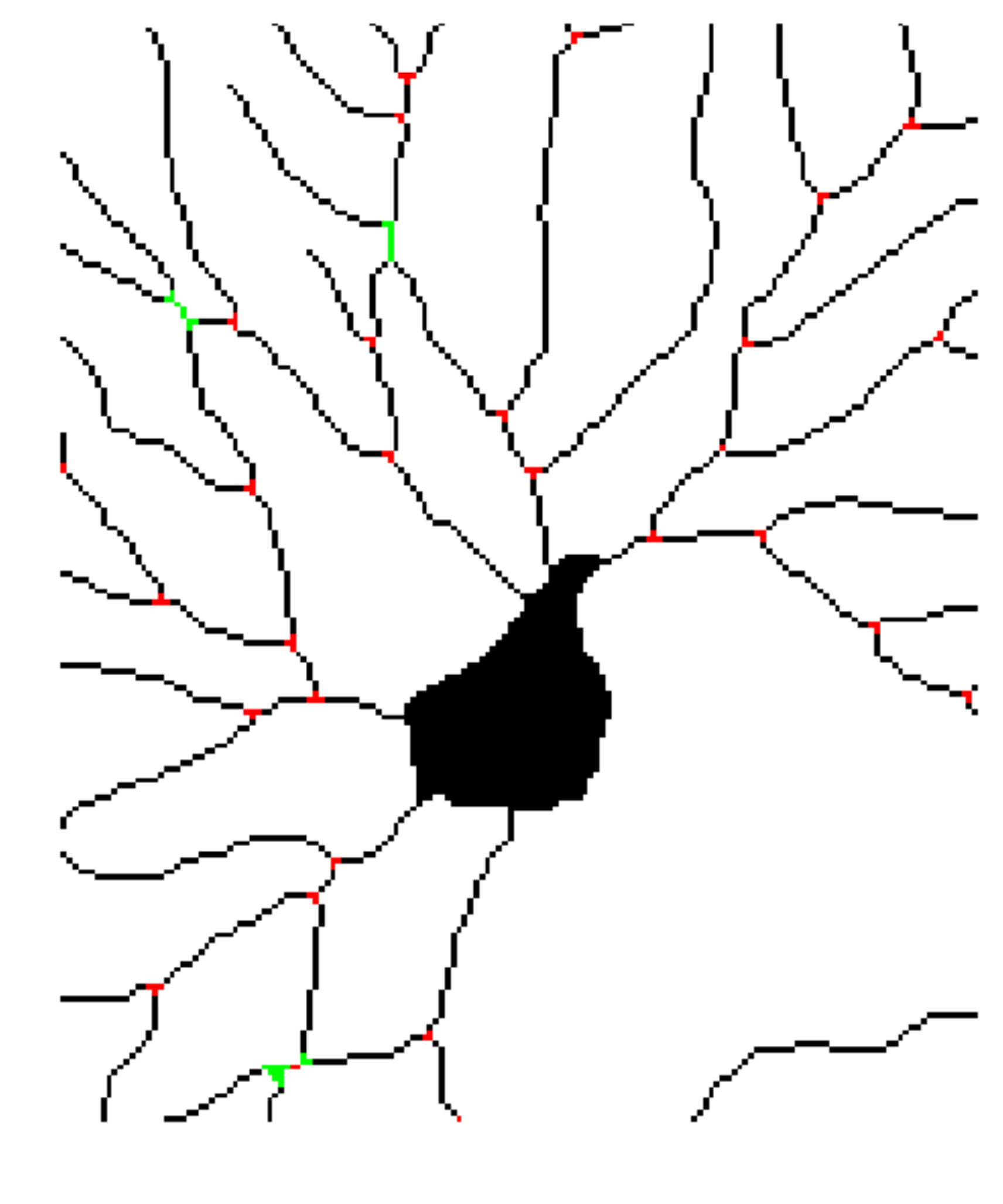}
\label{figpreproc_d}}}
    \caption{Preprocessing results: (a) The darkest pixels were removed
    by the \emph{Hit-or-Miss} filtering yielding the $8$-connected
    skeleton with one-pixel wide branches shown in lighter cyan; (b) Pruned
    $8$-connected skeleton (cyan) with one-pixel wide branches superimposed to the
    skeleton (black); (c) Soma (red), seeds (blue), critical regions
    (green) and skeleton(black); (d) Critical Regions (green and red)
    and skeleton (black).}~\label{fig:preprocessing}
\end{figure*}

\subsection{Tracking of Branches} 
\label{sec-bta}

One of the main goals at this stage is to label each dendritic branch
as a whole object on its own.  This is achieved by pixel-by-pixel
labeling of each branch.  
Considering the sequential nature of such a
processing, this problem may be described as estimating the spatial
coordinates $(x,y)$ of each subsequent branch pixel.  
Because this is analogous to tracking problems~\cite{arulampalam02tutorial} in the
computer vision literature, this algorithm is called \emph{Branches
Tracking Algorithm (BTA)}.

\emph{Tracking} is usually divided into \emph{Prediction}, 
\emph{Measure} and \emph{Update} stages~\cite{arulampalam02tutorial}.
During the Prediction stage, the algorithm estimates the next state of
the system. On the Measure stage, the algorithm probes the system by
looking for plausible states nearby, in this case valid pixels,
through some measures, herein the spatial coordinates $(x,y)$ of
pixels.  During the Update stage, the algorithm merges both pieces of
information gathered on the previous two stages, through a linear
combination, giving as a result the optimal estimation for the next
state.  So, in terms of Tracking, the \emph{BTA} Prediction and
Measure stages are carried out in a single step, through the
$8$-neighborhood scanning by using the chain-code~\cite{costabook01}.

The \emph{BTA} Update stage is related to the pixel labeling.  This
stage labels each dendritic subtree growing out of the soma in the
same way, i.e. by starting from the related primary seed and labeling
the entire branch adjacent to it, up to its termination. Meanwhile,
its branches are marked to be labeled afterwards. Thereafter,
every branch is labeled as the first branch was, and the respective
encountered branches are similarly marked to be labeled afterwards in a
recursive-like fashion until the whole dendritic subtree is labeled.

\begin{figure}[!htbp]
\centering
  \centerline{\includegraphics[height=21cm]{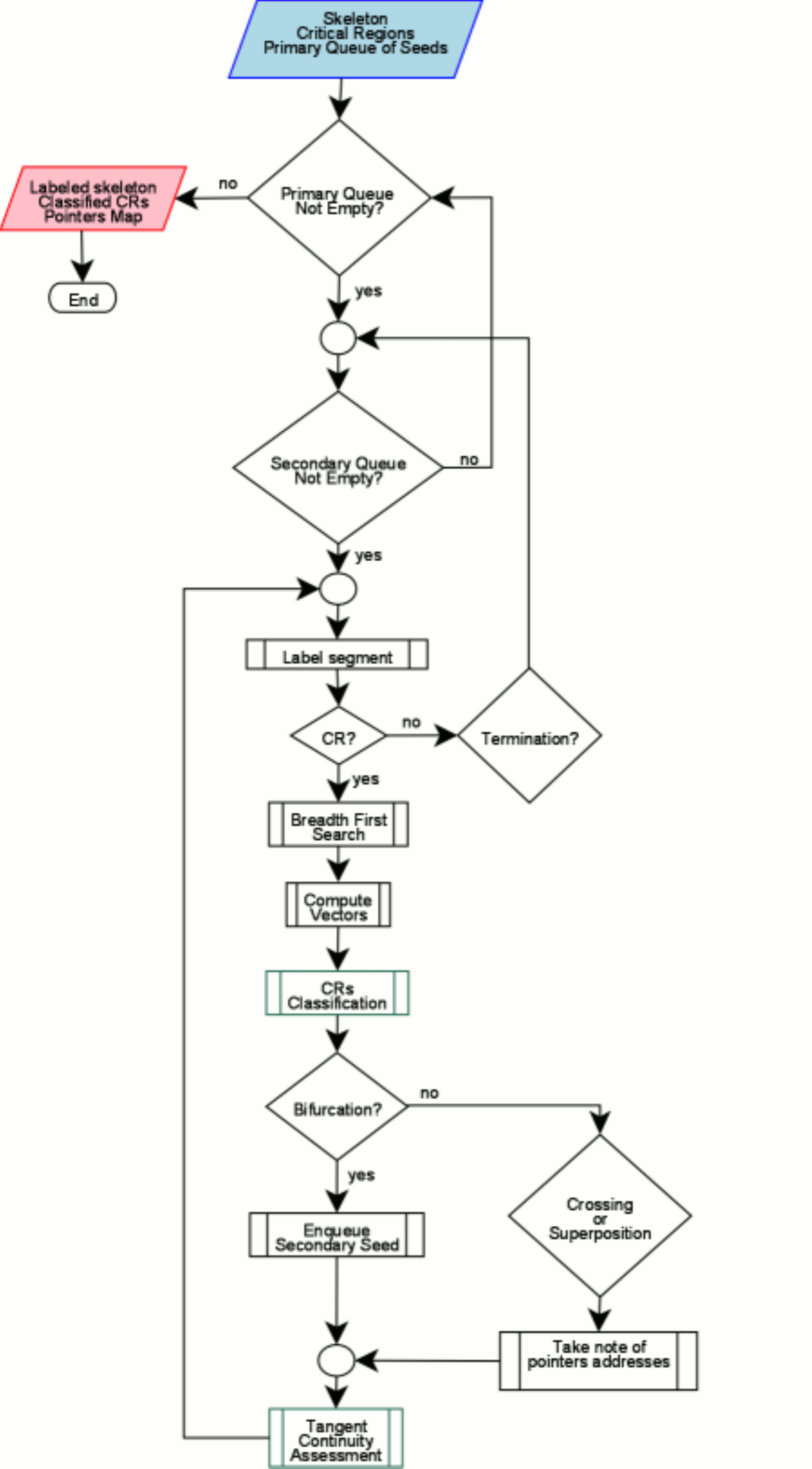}}
     \caption{Flowchart of \emph{BTA}.}
      \label{fig:bta-dia-eng} 
\end{figure}

The \emph{BTA} is mainly composed of two nested loops. 
The outermost loop is on primary seeds, being related to the labeling of 
each dendrite having root in the soma. The innermost loop is on secondary seeds, 
being related to the labeling of each branch within a given dendrite. This algorithm 
is depicted in the flowchart of Fig.~\ref{fig:bta-dia-eng}. It is worth mentioning that, for our purposes, valid pixels are defined as
simultaneously non-labeled and non-critical foreground pixels.  Then, 
for each primary seed, the \emph{BTA} starts
by subsequently stacking every valid pixel from a segment to be labeled afterwards,
until either a termination or a critical region is reached.

On arriving at a critical region, the \emph{BTA} may perform one or
two of the following tasks, \emph{Continuity of the Tangent
Orientation Assessment} and
\emph{Critical Regions Classification}.
The former (detailed in the Section~\ref{sec-continuity}) is always
carried out, while the latter (described in the
Section~\ref{sec-classificaregioes}) is performed only if the current
critical region has not been classified yet.  Notice that though the
critical regions are now available from the previous preprocessing
step, they are not classified yet, i.e. we do not know which is a
bifurcation, a crossing or a superposition. This classification is
important for the contour extraction step.

\subsubsection{Continuity of the Tangent Orientation Assessment}
\label{sec-continuity}
Analogously to the tracking process during branches labeling 
as described in \ref{sec-bta}, this step also comprises Prediction, Measure and Update, 
however in a slightly different fashion. Coming to a critical region in 
this step is similar to approaching the occlusion case in tracking problems 
~\cite{gabriel03}, where different objects follow trajectories which apparently overlap. 

So, after arriving at a critical region, the Prediction stage is
performed by computing the \emph{inwards direction vector}
$\hat{v}_0$.  In the Measure stage the algorithm calculates all
\emph{outwards direction vectors} $\hat{v}_i$.  Finally, in the Update
stage the outcomes from the Prediction and Measure stages are merged.
The reason to do that is to estimate the best candidate segment among
all the alternatives so as to carry forward the tracking
procedure. This merging is achieved by the calculation of inner
products (projections) between the inwards direction vector
$\hat{v}_0$ and each outwards direction vector $\hat{v}_i$ (observe
that these two vectors have unitary magnitude), according to
Eq. \ref{eq:inner-prod}:

\begin{equation}
	\label{eq:inner-prod}		
	k = \argmax_{i} ( < \hat{v}_0 , \hat{v}_i > )	
\end{equation}

\noindent where the index $k$ is assigned to the direction vector 
$\hat{v}_k$ among $\hat{v}_i$, for which the inner product result is
the largest. The extremity points for properly computing each vector
are determined through a \emph{Breadth-First Search} approach \protect
\footnote{The Breadth-First Search is a suitable method to find the
shortest path between two nodes in a graph, implemented by using a
queue of nodes as a data structure~\cite{sedgewick:1983}.}.

Every time a critical region is encountered, the Breadth-First Search
is triggered and all the forward neighboring pixels are iteratively
enqueued into an auxiliary queue, while passing across the just
detected critical region. At each Breadth-First Search iteration, the
auxiliary queue is run through in search of critical pixels. The stop
condition for the Breadth-First Search is set beforehand as a number
$C$ of consecutive executions through the auxiliary queue without
finding any critical pixel. This procedure is detailed in an example in Appendix \ref{sec-bta-bfs-examples}.

The starting pixel of the optimum segment to proceed is lastly stacked
and labeled. Also, the alternative path origin is considered as a
secondary seed, that is a side branch seed to be enqueued in case a
bifurcation is detected. Conversely, in case either a superposition or
a crossing is detected, the next segment starting point $V_{n+1}$ and
the current segment last point $V_{n}$ (Fig.~\ref{fig:supcontour_b})
addresses are stored into the Pointers Map.

\subsubsection{Critical Regions Classification}
\label{sec-classificaregioes}

\begin{figure*}[htb]
\centering
\centerline{\includegraphics[width=100mm]{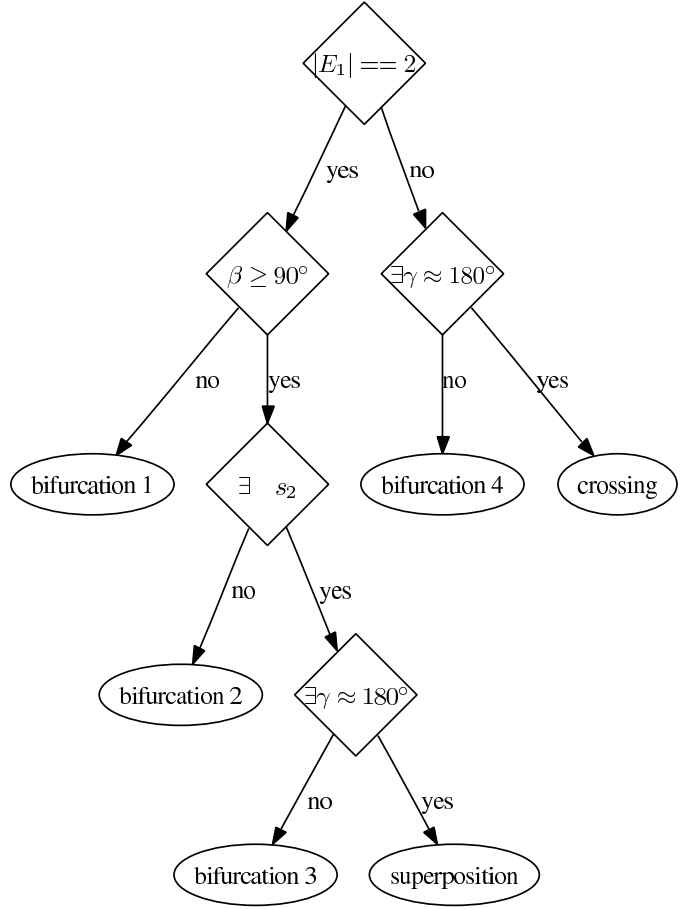}}
     \caption{Decision tree representing the sequence of rules
     applied for Critical Regions Classification by the
     \textbf{BTA}. The variables $s_1$ and $s_2$ represent
     respectively the current critical region and other nearby in the
     set $S$ of critical regions (see Fig. \protect
     \ref{fig:crclassification}). The variables $E_1$ and $E_2$ stand
     for the sets of unitary \textbf{outwards vectors} related to
     $s_1$ and $s_2$ respectively. Since a critical region is individually
     classified as it is found during the labeling process, it is
     clear that the current critical region $s_1$ exists. 
     The variable $\beta$ measures the angle between the inwards
     direction vector and the outwards direction vector, while the
     variable $\gamma$ measures the angle between any two outwards
     direction vectors. Refer to the Appendix~\protect \ref{sec-bta-cr-class-rules} for further explanation on notations.}
\label{fig:crclassificationdecisiontree}
\end{figure*}

While assessing the orientation of the tangent direction vectors at
each critical region, as described in section~\ref{sec-continuity},
the \emph{BTA} also gathers enough information to classify the current
critical region into one of the $6$ different classes (see
Fig. \ref{fig:crclassification}), which have been identified as being
critical for the skeletons. \emph{Critical Regions Classification} is
a crucial concept for the proper functioning of our \emph{Contour
Extraction} algorithm presented in
section~\ref{sec-contourextraction}. Although \emph{Critical Regions
Classification} is not an algorithm on its own, it is an important
part of the \emph{BTA}. Therefore, each critical region is classified
according to some special rules.  The decision tree depicted in
Fig. \ref{fig:crclassificationdecisiontree} details both the
classification rules themselves and the order in which they should be
considered, i.e. it illustrates the flow of decisions required to
properly classify a critical region into one of the $6$ classes showed
in Fig~\ref{fig:crclassification}.  These classes have been abstracted
from the analysis of several images, during the development of our
methodology. We started from the assumption that $2D$ branching
structures are comprised of only bifurcations $1$ and crossings. So
the number of adjacent segments ($3$ or $4$) at every critical region
should be enough to classify them.  However, misclassifications during
the system development implied a more complete description. The system
became more and more robust, as we moved further by taking into
account new pieces of information, such as orientation between
incoming and outgoing direction vectors, proximity relation between
neighbor crossing regions, besides the basic and first criterion of
number of adjacent segments to each crossing region.

In brief, the critical regions classification rules take into account
the angle $\beta$ between the inwards direction vector - $\hat{v}_0$ -
at the current critical region $s_1$ and its outwards direction vector
- $\hat{v}_2$; the angle $\gamma$ between any pair of outwards
direction vectors ${\hat{v}_i,\hat{v}_j}$; and the cardinality $|E_1|$
of the set of outwards direction vectors $E_1$ related to the current
critical region $s_1$.

\subsection{Contour Extraction}
\label{sec-contourextraction}

\begin{figure}[!htbp]
\centering
  \centerline{\includegraphics[height=21cm]{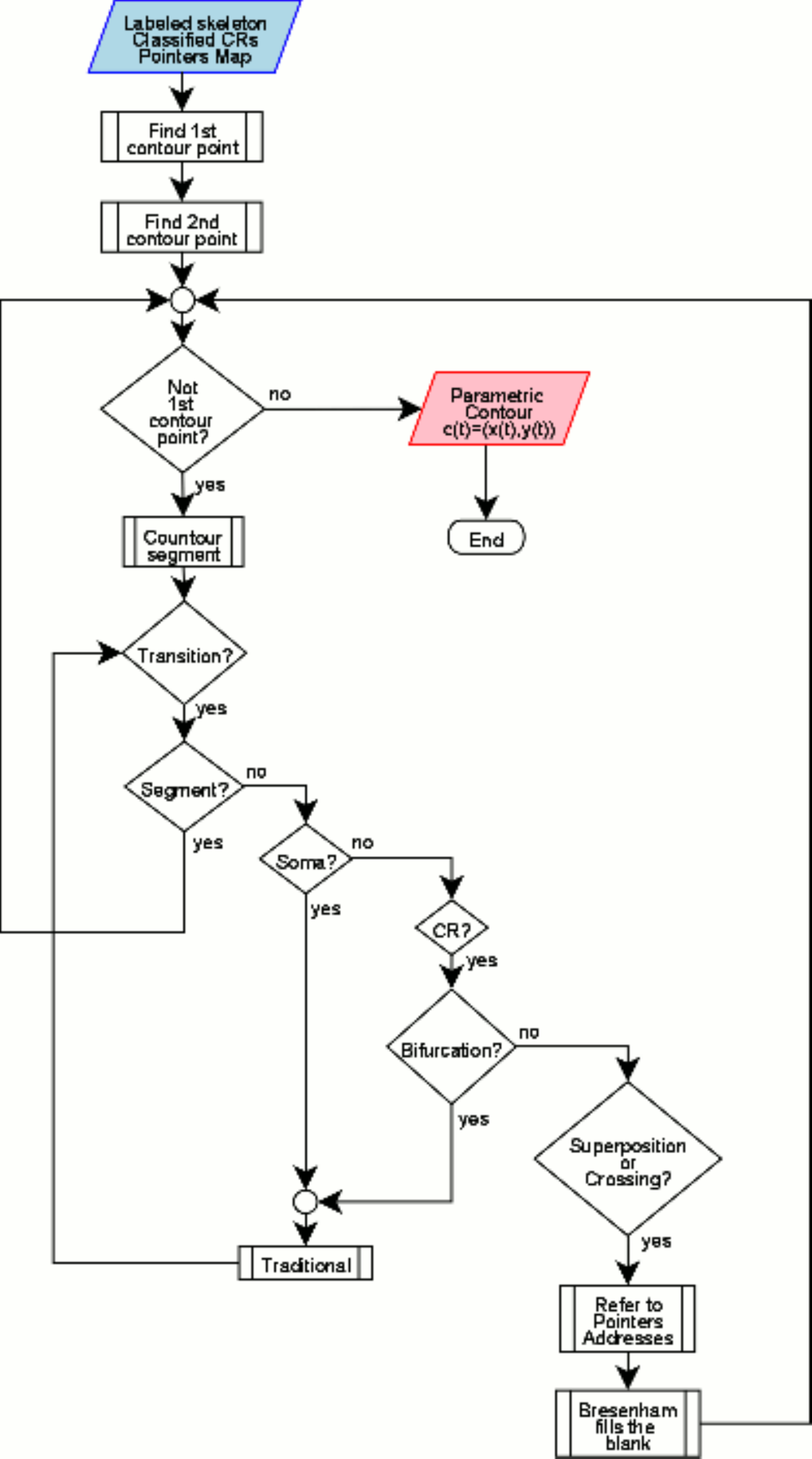}}
     \caption{Flowchart of the \emph{BSCEA}.}
      \label{fig:bscea-dia-eng} 
\end{figure}

Consider the following conventions for the \emph{Branching Structures
Contour Extraction Algorithm} (\textbf{BSCEA}) description:

\begin{enumerate}
\item[i] input: union of labeled skeleton and soma images.
\item[ii] directions related to the current $C$ pixel are identified
according to the chain-code.
\item[iii] the input is followed in a counter-clockwise sense.
\item[iv] all the $N$ points of the parametric contour are stored in a suitable data structure $E(1..N)$. 
Each element $E(n)$ keeps the $n^{th}$ contour point coordinates, 
i.e. $E(n).x$ and $E(n).y$, which are the computational representation 
for $x(t=n)$ and $y(t=n)$ respectively. When the contour is closed, $x(t=1)=x(t=N)$ 
and $y(t=1)=y(t=N)$.
\end{enumerate}

The main steps composing the \emph{BSCEA} are depicted on the
respective flowchart in Fig.~\ref{fig:bscea-dia-eng}. The contour
following algorithm explained in \cite{costabook01} has been adopted
in this work.

\subsubsection{Finding the first pixel}
\label{sec-firstpixel}
The \textbf{BSCEA} starts by a raster scanning, i.e., from left to
the right, from top to the bottom, in search of the first contour
pixel $E(1)$, which should be the first background pixel found that is
also a neighbor of a foreground pixel.  In the sequel, the BSCEA will
contour the shape all the way, until coming back to the
first pixel, closing the cycle and having $E(1)=E(N)$.
\subsubsection{Finding the next pixel}
\label{sec-nextpixel}
\begin{figure}[htb] 
\centering
\includegraphics[width=80mm]{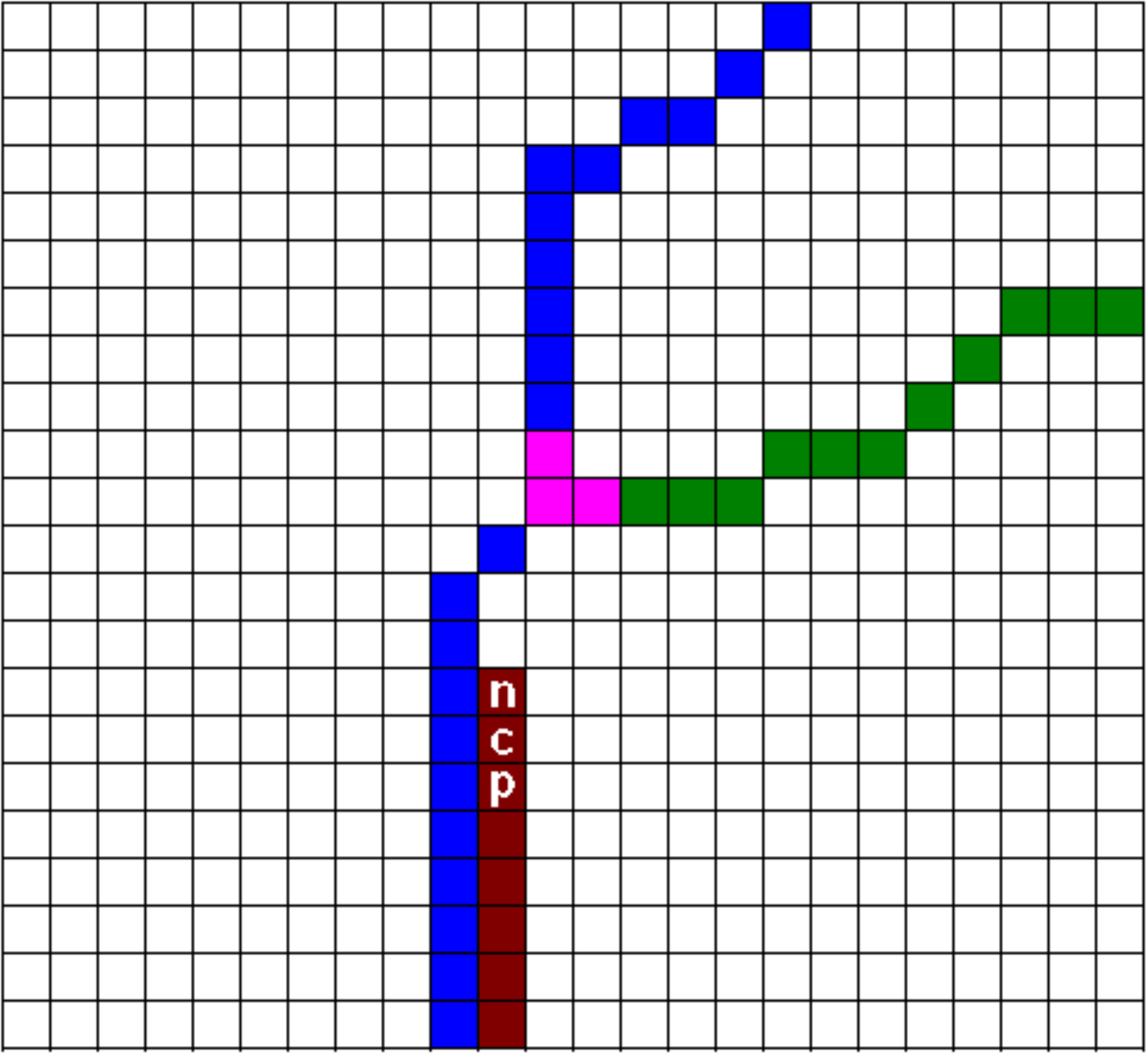}
  \caption{Previous (\textbf{p}), Current (\textbf{c}) and Next (\textbf{n}) pixels
  represented in an iteration of the \emph{BSCEA}. The direction relationship $d_{pc}$ between 
  the pixels \textbf{p} and \textbf{c}, besides the labels assigned to the segments, determine 
  the Next pixel \textbf{n}.}
  \label{fig:previouscurrentnext}
\end{figure}

From the second pixel on, the chain-code will be used to scan the
current pixel vicinity.  In so doing, the second contour pixel will be
the first neighbor in the chain-code sequence $4, 5, 6, 7, 8$, which is also
a background pixel and a neighbor of a foreground pixel. Herein this scanning in search 
of the next pixel is done analogously to that by the traditional
contour extraction algorithm~\cite{costabook01}. 
Besides direction relationship between \emph{current} and \emph{previous pixels} (see Fig.\ref{fig:previouscurrentnext})
to properly decide about the next contour pixel, it shall be considered the transition between labels,
so as to know if the \emph{BSCEA} is contouring a branch, the soma or a critical region.
The \emph{BSCEA} contouring strategy is in accordance with the specific structure being contoured.
So, the main \emph{BSCEA} parameters are:
\begin{itemize}
\item the current contour pixel $E(n)$
\item the direction $d_{cp}$ from the current to the previous pixel
\item the previous pixel label 
\item the current pixel label
\end{itemize}
Providing the \emph{BSCEA} with the $E(n)$ pixel and direction
$d_{cp}$ allows the identification of the starting point to scan the
neighborhood in search of the next pixel~\cite{costabook01}, according to the chain-code
sequence.

Since the input for the \emph{BSCEA} is a union of the
labeled skeleton and the soma images, it is necessary to adopt a
policy to properly find the next pixel in each case. Hence, the
\emph{BSCEA} considers contouring branches as the default case, 
taking the first background pixel which is also neighbor of a
foreground pixel in the neighborhood defined by the chain-code.
Conversely, the \emph{BSCEA} considers contouring the soma as a
particular case, taking the last pixel, instead of the first one, to
be included as contour.  By so doing, the \emph{BSCEA} is able to
contour branches, while preserving the ability of more traditional
approaches to circumvent the problem of contouring occasional
one-pixel wide entrances into the soma, consequently allowing the
contour to be closed~\cite{costabook01}.

\subsubsection{Traversing critical regions}
\label{sec-traversing}

It is also necessary to devise a strategy for critical regions
processing, according to their classes, as described in section
\ref{sec-classificaregioes}.  Regions classified as Bifurcation
should be contoured outwards, while those ones classified as
either Superposition or Crossing should be contoured
inwards, through pointer addresses written to the
\emph{Pointers Map} data structure during the tracking stage.  
The integration between soma and labeled skeleton is critical for the
successful contour extraction, since it guarantees the contour
closing.

\begin{figure*}[htb]
\centering
  \centerline{\subfigure[]{\includegraphics[width=50mm]{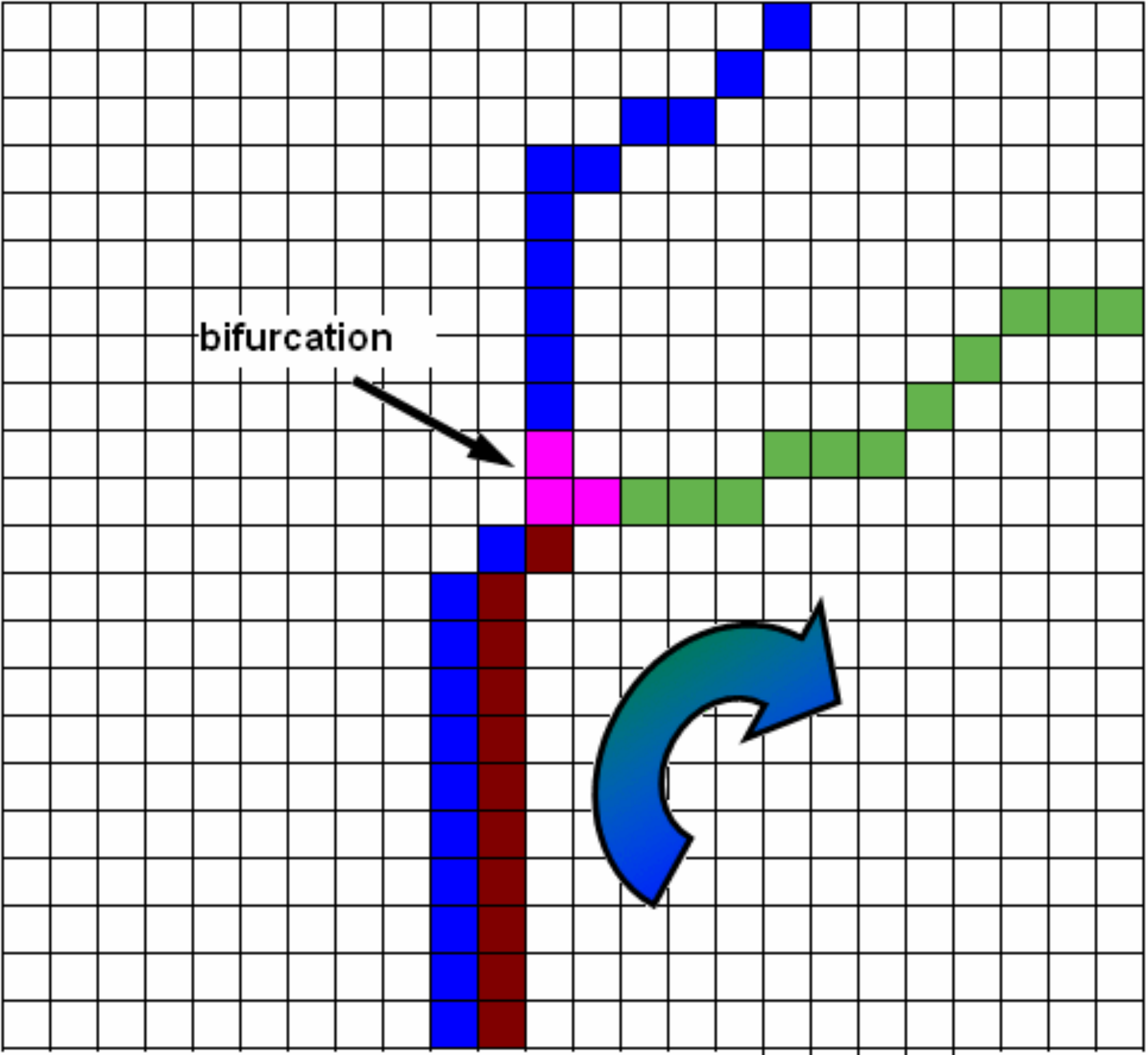}
\label{fig:bif1contour_a}}
\hfil
\subfigure[]{\includegraphics[width=50mm]{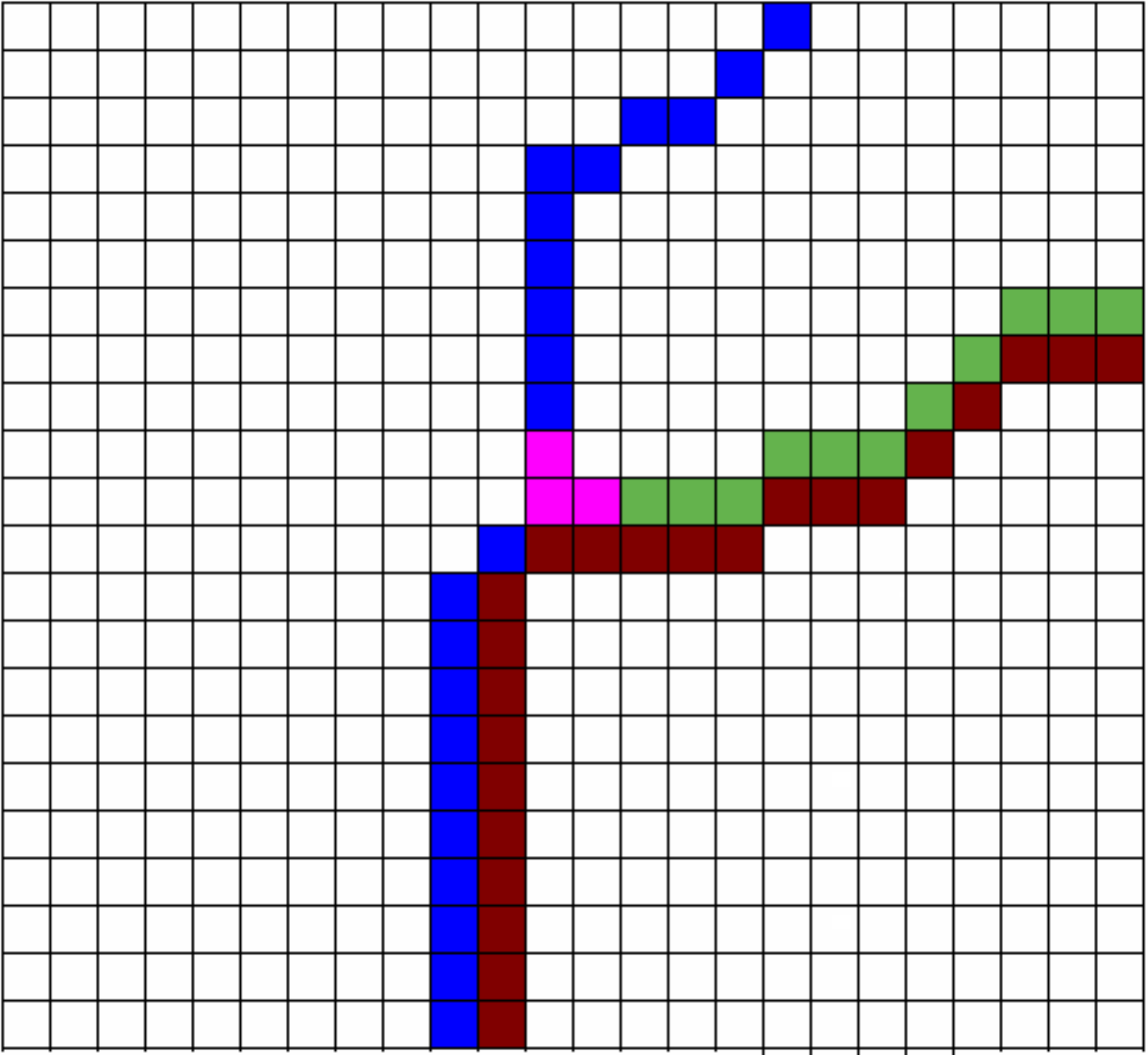}
\label{fig:bif1contour_b}}}
  \caption{Contouring a bifurcation. Branches appear labeled in blue
  and green, while the critical region previously classified as a
  bifurcation appears in magenta. The contour is shown in brown. (a)
  By detecting labels transition, the \emph{BSCEA} identifies that it
  has arrived at a bifurcation, thus deciding to contour the shape
  outwards. (b) Having left the critical region behind, it proceeds
  until reaching another critical region.}
\label{fig:bif1contour}
\end{figure*}

\begin{figure*}[htbp]
\centering  
  \centerline{\subfigure[]{\includegraphics[width=50mm]{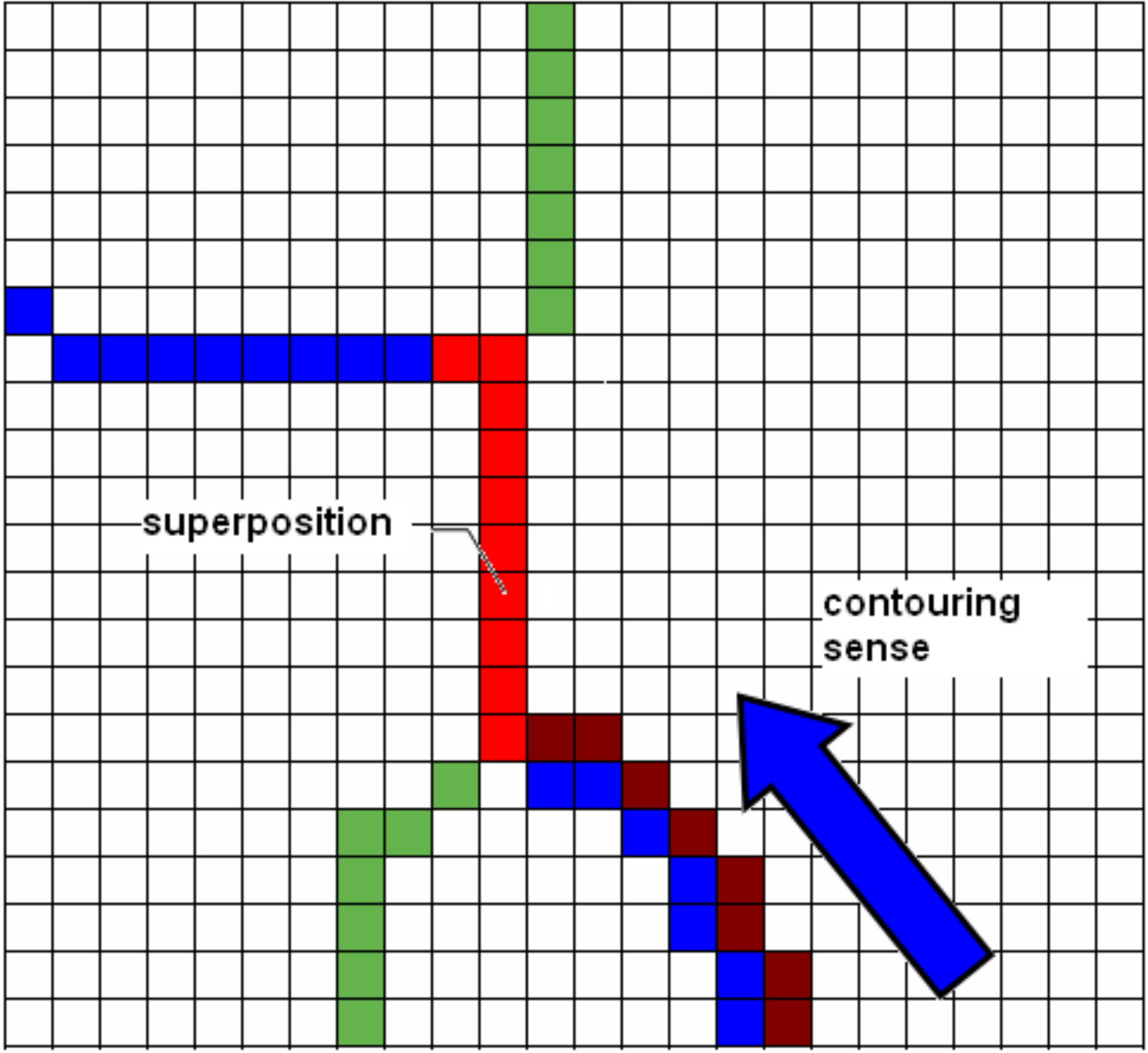}
\label{fig:supcontour_a}}
\hfil
\subfigure[]{\includegraphics[width=50mm]{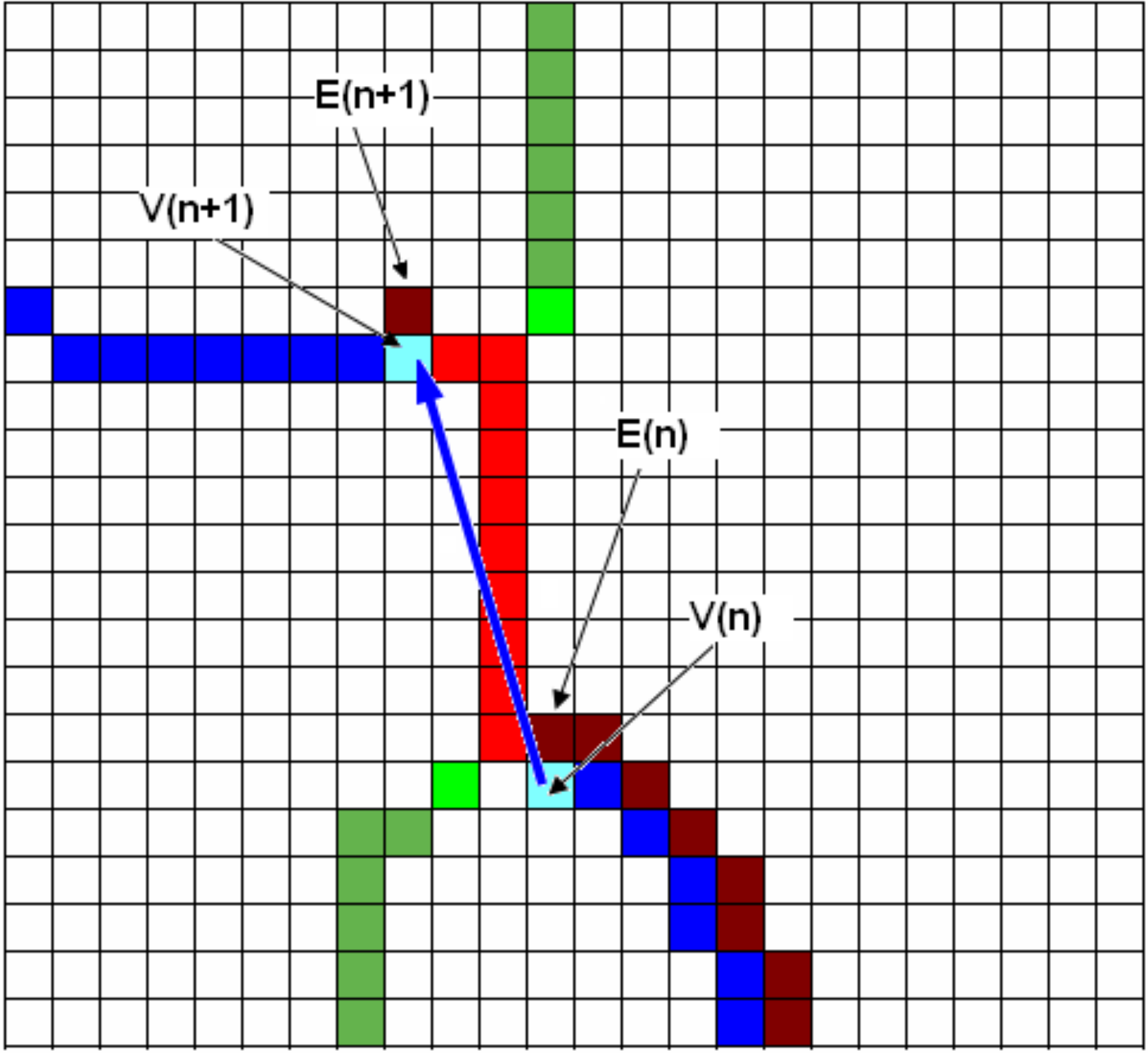}
\label{fig:supcontour_b}}}
 \centerline{\subfigure[]{\includegraphics[width=50mm]{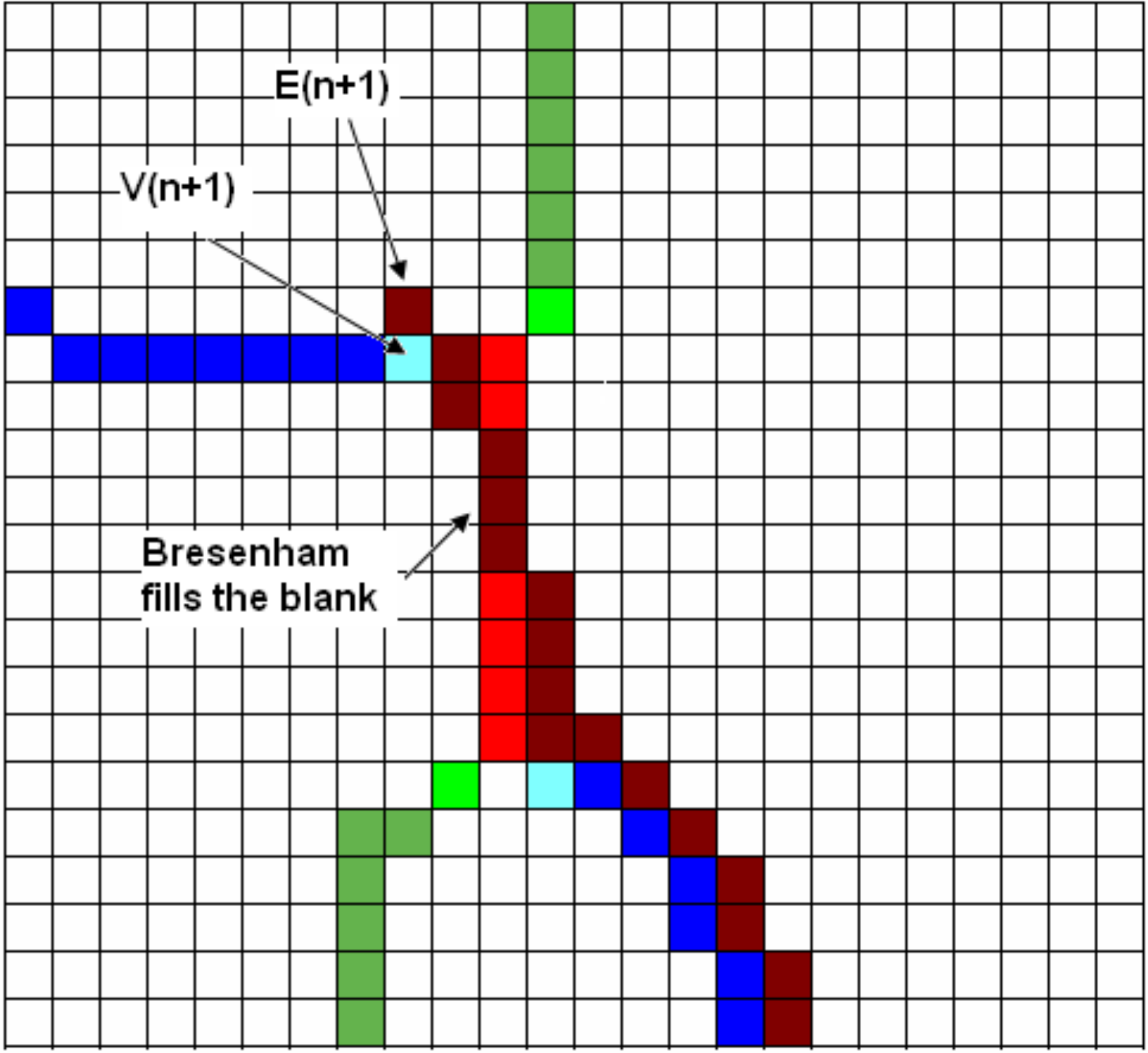}
 \label{fig:supcontour_c}}
\hfil
\subfigure[]{\includegraphics[width=50mm]{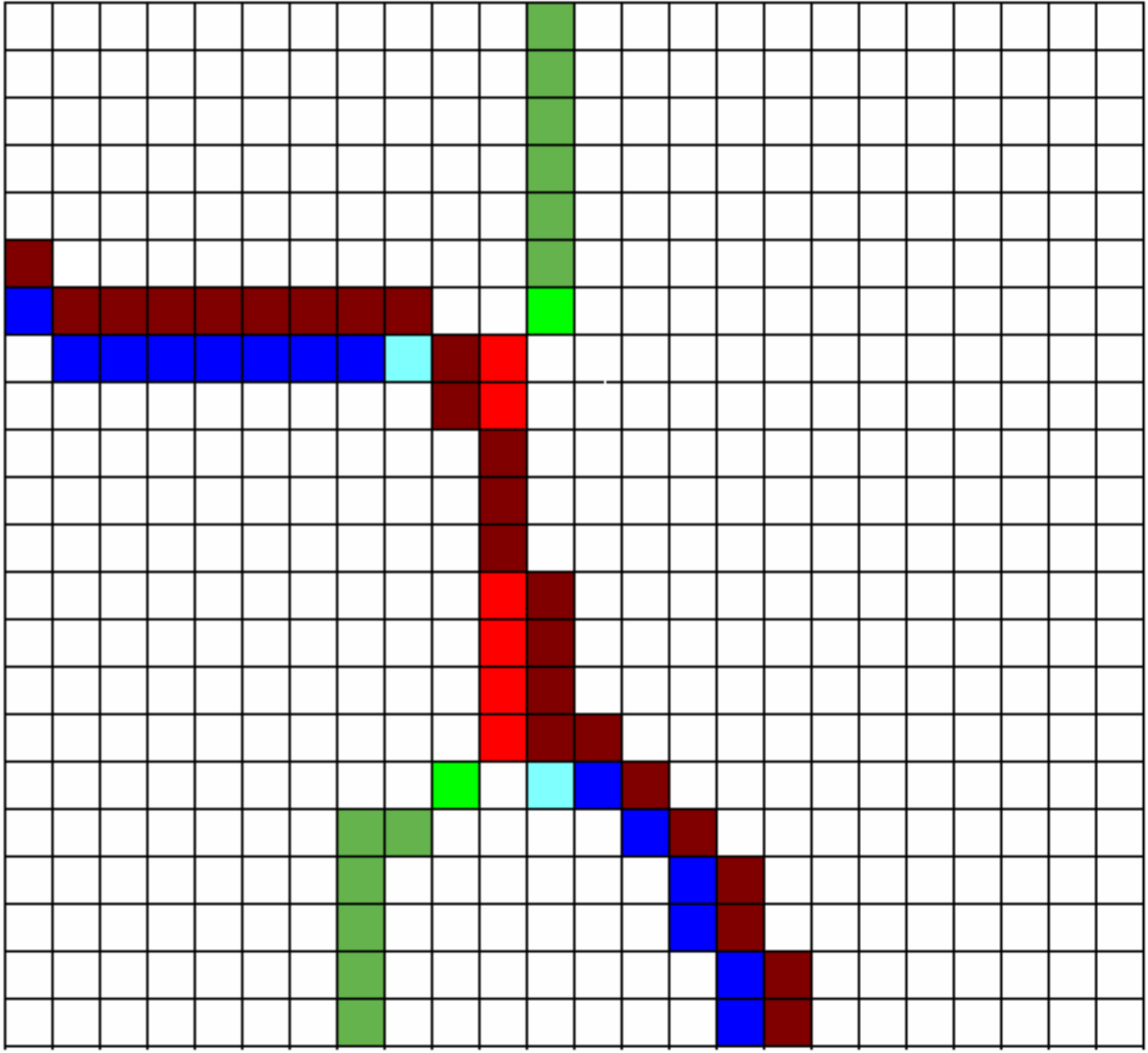}
\label{fig:supcontour_d}}}
    \caption{Contouring an overlap. Branches appear labeled in
    blue and green, while the critical region previously classified as
    a superposition appears in red. The contour is shown in brown.
    (a) By detecting labels transition, the \emph{BSCEA} identifies
    that it has arrived at an overlap, thus deciding to contour
    the shape inwards.  (b) Firstly the \emph{BSCEA} looks in the
    vicinity for the pointer $P_n$, related to the current segment end
    point pixel $V_n$. The pointer $P_n$ stores the address of the
    next segment pixel beyond the critical region, namely
    $V_{n+1}$. Subsequently the algorithm determines the direction
    relationship $d_{E(n) \Longleftrightarrow V_{n}}$.  Assuming
    $d_{E(n+1) \Longleftrightarrow V_{n+1}}=d_{E(n)
    \Longleftrightarrow V_{n}}$ , it finds the next segment starting
    pixel $E(n+1)$. (c) The Bresenham's algorithm is applied to trace
    a digital line between $E(n)$ and $E(n+1)$, filling the blank in
    the parametric contour signal over the \emph{critical region}.
    (d) Having left the critical region behind, it proceeds until
    reaching another critical region.}
    \label{fig:supcontour}
\end{figure*}

The \emph{BSCEA} can deal with both cases by taking into account the
labels of previous and current pixels, which convey valuable
information concerning particular situations, i.e.  if the critical
region is a bifurcation, "contour it outwards" (see
Fig.~\ref{fig:bscea-dia-eng} and Fig.\ref{fig:bif1contour}), as well
as the traditional contour extraction algorithm
would~\cite{costabook01}. In case it is a superposition or a crossing,
"contour it inwards", (see Fig.~\ref{fig:bscea-dia-eng} and
Fig.~\ref{fig:supcontour}), which means to trace a line between the
current segment end point and the next segment starting point. Both
points are known from the pointers marked by the \emph{BTA}. The line
is traced by using the Bresenham algorithm~\cite{bresenham1965} for
tracing a digital straight line segment.

\begin{itemize}
\item \emph{case 1:} \textbf{BSCEA} is contouring some branch
\begin{itemize}
\item take the $1^{st}$ candidate in the chain-code sequence.
\end{itemize}
\item \emph{case 2:} \textbf{BSCEA} is at a transition between a branch and the soma
\begin{itemize}
\item take the last candidate in the chain-code sequence.
\end{itemize}
\item \emph{case 3:} \textbf{BSCEA} is at a transition between a branch and a critical region
\begin{itemize}
\item[(a)] if the critical region is a bifurcation, "contour it outwards" (see Fig.~\ref{fig:bscea-dia-eng} and Fig.\ref{fig:bif1contour}).
\item[(b)] if the critical region is either a superposition or a crossing, "contour it inwards", (see Fig.~\ref{fig:bscea-dia-eng} and Fig.~\ref{fig:supcontour}).
\end{itemize}
\end{itemize}

In the case $3$-a, "contour outwards" means contouring the shape as
the traditional contour extraction algorithm would~\cite{costabook01}, as shown in Fig.\ref{fig:bif1contour}.

In the case $3$-b, "contour inwards" means:
\begin{itemize}
\item probing the current pixel $E(n)$ vicinity in search of the respective pointer $P_n$ in the \emph{Pointers Map} data structure
\item determining the direction relashionship $d_{E(n) \Longleftrightarrow V_{n}}$ between $E(n)$ and the current segment end point pixel $V_n$
\item accessing the next segment starting pixel $V_{n+1}$, pointed to by $P_n$
\item assuming $d_{E(n+1) \Longleftrightarrow V_{n+1}}=d_{E(n) \Longleftrightarrow V_{n}}$ and finding $E_{n+1}$ accordingly
\item filling the blank in the contour over the critical region with a digital line between $E(n)$ and $E(n+1)$, 
by using the Bresenham's algorithm~\cite{bresenham1965}.
\end{itemize}

Notice that the \emph{BSCEA} cannot tell which pixels of a
superposition or crossing are related one another or to a branch,
since the projection from the $3D$ neuron onto the $2D$ plane
suppresses this information. Such a problem is circumvented by
replacing the shared pixels in the critical region by two short
intercepting segments given by the Bresenham's algorithm, as
illustrated in Fig.\ref{fig:supcontour}.

\section{Results}
\label{sec-results}

All the methods described in this work have been implemented as
\emph{Matlab}\TReg\ scripts, using the
\emph{SDC Morphology Toolbox for MatLab}\TTra\ \cite{mmorph:2006}. 
The overall method has been evaluated with respect to a data set
containing several images of ganglion cells in the retina of cats,
acquired by camera lucida. In order to reflect an important biological
investigation, we have chosen $2D$ neuron images from
\cite{masland01}.

\begin{figure*}[htb]
\centering
\centerline{\subfigure[]{\includegraphics[width=50mm]{alfa9}
\label{fig:alfa_a}}
\hfil
\subfigure[]{\includegraphics[width=50mm]{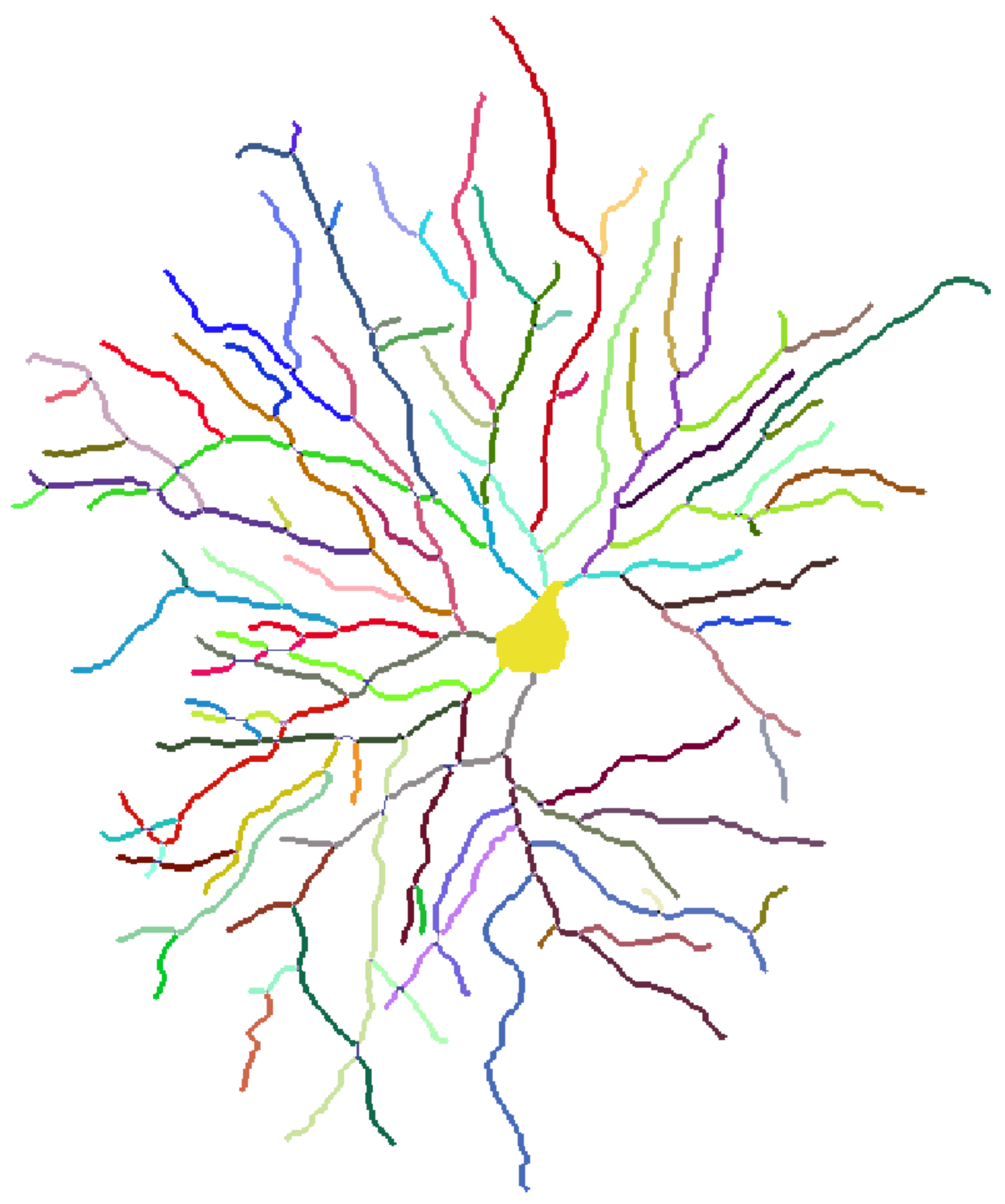}
\label{fig:alfa_b}}}

\centering
\centerline{\subfigure[]{\includegraphics[width=50mm]{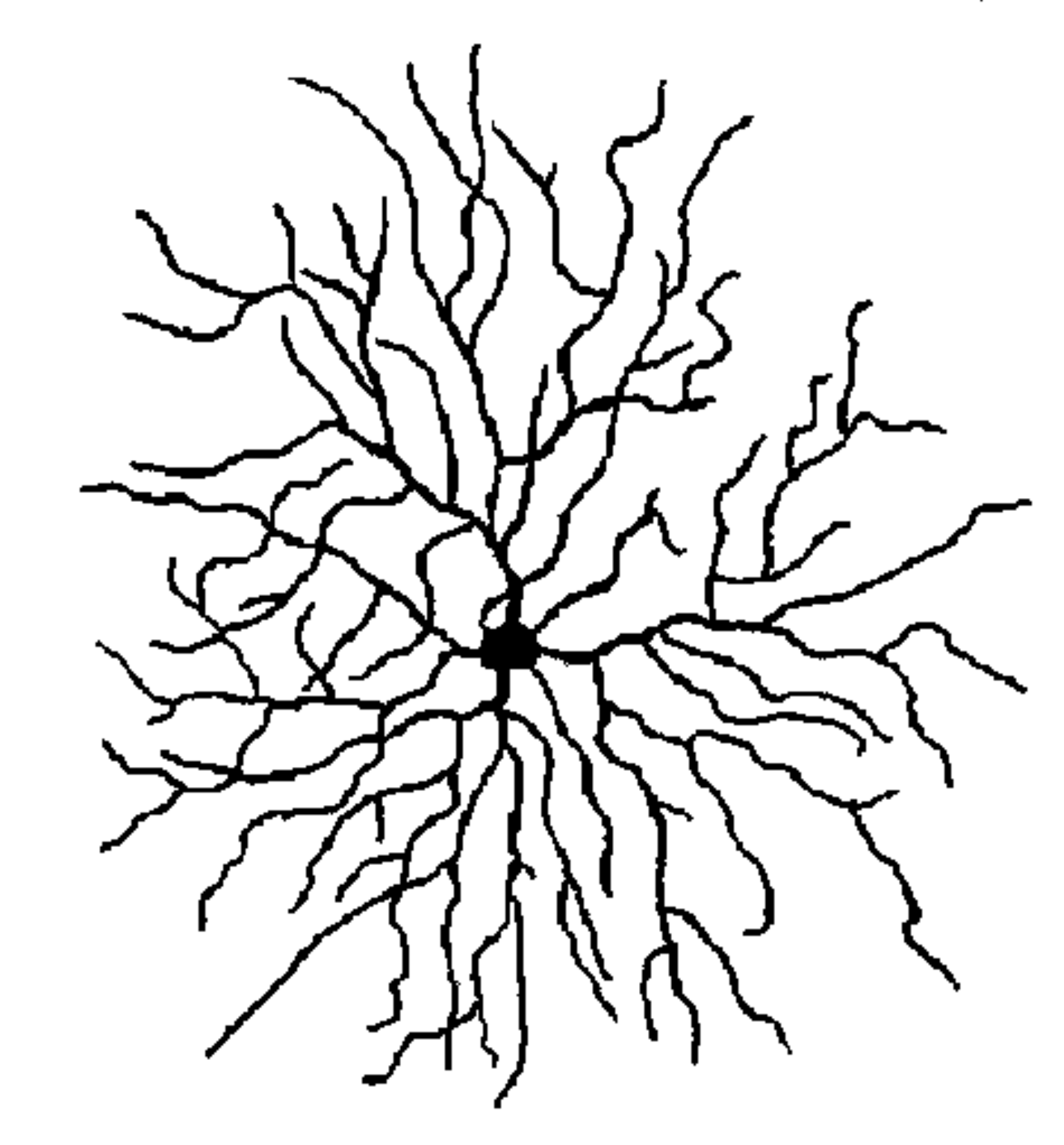}
\label{fig:delta_a}}
\hfil
\subfigure[]{\includegraphics[width=50mm]{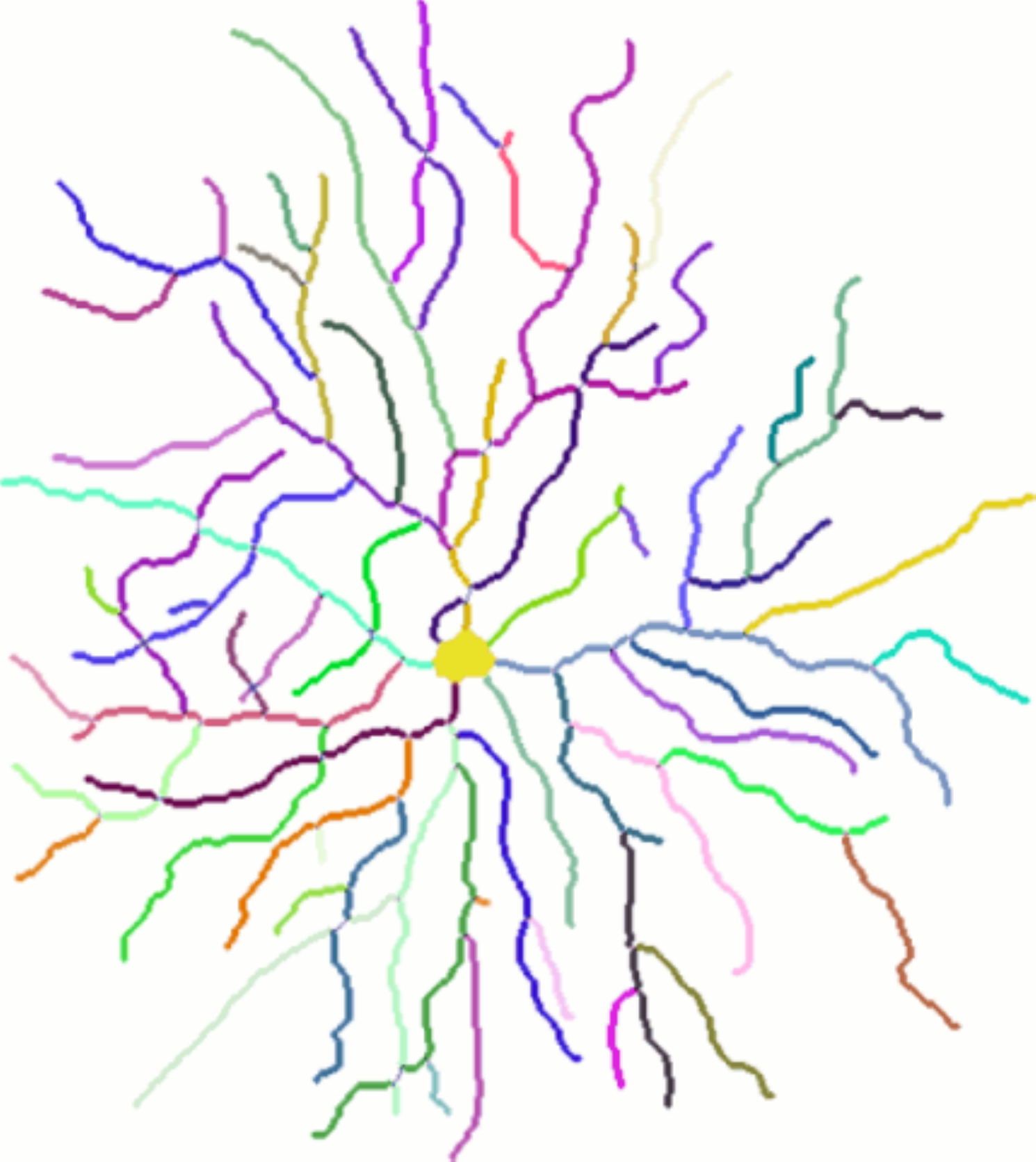}
\label{fig:delta_b}}}

\centering
\centerline{\subfigure[]{\includegraphics[width=50mm]{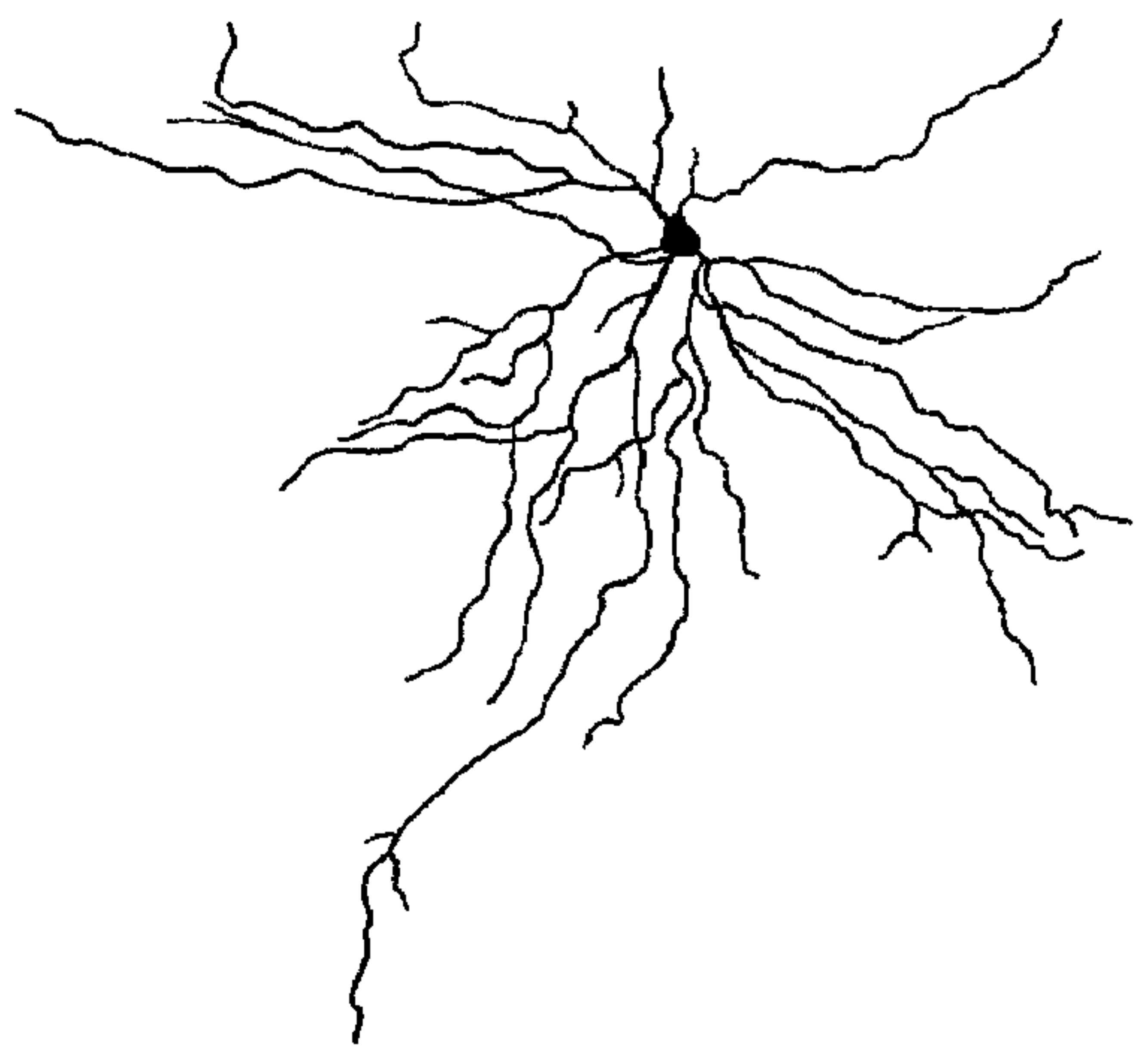}
\label{fig:epsilon_a}}
\hfil
\subfigure[]{\includegraphics[width=50mm]{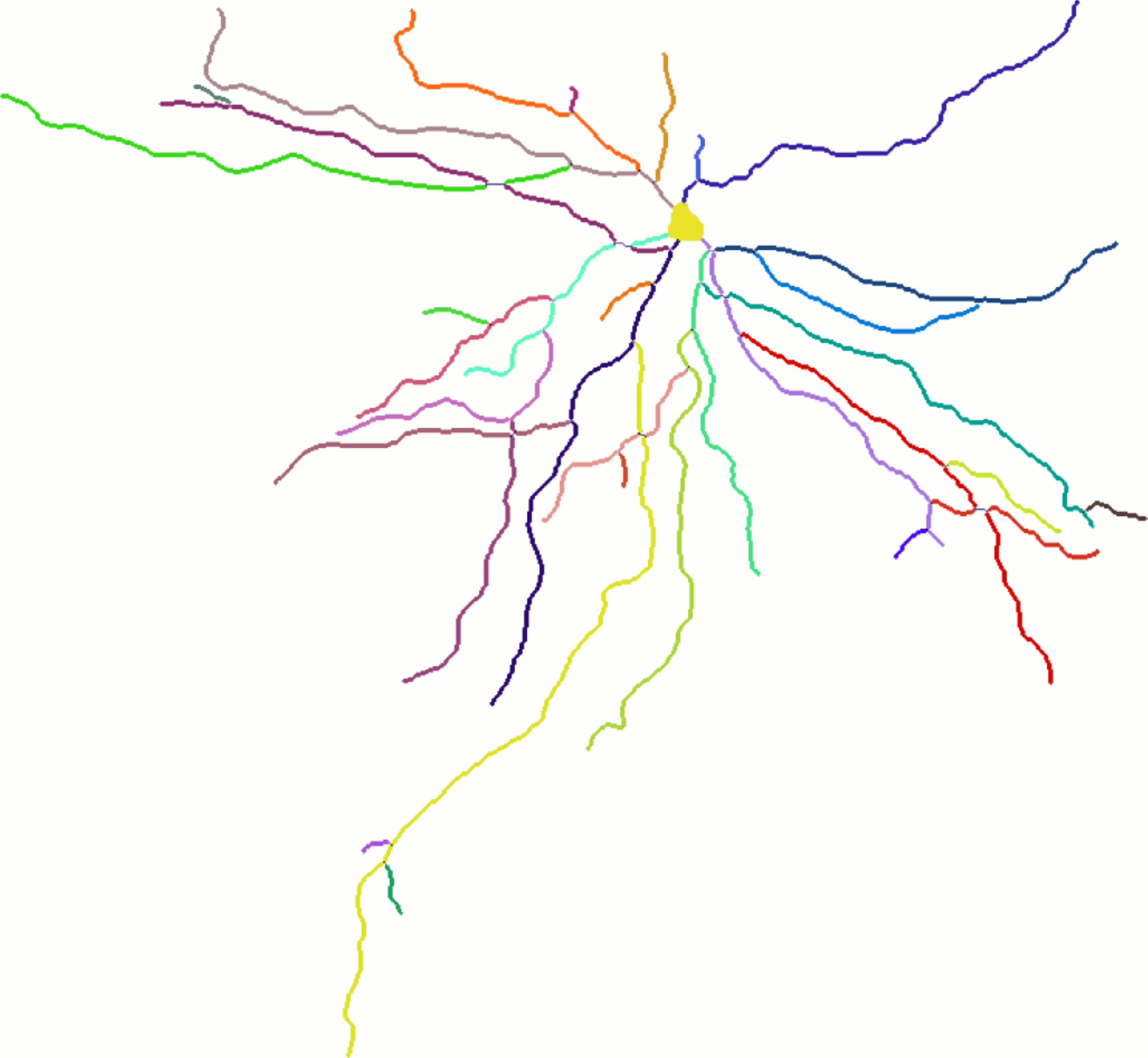}
\label{fig:epsilon_b}}}
\caption{Results obtained by the \emph{Branches Tracking Algorithm} when 
applied to alpha, delta and epsilon neuron images. 
(a) Alpha (c) Delta and (e) Epsilon neuron images and respective
labeled images in (b), (d) and (f). Distinct branches appear in
different colours.}
\label{fig:bta-results}
\end{figure*}

Results obtained for three images were chosen to be presented in this
work.  Figure \ref{fig:bta-results} shows labeled neuronal images
(right column) obtained from alpha, delta and epsilon (left column)
types of neurons.  New labels were assigned to dendrite segments
originating from branches. The algorithm is able to distinguish
between critical region classes, reflected by the correct assigned
labels for the outwards segments from such structures. Notice how the
cases of close parallelism imply the \emph{BTA} to label the clumped
segments as superposition regions. Moreover, long overlaps are also
labeled as superposition regions, as long as such an overlap is
smaller than $D_{max}$, which is the minimum path length allowed
between two critical regions. Extremely close bifurcations of type $1$
can be labeled as a bifurcation of type $4$.  All the bifurcations,
superpositions and crossings have been correctly labeled.

\begin{figure*}[htp]
\centering
\centerline{\subfigure[]{\includegraphics[width=50mm]{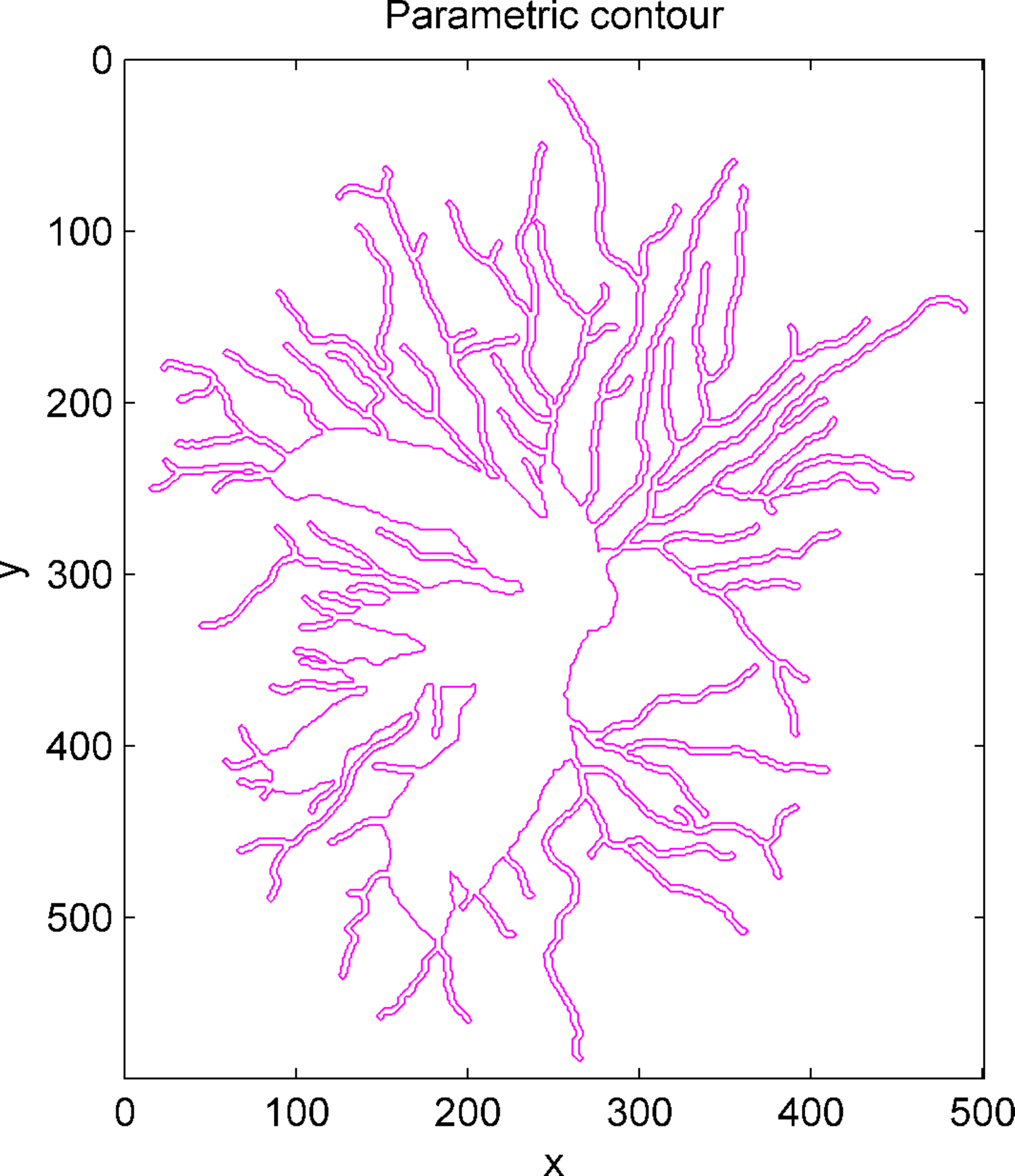}
\label{fig:comp_alfa_trad_cont}}
\hfil
\subfigure[]{\includegraphics[width=50mm]{alfa9-onlyparametriccontour}
\label{fig:comp_alfa_BSCEA_cont}}}

\centerline{\subfigure[]{\includegraphics[width=50mm]{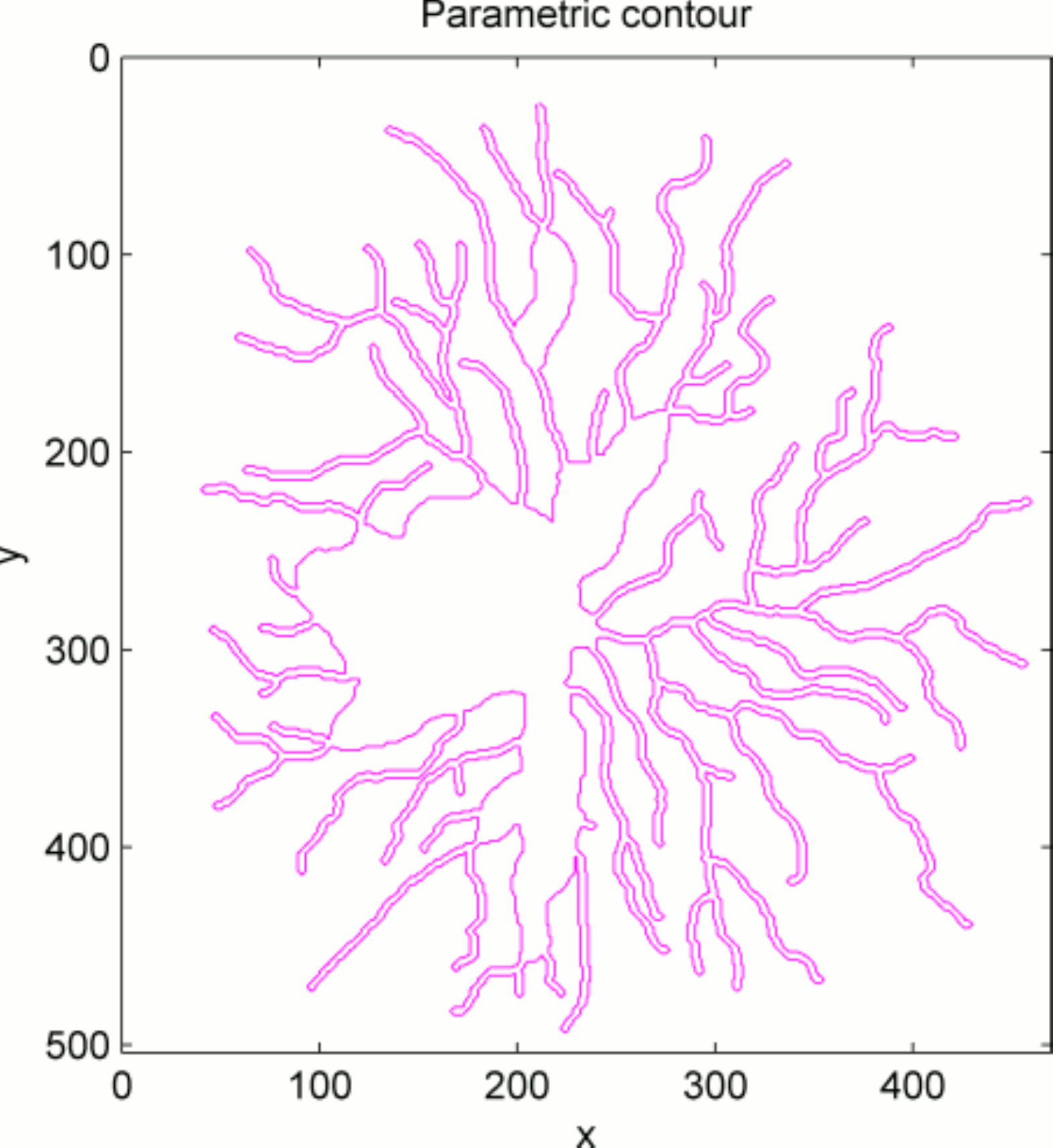}
\label{fig:comp_delta_trad_cont}}
\hfil
\subfigure[]{\includegraphics[width=50mm]{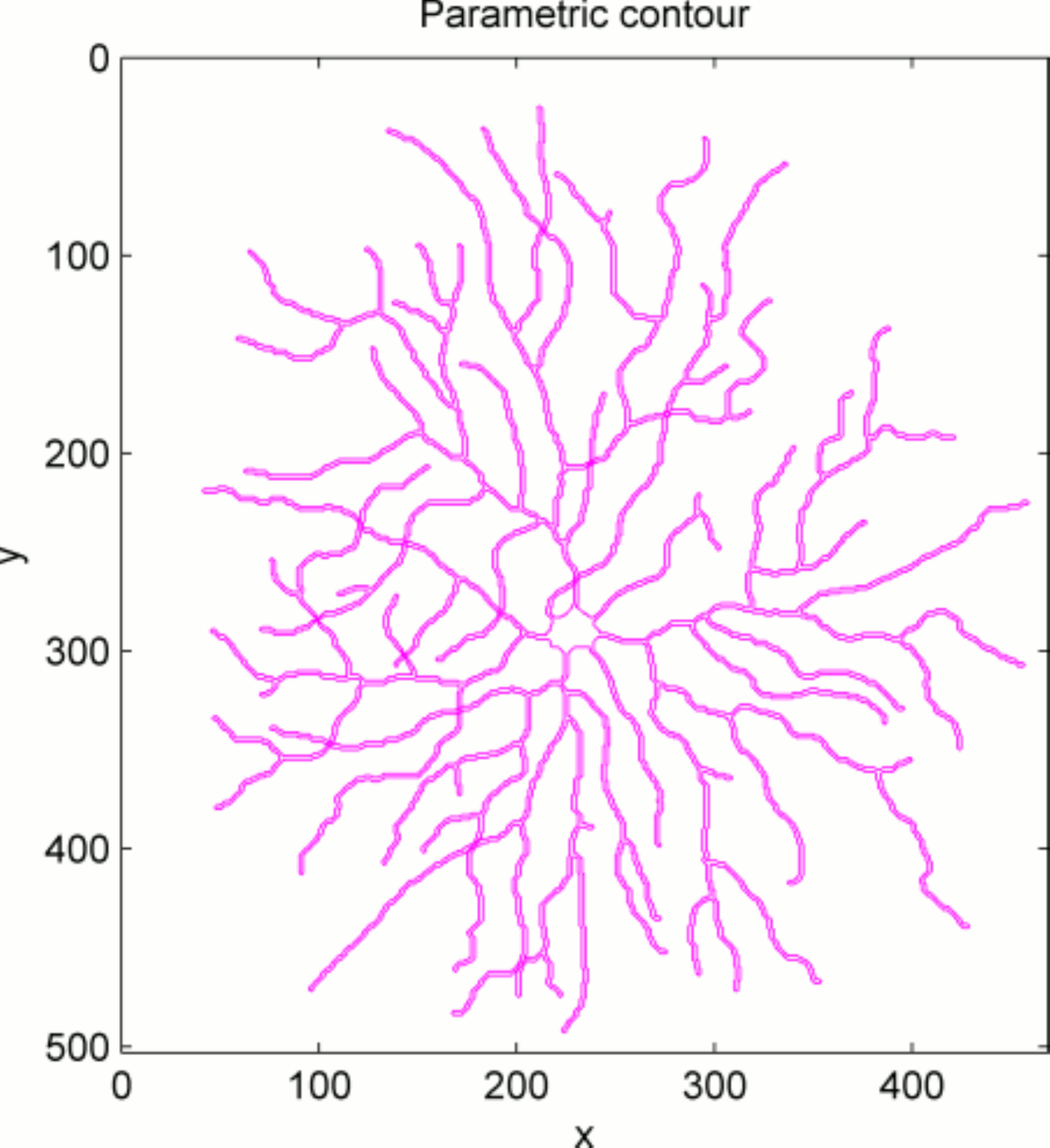}
\label{fig:comp_delta_BSCEA_cont}}}

\centerline{\subfigure[]{\includegraphics[width=50mm]{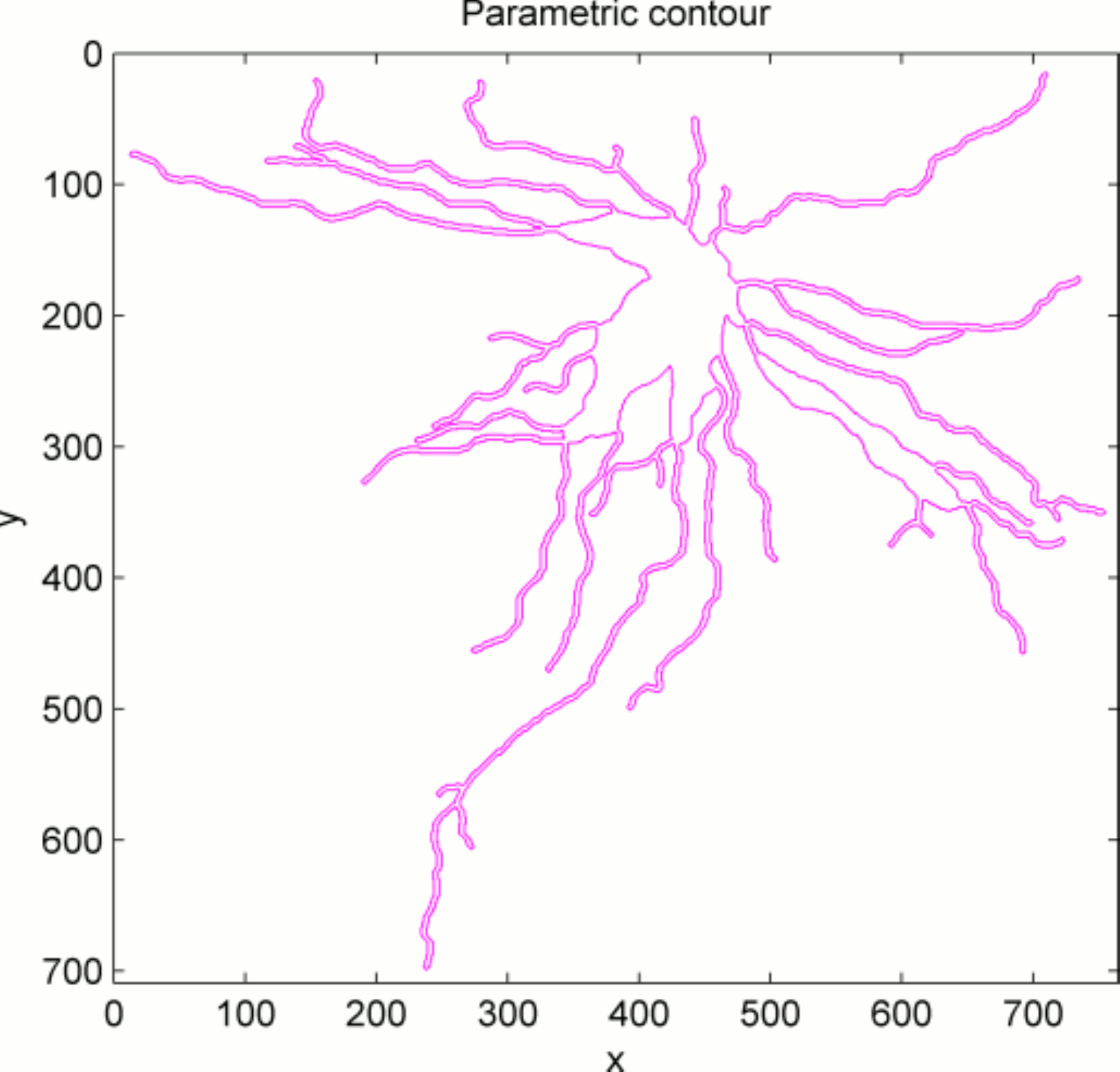}
\label{fig:comp_epsilon_trad_cont}}
\hfil
\subfigure[]{\includegraphics[width=50mm]{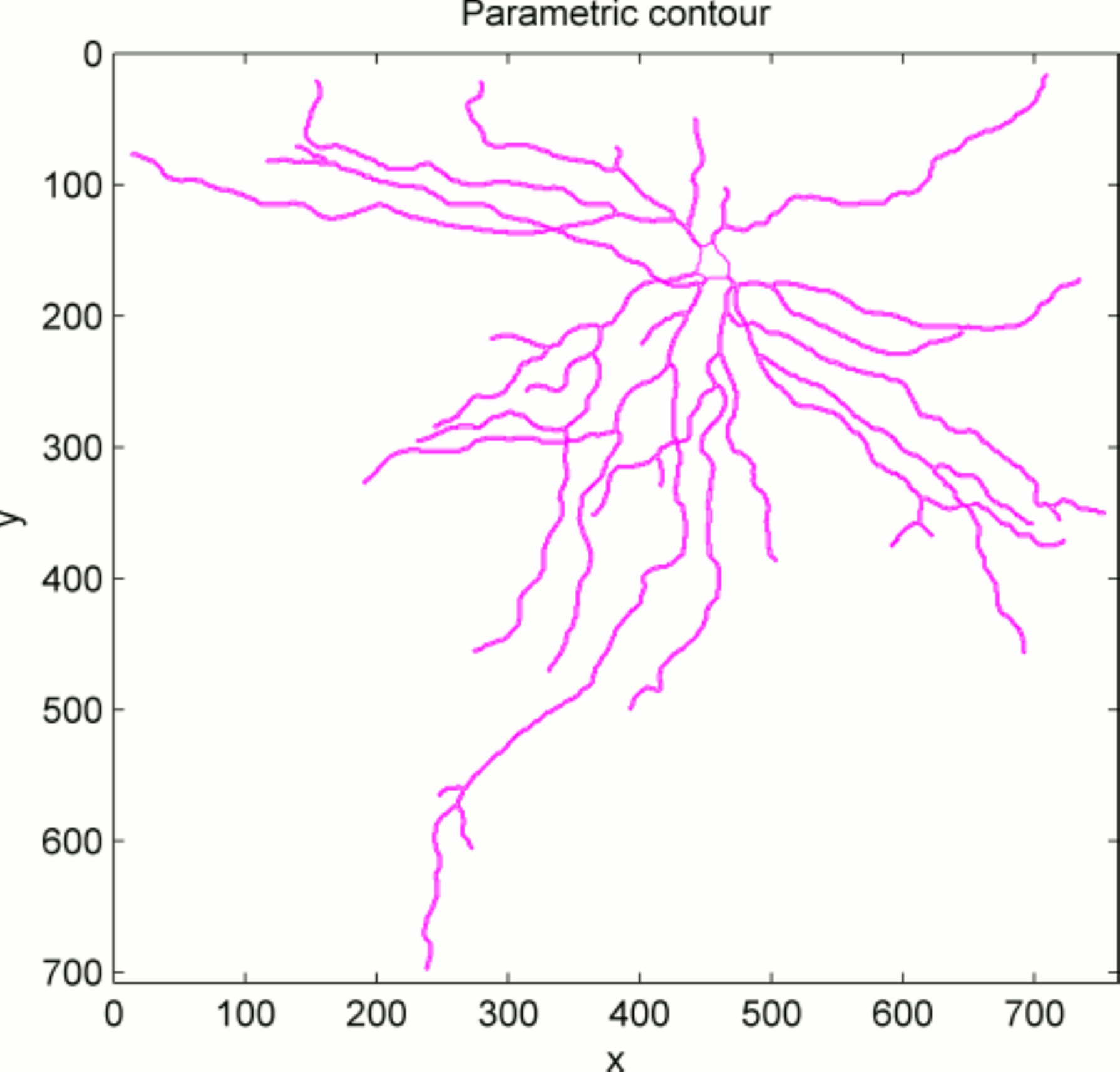}
\label{fig:comp_epsilon_BSCEA_cont}}}
     \caption{Comparison for alpha, delta and epsilon neuron images
     between results yielded by the traditional (a-c-e) and
     (b-d-f) BSCEA algorithms.  The \emph{BSCEA} enters all the regions,
     surpassing the traditional algorithm.}
     \label{fig:alfa-delta-epsilon-contorno-comparison}
\end{figure*}

Results for the \emph{Branching Structures Contour Extraction Algorithm} are
presented in Figure \ref{fig:alfa-delta-epsilon-contorno-comparison}, where one can see
the parametric contour trace for the shape and a comparison between
the results obtained by using both the traditional and the
\emph{BSCEA} approaches. Observe from Figures
\ref{fig:comp_alfa_trad_cont}, \ref{fig:comp_delta_trad_cont} and
\ref{fig:comp_epsilon_trad_cont} how the traditional algorithm did not
afford access to the innermost neuron contour portions, while the
\emph{BSCEA} conversely ensured full access to all neuronal
processes, as shown in Figures \ref{fig:comp_alfa_BSCEA_cont},
\ref{fig:comp_delta_BSCEA_cont} and \ref{fig:comp_epsilon_BSCEA_cont}.

\begin{figure*}[p]
  \centering
\includegraphics[height=180mm]{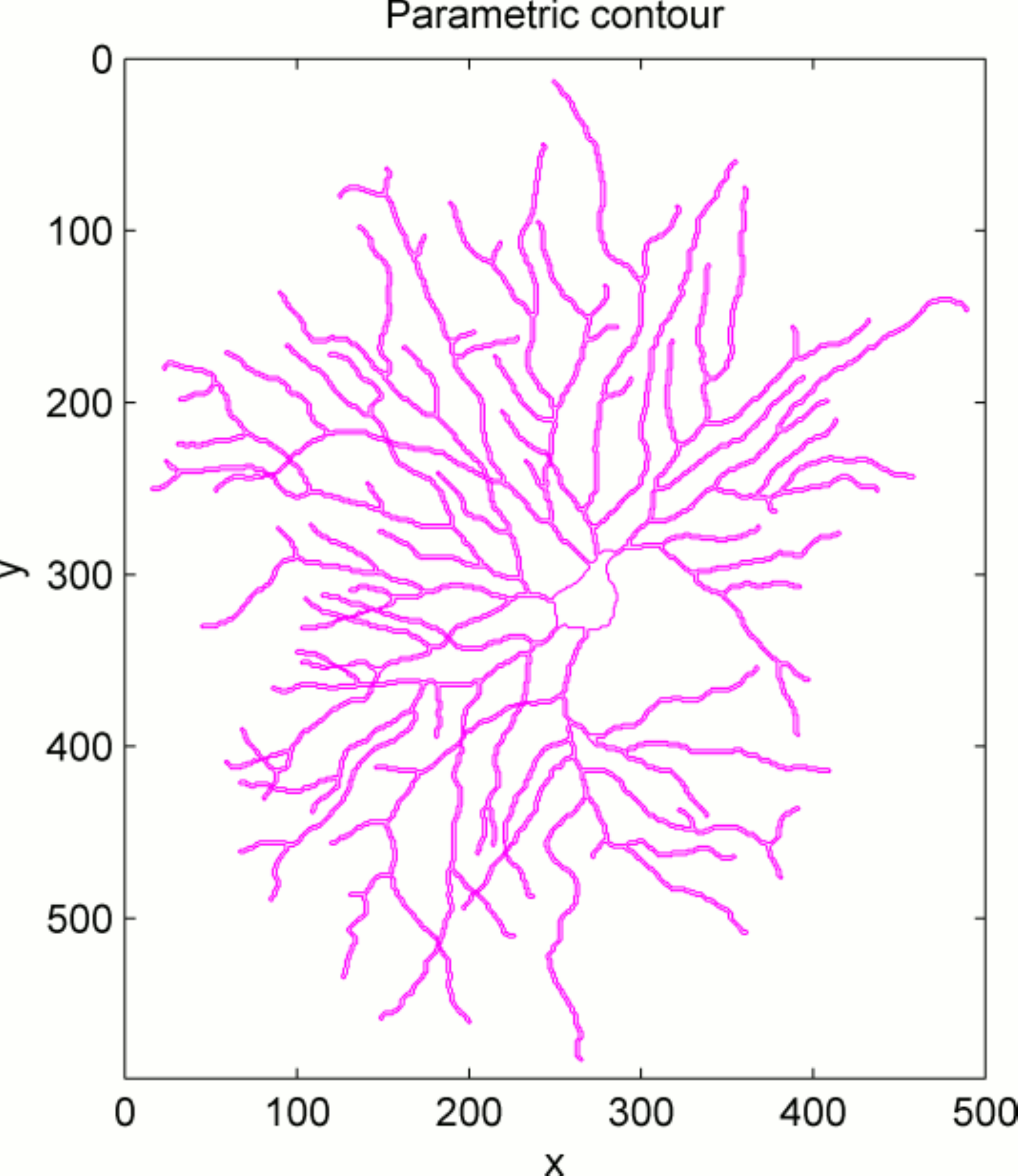}
    \caption{Parametric contour resulting from the \emph{BSCEA} application on a neuron \emph{Alfa} image. 
    Notice how the \emph{BSCEA} grants full admittance to all the innermost regions within the neuron image. 
    Also, note the contour continuity across critical regions.}
  \label{fig:results-contour-alfa}
\end{figure*}

\begin{figure*}[htbp]
\centering
 \centerline{\subfigure[]{\includegraphics[width=65mm]{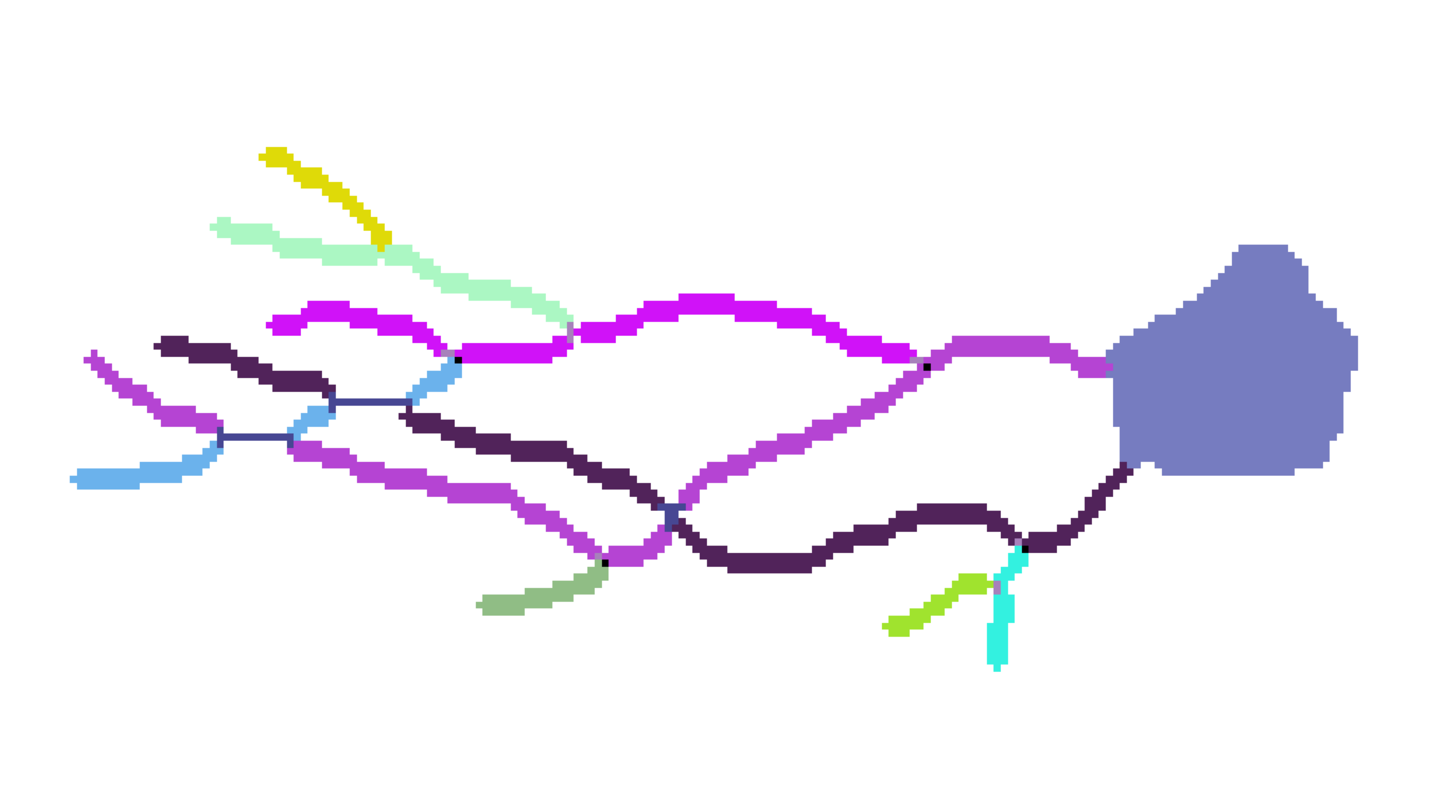}
\label{fig:auto-lbl}}
\hfil
\subfigure[]{\includegraphics[width=65mm]{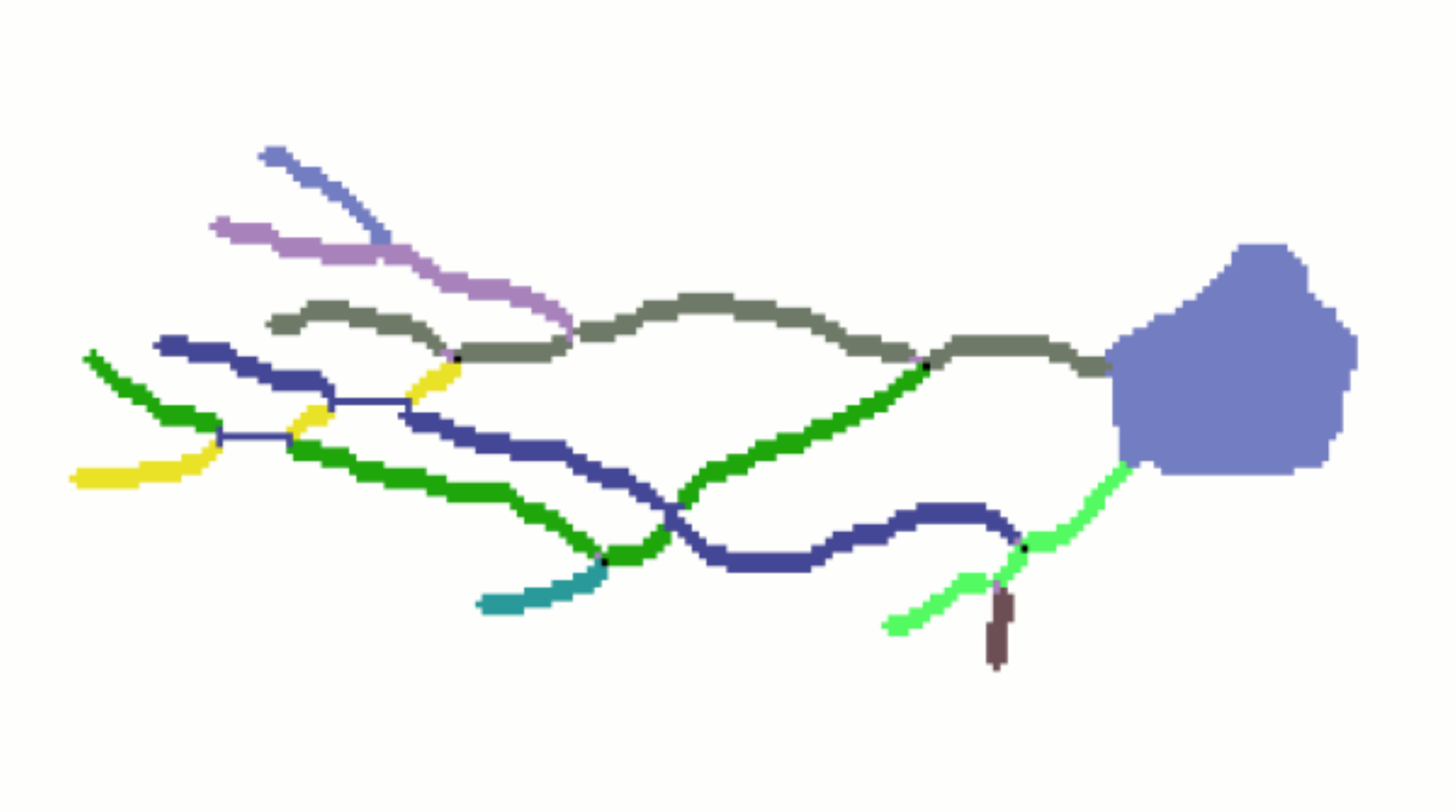}
\label{fig:man-lbl}}}
\centerline{\subfigure[]{\includegraphics[width=65mm]{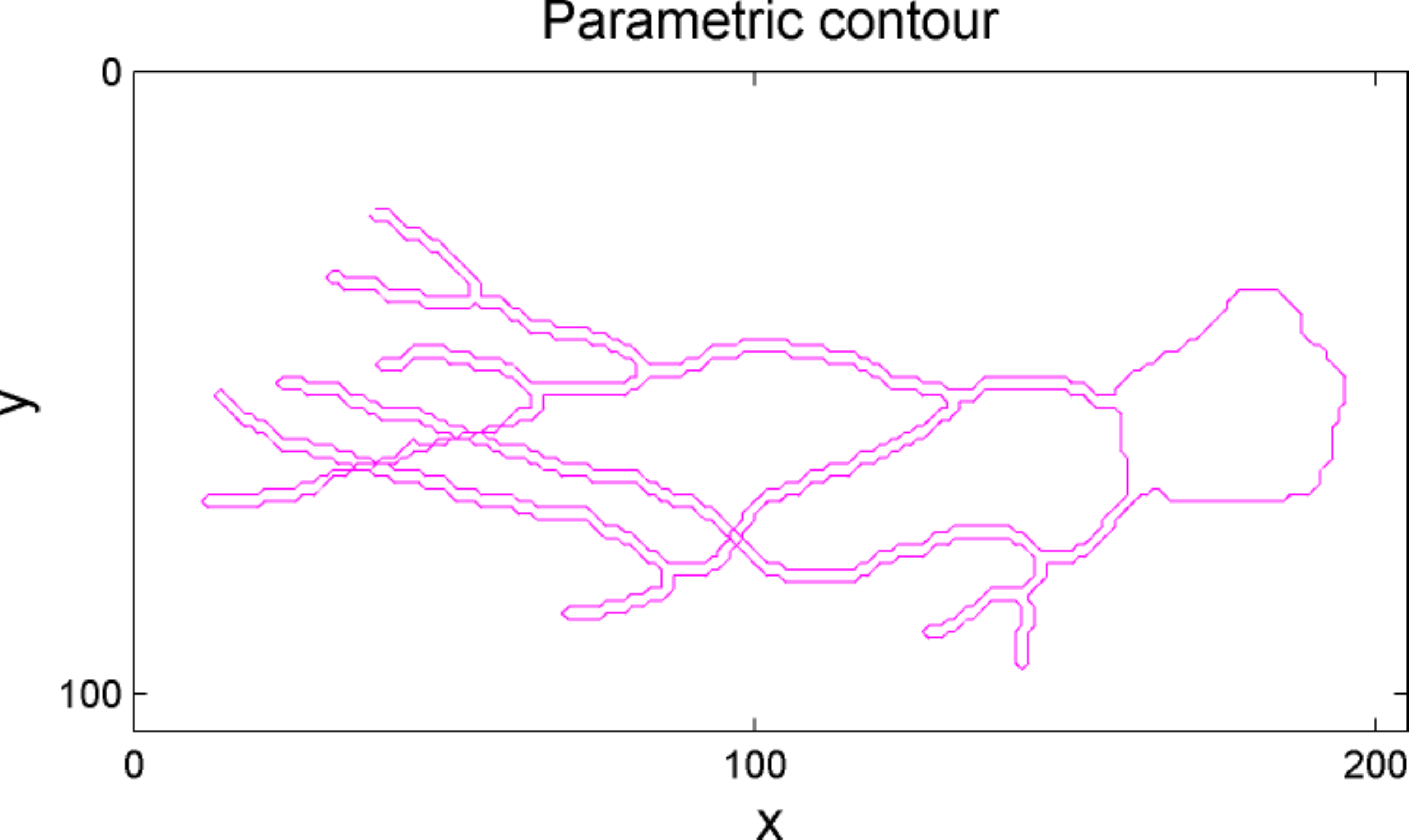}
\label{fig:auto-cont}}
\hfil
\subfigure[]{\includegraphics[width=65mm]{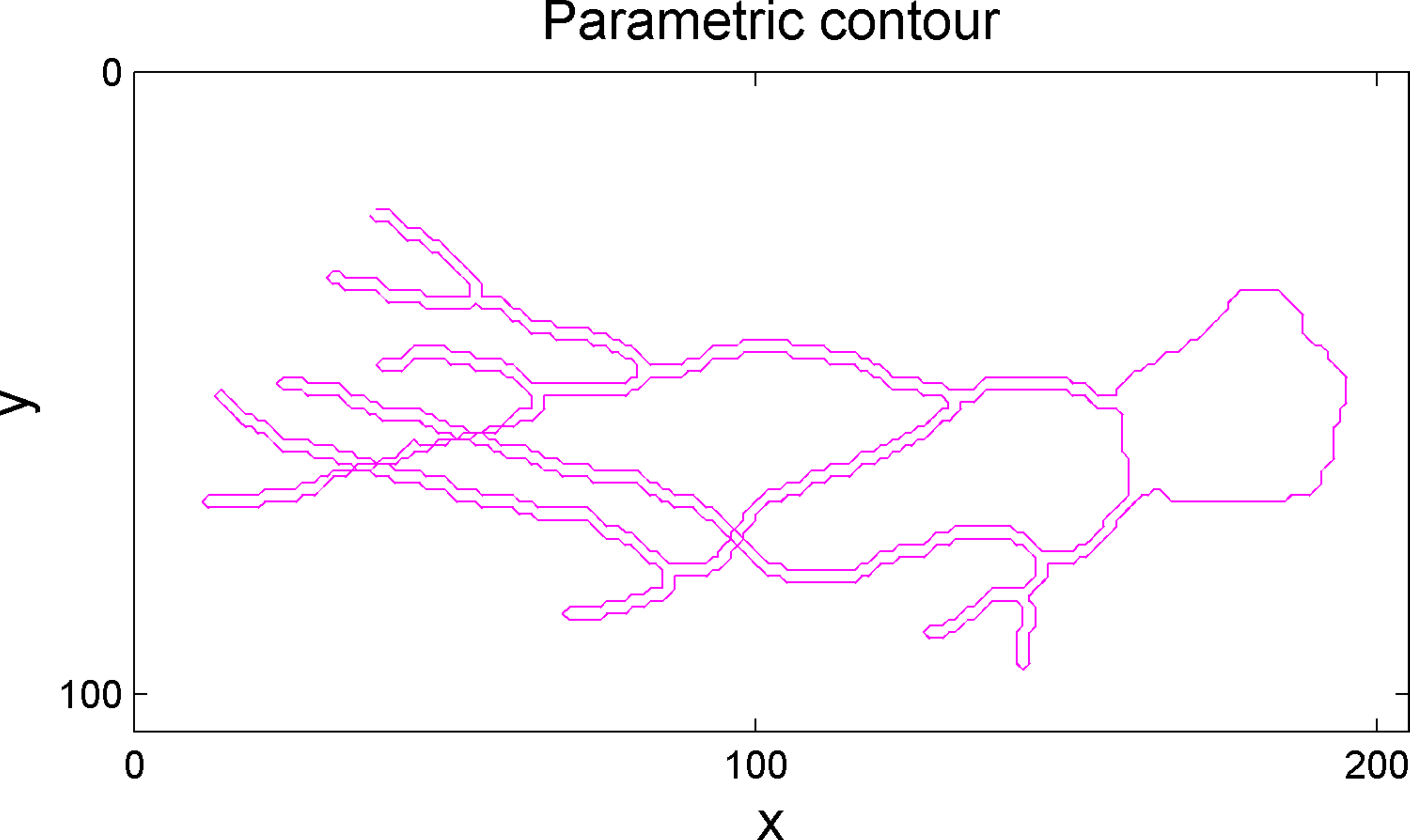}
\label{fig:man-cont}}}
\caption{Comparison between a synthetic neuron shape labeled automatically (a) and manually (b).
Parametric contours provided by the BSCEA for the automatic (c) and manual (d) labelings. 
Despite minor differences between automatic and manual labelings, notice that both contours are in complete accordance.}     
\label{fig:autoversusmanual}
\end{figure*}

\begin{figure*}[htbp]
\centering
\centerline{\subfigure[]{\includegraphics[width=65mm]{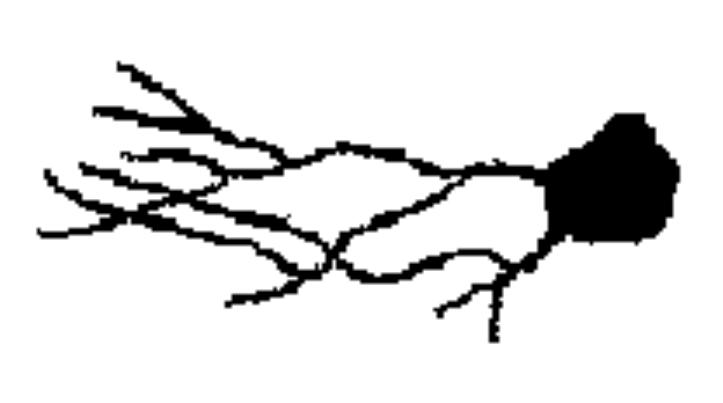}
\label{fig:alfa-synthetic}}
\hfil
\subfigure[]{\includegraphics[width=65mm]{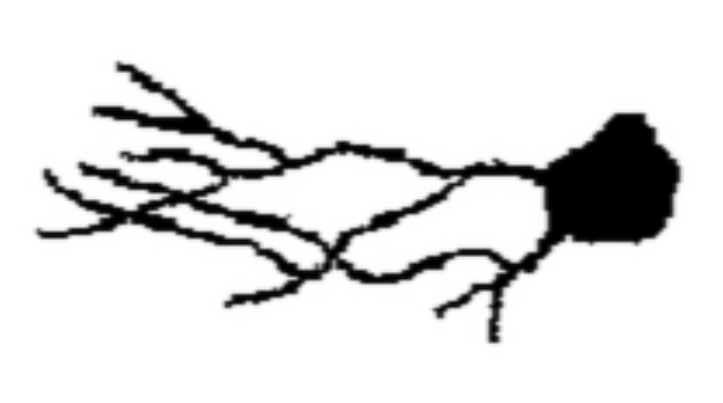}
\label{fig:2e-6}}}
\centerline{\subfigure[]{\includegraphics[width=65mm]{alfasynthetic-labeled}
\label{fig:alfa-synthetic-labeled}}
\hfil
\subfigure[]{\includegraphics[width=65mm]{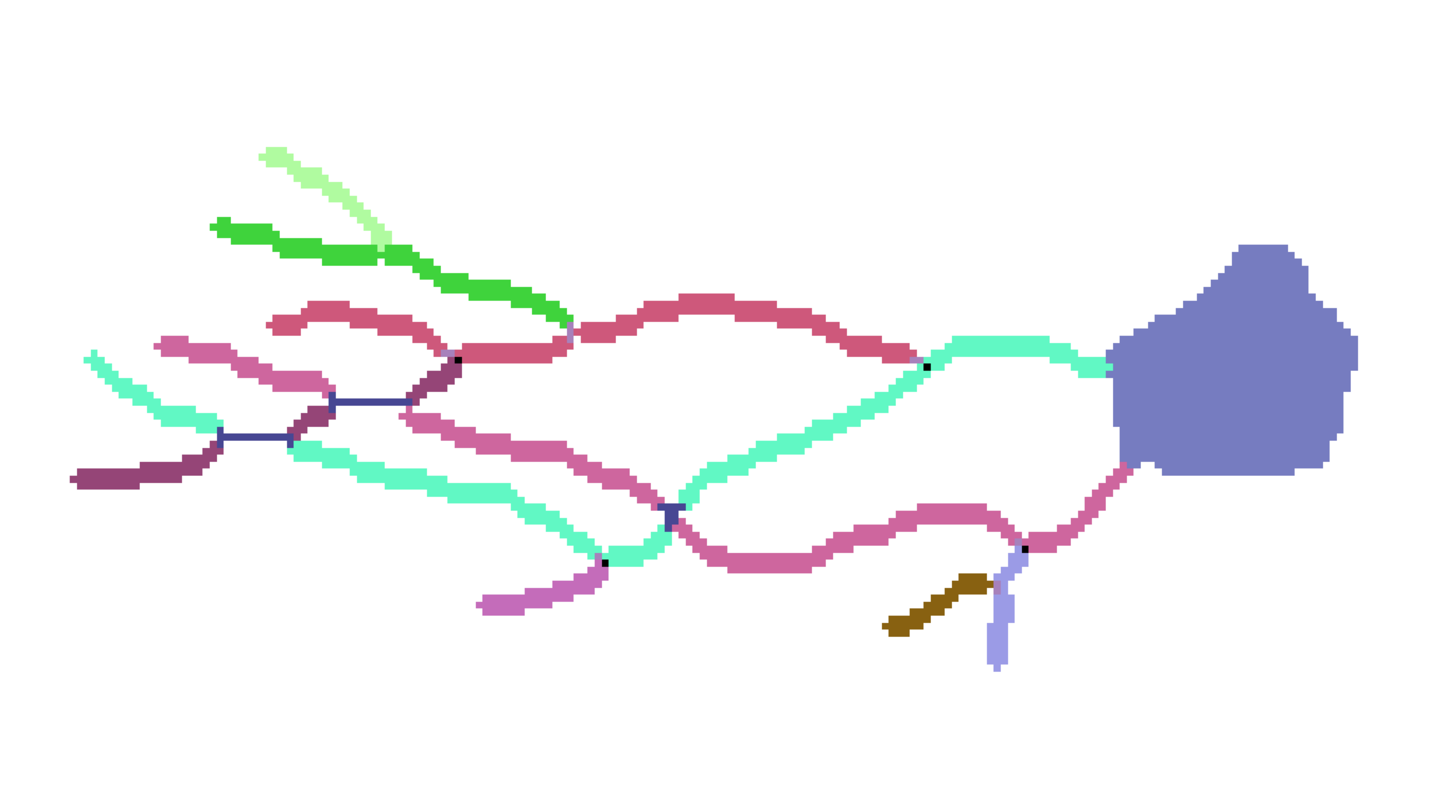}
\label{fig:2e-6-thr111-lbl}}}
\centerline{\subfigure[]{\includegraphics[width=65mm]{alfasynthetic-onlyparametriccontour}
\label{fig:alfa-synthetic-contour}}
\hfil
\subfigure[]{\includegraphics[width=65mm]{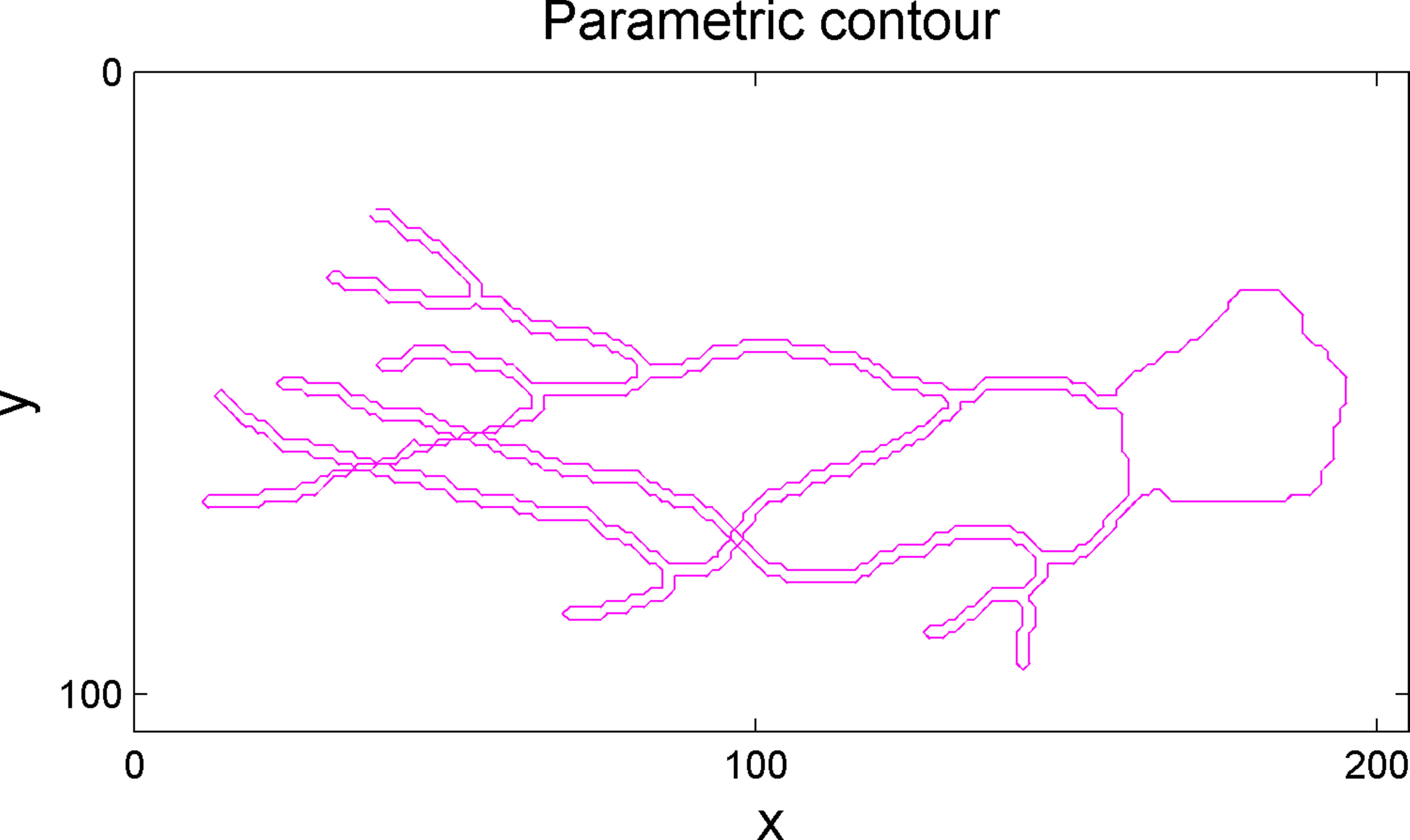}
\label{fig:2e-6-cont}}}
  \caption{Assessment of noise effects on neuron shape labeling and
  respective contour extraction.  A synthetic neuron shape has been
  corrupted by $2D$ Gaussians with different bandwidths.  The Gaussian
  scale parameters $\sigma$ in Fourier domain varied from $10^{-8}$ up
  to $2\cdot10^{-6}$. Both BTA and BSCEA ensured robustness
  within this range of smoothing bandwidths.}
\label{fig:noiseeffects2}
\end{figure*}

Additional experiments have been carried out in order to validate the
algorithms and to assess their performance with respect to noisy
neuronal images. A validation test was performed with a synthetic
neuron image, by labeling it manually and automatically through the
\emph{BTA} (Fig.~\ref{fig:autoversusmanual}). Despite differences
between manual and automatic labelings, due to distinct assessments of
tangent continuity for some bifurcations in both approaches, notice
that both labelings are consistent, providing suitable input for the
\emph{BSCEA} which yielded identical parametric contours for both
labelings. The robustness of the proposed methodology for noisy
versions of the same synthetic neuron image has been tested by
convolving the original synthetic image with six different $2D$
Gaussians, using bandwidths parametrized by values of standard
deviations spanning from $10^{-8}$ up to $2\cdot10^{-6}$ in the
Fourier domain. Figure~\ref{fig:noiseeffects2} shows results for
extreme cases, namely the original synthetic image and the smoothest
version of it. Notice that the \emph{BTA} provided consistent
labelings for the original and the smoothed versions, and the
\emph{BSCEA} yielded identical parametric contours for both cases.

\begin{figure*}[htp]
\centering
\centerline{\subfigure[]{\includegraphics[width=80mm]{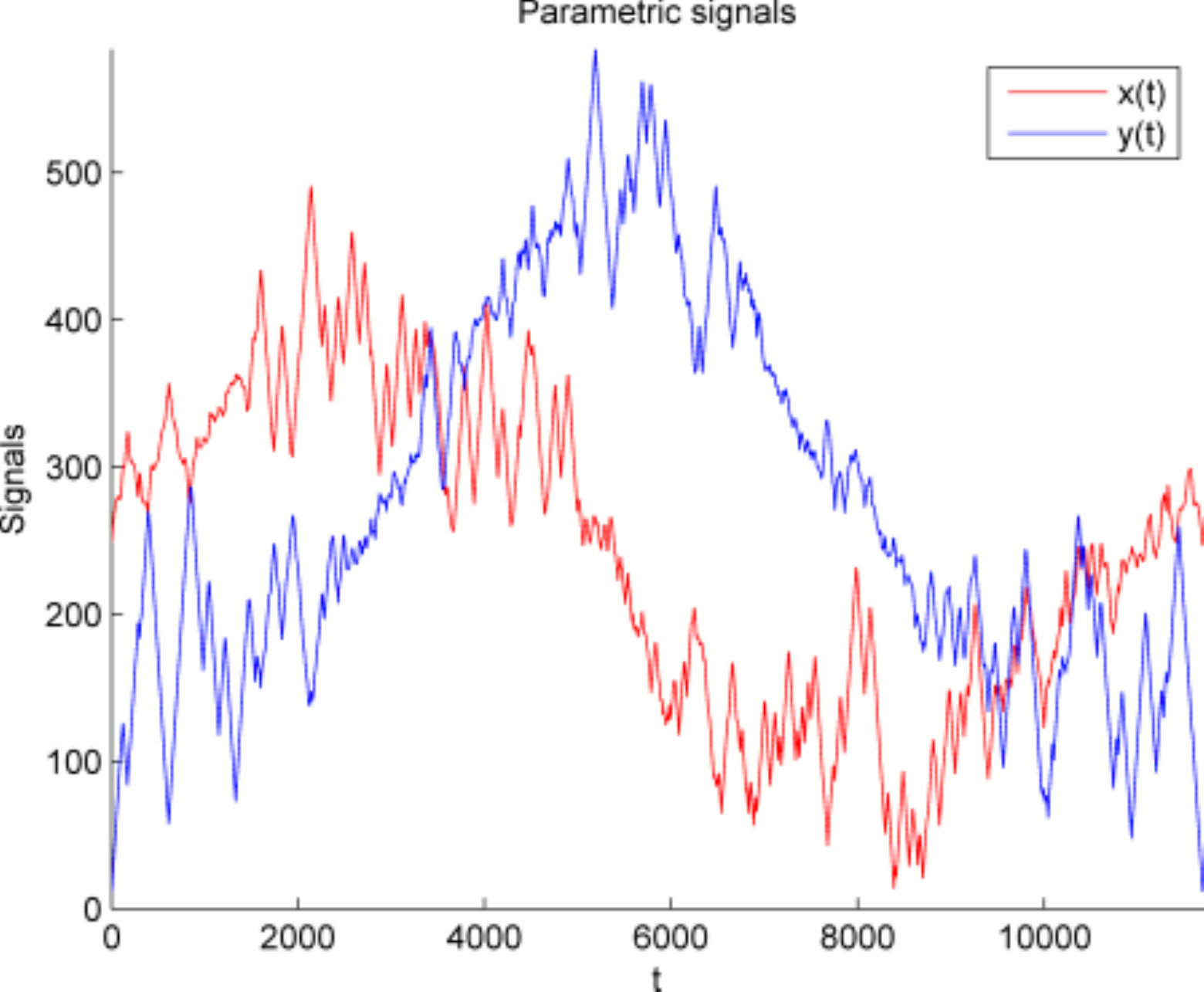}
\label{fig:comp_alfa_trad_sig}}
\hfil
\subfigure[]{\includegraphics[width=80mm]{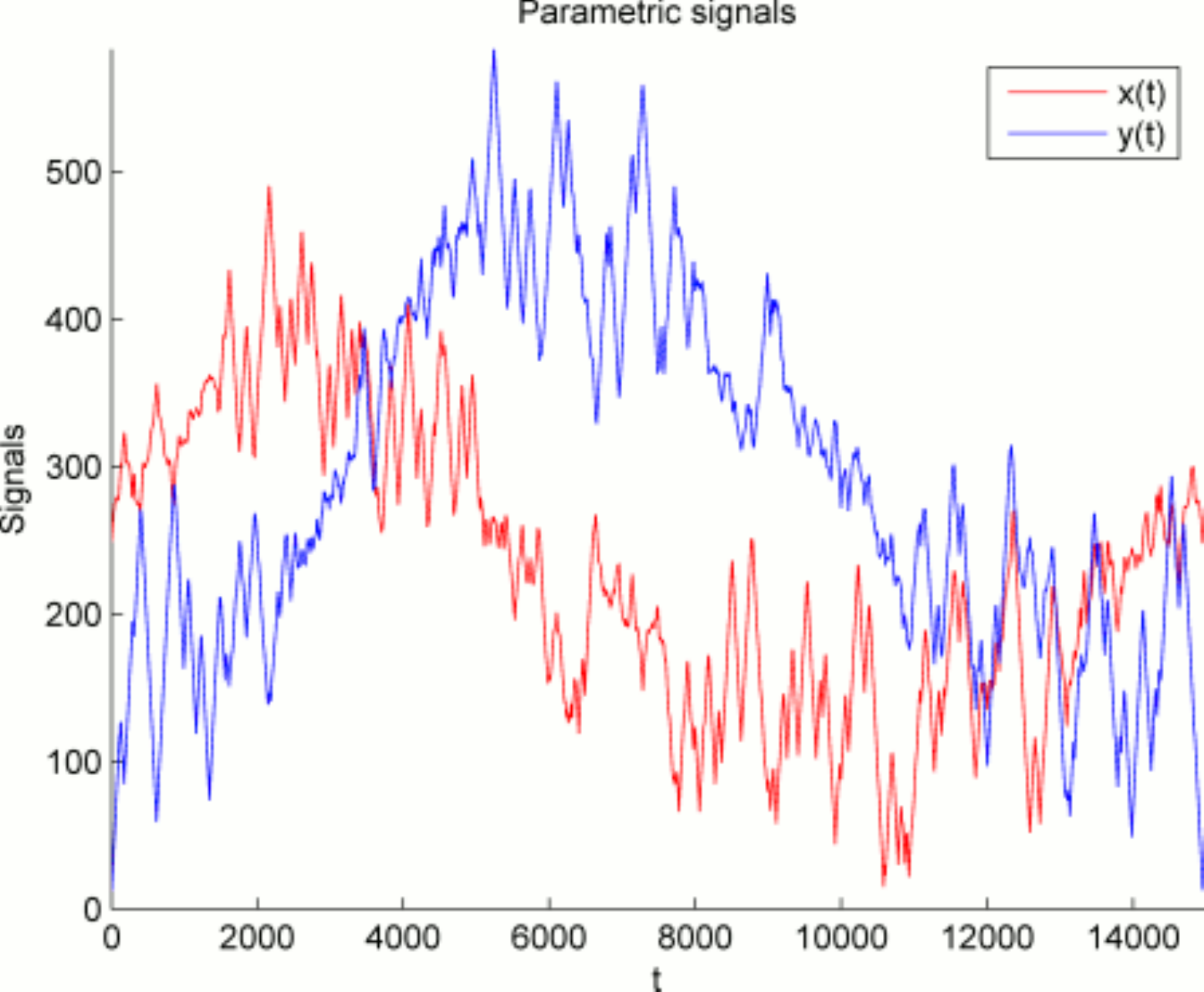}
\label{fig:comp_alfa_BSCEA_sig}}}

\centerline{\subfigure[]{\includegraphics[width=80mm]{alfa9-onlytradparametriccontour}
\label{fig:comp_alfa_trad_cont_}}
\hfil
\subfigure[]{\includegraphics[width=80mm]{alfa9-onlyparametriccontour}
\label{fig:comp_alfa_BSCEA_cont_}}}
     \caption{Comparison for an \emph{alpha} neuron image between results yielded by the algorithms (a-c) Traditional  and (b-d) BSCEA.
     The \emph{BSCEA} enters all the regions, surpassing the traditional algorithm in
     over than two thousand contour points.}
     \label{fig:alfa-contorno-comparison}
\end{figure*}

\section{Concluding Remarks}
\label{sec-concluding}

The proper shape characterization of branching structures is a
particularly important problem, as it plays a central role in several
areas of medicine and biology, especially in neuroscience.  Indeed,
the current understanding of the physiological dynamics in biological
neuronal networks can be reinforced through the proper
characterization of neuronal cells shapes, since both the amount of
synapes and the way in which neurons organize in networks are strongly
related to the cells shapes.

Local information is retrieved through parametric contour analysis, while
global information can be acquired from statistical measures considering the shape of the whole cells. 
However, accurate contour following in neuronal shapes has not been possible so far 
because of the presence of overlappings among neuronal processes implied by 
the projection from $3D$ neuronal shapes onto $2D$ images. Whenever a crossing takes place
in these images, the traditional contour following algorithm, which is based on the chain-code, fails to enter 
their innermost regions. The present work described an original methodology capable of properly tackling the problem of following the contour of branching structures, 
even in the presence of intercepting branches. The main original contributions of our
method\footnote{Preliminary results of the proposed approach have been
described in conferences (\cite{ leandrobranching,
leandrobioimage:2008}).} include both the tracking of the branching
structures, such as neurons, as well as the extraction of the
respective parametric contours. In addition, the features adopted in
this work to classify critical regions -- such as shape, number of
adjacent segments and angles among segments, have intrinsic potential
for providing additional information to be used in neuronal
characterization and classification. Our system is basicaly comprised of three parts, i.e.:
(i) \emph{Preprocessing Algorithm}, (ii) \emph{Branches Tracking Algorithm - BTA} and (iii) \emph{Branching Structures Contour Extraction Algorithm - BSCEA}.

Because the proposed system begins with a series of transformations (preprocessing) on the
$2D$ projection of a $3D$ branching structure image, so as to obtain a
suitable skeleton, obviously any skeletonization
scheme other than the morphological thinning might be adopted, such as
exact dilations~\cite{costabook01}, medial axis transform, and so on,
provided that an $8$-connected skeleton with one-pixel wide branches is obtained as a
result. Besides, the skeletonization scheme will affect the
choice of all the preprocessing parameters, which in this work have
been picked out by trial and error. One should bear in mind that the
method gist is supplying the tracking algorithms with an adequate
skeleton as input.

Apart from the skeleton, there are a number of separate components obtained through
the preprocessing step. Although the data structure to store such separate components  
is immaterial to the preprocessing pipeline implementation, since it may 
be implemented in alternative ways, we particularly have them separated 
and stored into separate images, for the sake of easier 
implementation. As will be shown in the sequel, the system dynamics  
involves comparisons among such pieces of information, which may be 
properly achieved by means of set operations (union, intersection, 
set difference, etc). Considering that mathematical morphology operations 
are usually described in terms of set operations (union, intersection, etc.) 
and that the \emph{SDC Morphology Toolbox for MatLab}\TTra\ \cite{mmorph:2006} 
implements such operations, we decided to take advantage of it 
by separating those structures into different images. However, instead 
of separating structures in images, one could label a pixel pertaining 
to a structure, by using the Object-Oriented Programming Paradigm to keep such information 
in the respective pixel attribute. 
Moreover, it should be emphasized that the one-pixel-wide 
restriction on the branches of the skeleton is mandatory in 
order to guarantee the proper operation of the tracking algorithm. In general, 
the dendritic tree should be one-pixel wide, because the tracking procedure 
is based on the stacking of nearby valid (background, unlabeled) pixels, 
following the chain code order. Hence, 
the thicker the structure, the larger the number of pixels in the 
vicinity of the current pixel. These irrelevant pixels would be indistinctly 
staked, putting the BTA in a forward-backward visitation of pixels, and not 
in a sequence. One should bear in mind that a one-pixel wide skeleton 
gathers enough information concerning the essential structure 
of the shape~\cite{costabook01}. 

Notice that the structuring elements dimensions have been obtained in an
empirical basis, specially for the images used in this work. Three images
have been used to calibrate these parameters: an \emph{alpha} neuron image of size $500\times598$, a \emph{delta} neuron image
of size $475\times511$ and an \emph{epsilon} neuron image of size $768\times712$. It should be
emphasized that images with very different sizes may require different structuring elements. 
Considering that Mathematical Morphology Image Processing depends on the structuring 
element sizes and shapes, choosing 
a suitable structuring element is important. 
Therefore, for different sized images from the mentioned above, 
one will have to test disks with different diameters, following the sequence of operations 
described.

The algorithms robustness regarding the contrast in input images is not 
affected, since the preprocessing first step is to binarize the input image. Also, 
for the purposes of this methodology, noise is related to redundant pixels, such as 
those wiped out through the hit-or-miss filtering operation. 

As for the \emph{BTA}, 
there may be particular cases for further consideration yet, for example images
with high density values of critical regions and/or the presence of 
structures whose topologies might favour the appearance of superpositions. 
The first case, i.e. high critical regions densities may be due to
particular shape topologies in the image or due to the image
resolution itself, causing the \emph{BTA} to cluster critical regions ocurring very close to one another.
Notice that, in an effort to fulfil the previously set stop condition for the Breadth-First
Search, the \emph{BTA} has bunched both bifurcations of type $1$
(Fig. \ref{fig:clumping}-(a)) into a cluster of bifurcations appearing
as a bifurcation of type $4$ (Fig. \ref{fig:clumping}-(b)).  A
possible solution is to use 
breadth-first search implemented with a tree data structure,
in addition to the auxiliary queue, in order to properly maintain memory of valid
paths between direction vector terminations and origins.
Immediately after achieving the stop condition, the \emph{BTA}
would retrieve the direction vectors end points hosted at the
tree leaves. The respective direction vectors origins would
simply be obtained by climbing the tree from the leaves upwards,
until reaching the respectives first non-critical pixels adjacent to
the precedent critical region. Moreover, this strategy might allow disambiguating 
between two possible origins that are equally far apart 
from a very same direction vector termination, since the
shortest distance between a direction vector termination and an origin candidate is
the only condition taken for granted. The auxiliary tree data structure would add a piece of information
regarding the existence of a valid path (segment) between a direction vector termination
and an origin candidate.

In the second case, that is Superpositions, one should notice that 
such regions may be resultant from three
distinct problems.  Firstly the hugely close parallelism, in which images 
containing many branches almost
parallel and very close to each other usually come to present short cycles
after the preprocessing stage. Short cycles are highly undesirable,
since they are error-prone structures. This problem may be
circumnvented by including a dilation step in the preprocessing
pipeline, just before the skeletonization.  In so doing, short cycles
are shrinked into closed Superposition regions.  Secondly, poor
resolution may yield images having ellipsoid cigarshaped
crossings sampled as almost line-shaped structures.  After the
preprocessing, these structures will apear as Superposition
regions. Lastly, long overlaps will also be labeled as superposition regions, 
provided that such overlapping is below the threshold $D_{max}$, that is 
the shortest allowed path length between two bifurcations.

Needless to say that the system performance should not be evaluated on the basis 
of the number of images, but rather on the number of fully accomplished tasks. 
Regarding the number 
of successful objects processed, the number of trespassed critical regions, 
during the contour following process, is a quantitative assessment \emph{per se}.
Notice that the parametric contour following process is a sequential procedure, 
in which any mistake will prevent 
the algorithms to continue and reach their goal. In the case of BTA, the 
task is to label every branch. In the case of BSCEA, the task is to obtain the parametric 
closed contour, coming back to the first pixel it started from. Having this in mind, 
BTA succeeds only and if only if it does not leave any branch out without 
being labeled. So the more labeled branches within a neuron, the more evident the BTA success.
As for the BSCEA, it succeeds only and if only if it could return to the first pixel it 
started from, after having the whole pattern contoured. So, the more critical regions within a $2D$ neuron 
image, the more evident the BSCEA success. Also, concerning the quantitative validation 
between results obtained from the traditional and the BSCEA approaches, 
Fig.~\ref{fig:alfa-contorno-comparison} displays the parametric signals 
($x(t)$ and $y(t)$) for both cases. Several global features may be calculated from these 
signals, such as the bending  energy. 

The most expensive operation in the BTA would be to check every pixel at some $8$-neighborhood 
to decide whether or not it should be labeled. However this is done at most a 
constant number of times. So, tracking would be eventually of $O(n)$ 
with respect to the number of object pixels (far less than the size of the image). Similarly, 
in BSCEA, every pixel in the neighborhood of a labeled pixel is visited to check whether it 
has a blank neighbor which will ultimately become a contour pixel, so it would also be of $O(n)$.

The main original contributions of the present work\footnote{Preliminary results of the proposed
approach have been described in conferences (\cite{ leandrobranching, leandrobioimage:2008}).} encompass
both the tracking and the parametric contour extraction from
branching structures, like neuron cells. Future developments
include the extension of the methodology to separate cells in images
containing multiple cells.  Several applications of the methodology
proposed in this work can be made regarding neural networks images
as well as other types of biological structures such as retinal vessel
trees.

\newpage

\appendix
\label{sec-appendices}

\section{Appendices}
\subsection{Breadth-First Search Example}
\label{sec-bta-bfs-examples}

\begin{figure}[htb]
\centering
 \centerline{\subfigure[]{\includegraphics[width=60mm]{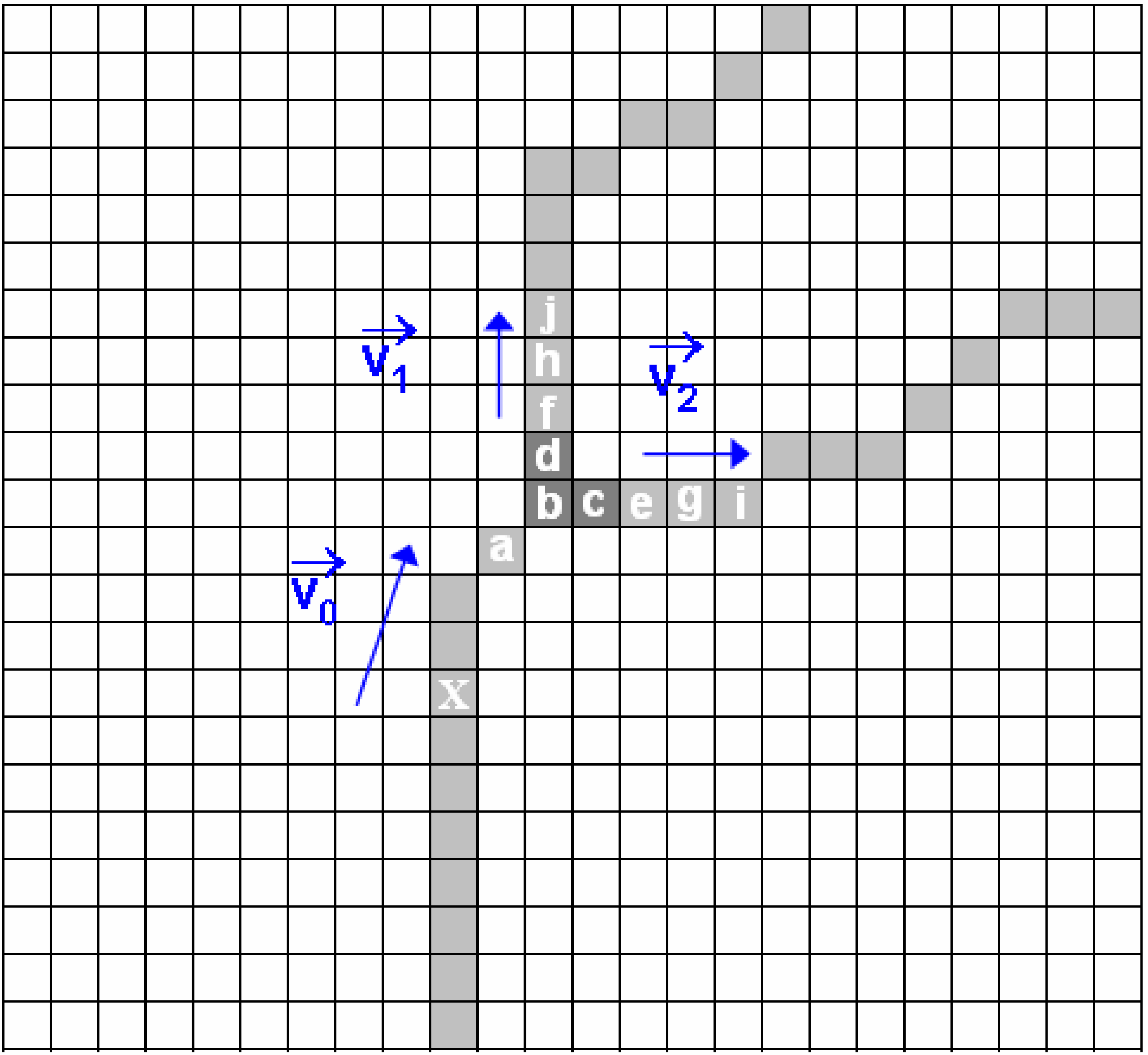}
\label{fig:crossandvec_a}}
\hfil
\subfigure[]{\includegraphics[width=60mm]{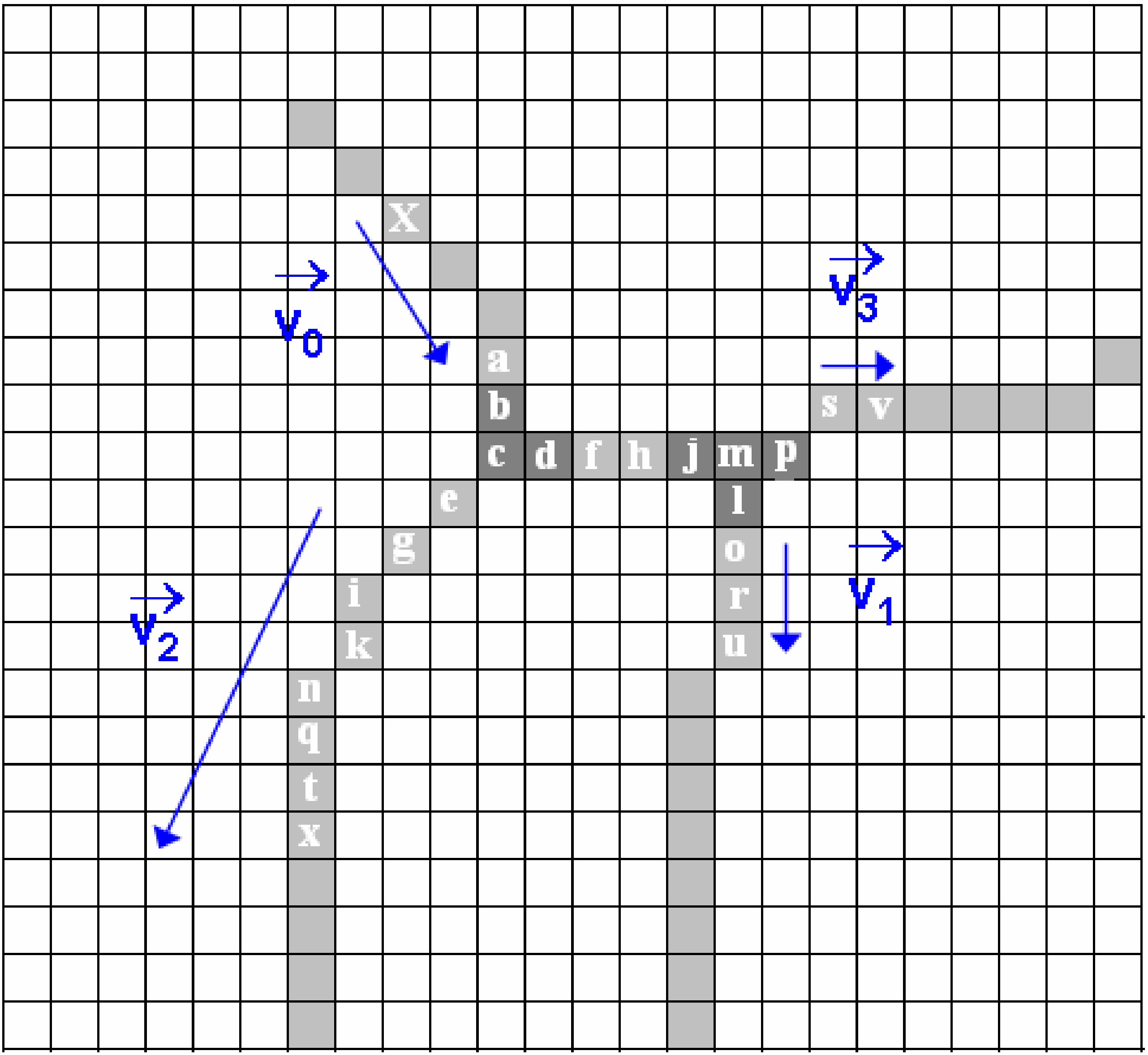}
\label{fig:crossandvec_b}}}

     \caption{\emph{Breadth First Search} application to find out extremity
     points for outwards direction vectors calculation. The \textbf{X}-marked
     pixel address is provided by the \emph{backpointer} as the origin of the inwards
     direction vector $\vec{v}_0$. Pixels appear in alphabetical order portraying the Breadth
     First Search visitation order, according to the chain-code scanning. (a) A bifurcation $1$ and its direction
     vectors. The Breadth-First Search starts up when the pixel \emph{b} is
     detected and ceases when there remain solely the pixels \emph{i} and \emph{j} in the auxiliary queue.     
     (b) A bifurcation $4$ and its direction vectors. When two critical regions of type bifurcation 1
     occur highly close to each other, the agglutination effect takes place
     and two bifurcations 1 are seen as a bifurcation 4. 
     The Breadth-First Search starts up when the pixel \emph{b} is
     detected and ceases when there remain solely the pixels \emph{u, v} and \emph{x} in the auxiliary queue.
     }
      \label{fig:crossingsandvectors}
\end{figure}

\begin{figure}[htb]
  \centering
\includegraphics[width=30mm]{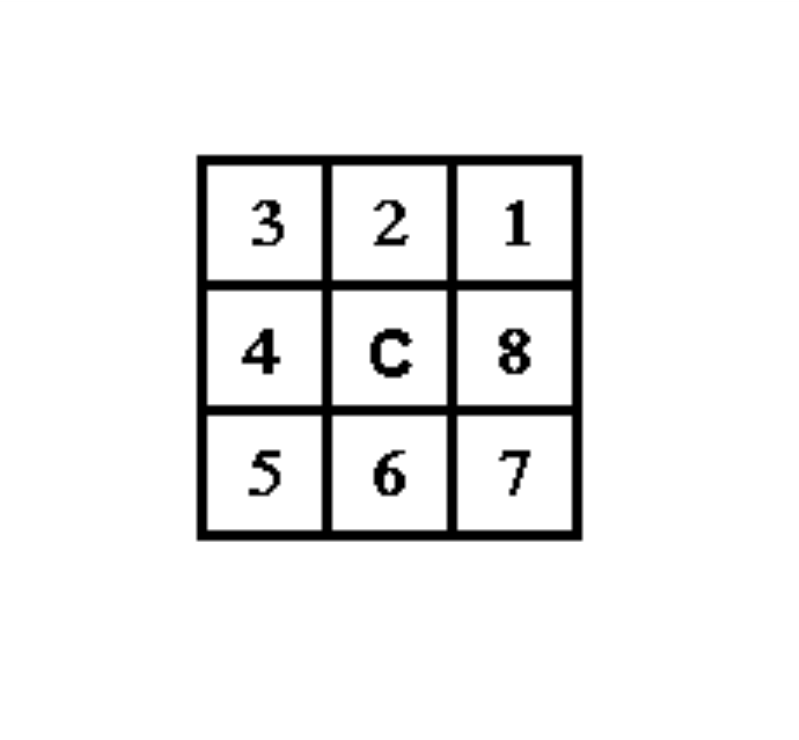}
    \caption{Neighborhood defined by the chain-code and used by the \emph{BTA} and \emph{BSCEA}}
  \label{fig:chainrulefreeman}
\end{figure}

This process is illustrated in Fig. \ref{fig:crossandvec_a} which is related to 
Table \ref{tab:table1}, Fig. \ref{fig:crossandvec_b} and Table \ref{tab:table2}. The former example illustrates the \emph{Breadth-First Search}
across a single bifurcation, while the latter example illustrates the Breadth-First Search
across two very close bifurcations, giving rise to the agglutination effect of
two bifurcations of type $1$ (Fig.~\ref{fig:crc_a}) into one bifurcation of type $4$ (Fig.~\ref{fig:crc_d}).
In both examples one may realize that the state $0$ has
been obtained by probing the vicinity of the pixel \textbf{a},
in the chain-code~(Fig.~\ref{fig:chainrulefreeman}) sequence $3, 4, 5, 6, 7, 8$. 
Having set the stop condition parameter $C$ to $5$ for these
examples, the Breadth-First Search continues until the auxiliary
queue achieves the state $07$ in the first case, and state $19$ in the second case, 
when precisely $\Sigma$ equals $C$, since the auxiliary queue has been continuously run through for $5$
times. The stop condition parameter $C$ has been empirically defined by taking into 
account a sufficient distance far away from the critical region 
to compute the outwards direction vectors.

\begin{table}
\renewcommand{\arraystretch}{1.3}
\caption{Breadth First Search across a bifurcation of type $1$: auxiliary queue states. \emph{B} is set to $1$ whenever every pixel
  in a specific state is non-critical and $0$ otherwise. $\Sigma$
  increases by one if the respective \emph{B} variable has been set to $1$, and zeroed
  otherwise. Already dequeued pixels have been concealed.}

\label{tab:table1}
\centering
\begin{tabular}[htb]{c|r|l|c|c}
\hline
state & current & auxiliary queue & B & $\Sigma$\\
\hline
00 & a & b & 0 & 0 \\
01 & b & c d & 0 & 0\\
02 & c & d e & 0 & 0\\
03 & d & e f & 1 & 1\\
04 & e & f g & 1 & 2 \\
05 & f & g h & 1 & 3 \\
06 & g & h i & 1 & 4 \\
07 & h & i j & 1 & 5 \\
\hline
\end{tabular}
\end{table}

Thus, the leftover pixels in the auxiliary queue, say
\textbf{i} and \textbf{j} in the first case, and \textbf{u}, \textbf{v} and \textbf{x} in the second case, are precisely the end points of
the required outwards direction vectors $\vec{v}_i$, obviously non-normalized yet.  
Besides the end points, the
origin points are also needed to compute each outward segment direction
vector.  In order to find the origin points, every candidate should
 satisfy two requisites at the same time: $(i)$ being the
closest neighbor to a critical region pixel and $(ii)$ having a path of valid pixels
between it and the corresponding end point.  For skeletons with low
densities of critical regions, one may relax these requirements
to only the first condition. However, in some cases the latter
requisite is mandatory so as to correctly find out each origin
point direction vector, since it may happen that there is a closest 
neighbor to a critical region which does not pertain to the current branch.
In the cases illustrated in Fig. \ref{fig:crossandvec_a} and Fig. \ref{fig:crossandvec_b}, it is 
sufficient to take into account only condition $(i)$.
  Afterwards, each outwards direction vector is normalized and inner products between the unitary
inwards direction vector $\hat{v}_0$ and each unitary outwards direction
vector are computed.  As a consequence, the unitary outwards direction
vector for which the inner product attains its maximum result, in
accordance with Eq. \ref{eq:inner-prod}, gives the optimum direction 
to continue the tracking with the very same label value, hence
providing the following pixel to be stacked. The remaining vectors
origins are enqueued as secondary seeds. 
Both examples show non-normalized outward direction vectors
, being $\vec{v}_1$ the optimum 
direction choice to continue the tracking beyond the critical region, whereas 
the remaining vectors $\vec{v}_i$ are shown as side branches directions to be considered later on.

\begin{table}
\renewcommand{\arraystretch}{1.3}
\caption{Breadth First Search across a bifurcation of type $4$: auxiliary queue states. \emph{B} is set to $1$ whenever every pixel
  in a specific state is non-critical and $0$ otherwise. $\Sigma$
  increases by one if the respective \emph{B} variable has been set to $1$, and zeroed
  otherwise. Already dequeued pixels have been concealed.}
  
\label{tab:table2}
\centering
\begin{tabular}[htb]{c|r|l|c|c}
\hline
state & current & auxiliary queue & B & $\Sigma$\\
\hline
00 & a & b & 0 & 0 \\
01 & b & c d  & 0 & 0\\
02 & c & d e   & 0 & 0\\
03 & d & e  f  & 1 & 1\\
04 & e & f g  & 1 & 2 \\
05 & f & g  h & 1 & 3 \\
06 & g & h i  & 1 & 4 \\
07 & h & i j  & 0 & 0 \\
08 & i & j k  & 0 & 0 \\
09 & j & k l m & 0 & 0 \\
10 & k & l m n  & 0 & 0 \\
11 & l & m n o p  & 0 & 0 \\
12 & m & n o p  & 0 & 0 \\
13 & n & o p q  & 0 & 0 \\
14 & o & p q r  & 0 & 0 \\
15 & p & q r s  & 1 & 1 \\
16 & q & r s  t & 1 & 2 \\
17 & r & s t u  & 1 & 3 \\
18 & s & t u  v & 1 & 4\\
19 & t & u v  x & 1 & 5 \\
\hline
\end{tabular}
\end{table}

\begin{figure}[htb]
\centering
\centerline{\subfigure[]{\includegraphics[width=60mm]{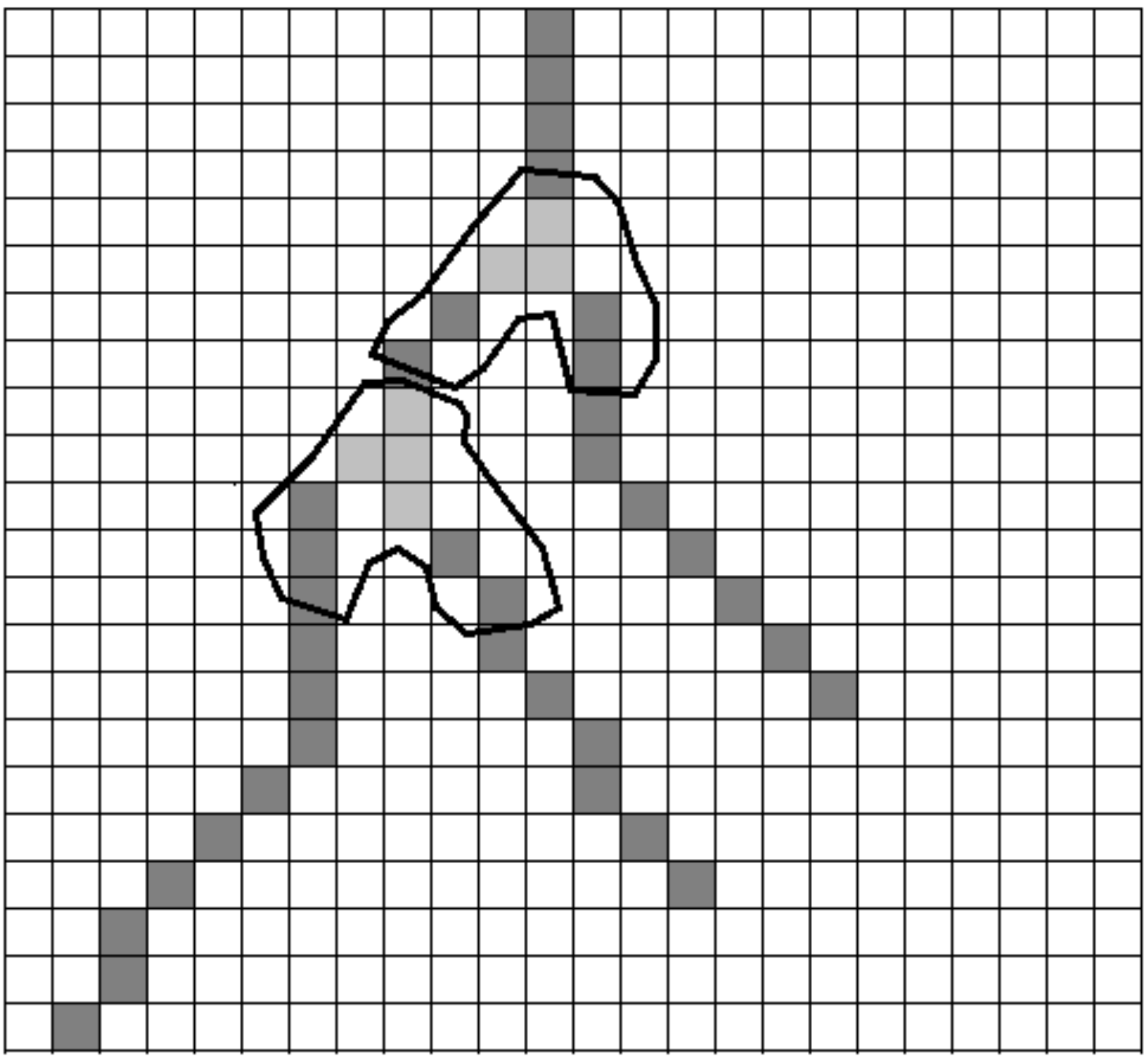}
\label{clump_a}}
\hfil
\subfigure[]{\includegraphics[width=60mm]{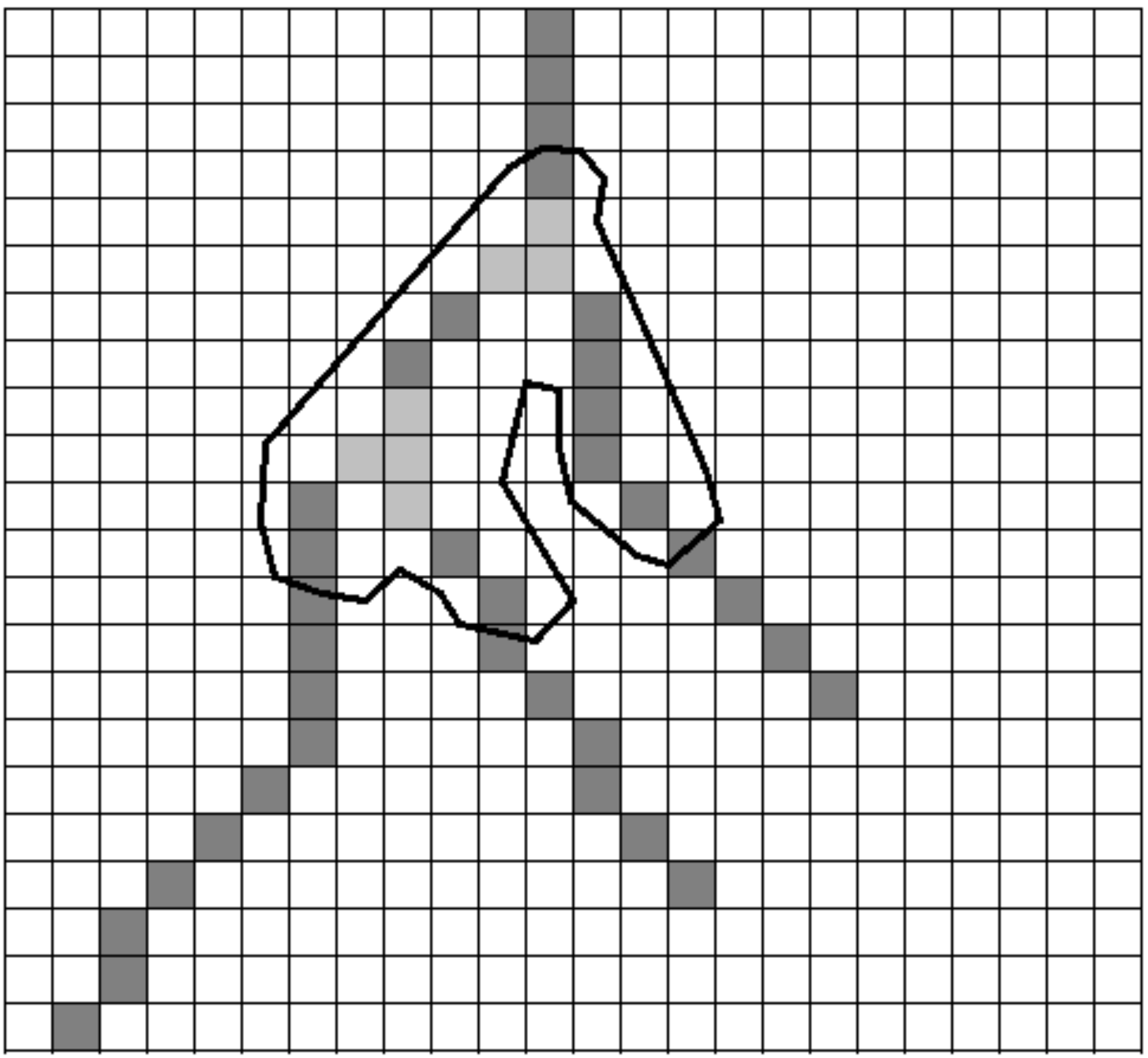}
\label{clump_b}}}
     \caption{(a) Two distinct bifurcations of type $1$ will be seen as (b) one bifurcation of type $4$, 
     an immediate consequence from the \emph{agglutinating effect} caused by the Breadth First Search 
     algorithm, when encountering two close bifurcations, as though the current local analysis had given place to a 
     more global analysis by switching into a larger analyzing scale}
      \label{fig:clumping}
\end{figure}

\newpage
\subsection{Critical Regions Classification Rules}
\label{sec-bta-cr-class-rules}

Let $S$ be a set of critical regions $S=\{s_1, s_2, s_3...\}$, $s_1$
being the current critical region and $D_{max}$ the shortest allowed
path length between two consecutive critical regions. Let $\hat{v}_0$
be the unitary \textbf{inwards} vector to $s_1$ and let $E_1$ and $E_2$
be the sets of unitary \textbf{outwards} vectors from close critical
regions $s_1$ and $s_2$ respectively.  Let $|E_i|$ be the cardinality
of $E_i$. Then, a critical region should be classified into one of the
eight classes depicted in the
Fig. \ref{fig:crclassificationdecisiontree}, according to the
following rules:

\begin{enumerate}
\item Bifurcation 1: Fig. \ref{fig:crc_a}\\
$|E_1|=2$ and\\
$< \hat{v}_0 . \hat{v}_i > > 0, \quad \forall \hat{v}_i \in E_1$

\item Bifurcation 2: Fig. \ref{fig:crc_b}\\
$|E_1|=2$  and\\
$\exists \hat{v}_i, \hat{v}_j \in E_1  \quad | $\\
$< \hat{v}_0 . \hat{v}_i > > 0$ and $ < \hat{v}_0 . \hat{v}_j > \leq 0$ and\\
$\nexists s_2 \quad | \quad dist(s_1, s_2) < D_{max}$

\item Bifurcation 3: Fig. \ref{fig:crc_c} \\
$|E_1|=2$, but\\
$\exists s_2  |  dist(s_1, s_2) < D_{max} \therefore |E_1|+|E_2|=3$\\
$\exists \hat{v}_j \in s_1$ e $\hat{v}_k, \hat{v}_l \in s_2$, such that \\
$< \hat{v}_0 . \hat{v}_k > \approx 1 \quad$ and $\quad < \hat{v}_j . \hat{v}_l > \neq -1$ 

\item Bifurcation 4: Fig. \ref{fig:crc_d} \\
$|E_1|=2$, but\\
$\exists s_2  |  dist(s_1, s_2) < D_{max} \therefore |E_1|+|E_2|=3$\\
$\exists \hat{v}_j \in s_1$ e $\hat{v}_k, \hat{v}_l \in s_2$, such that \\
$< \hat{v}_0 . \hat{v}_k > \approx 1 \quad$, $\quad < \hat{v}_j . \hat{v}_l > \neq -1$ \\
and $\quad < \hat{v}_0 . \hat{v}_j > \leq 0$ 

\item Superposition: Fig. \ref{fig:crc_e}\\
$|E_1|=2$, but\\
$\exists s_2  |  dist(s_1, s_2) < D_{max} \therefore |E_1|+|E_2|=3$\\
$\exists \hat{v}_j \in s_1$ and $\hat{v}_k, \hat{v}_l \in s_2$, such that\\
$< \hat{v}_0 . \hat{v}_j >  \leq 0,$ and\\
$< \hat{v}_0 . \hat{v}_k > \approx 1 \quad and \quad < \hat{v}_j . \hat{v}_l > \approx -1$ 

\item Crossing: Fig. \ref{fig:crc_f}\\
$|E_1|=3$ and\\
$\exists \hat{v}_j, \hat{v}_k, \hat{v}_l \in s_1$ such that\\
$<\hat{v}_0 . \hat{v}_k > \approx 1 \quad$ and $\quad < \hat{v}_j . \hat{v}_l > \approx -1$ 
\end{enumerate}

\section*{Acknowledgement} 

The authors are grateful to Fapesp, CNPq, Capes and Finep for
financial support, as well as to the \emph{Nature Publishing
Group} for the license to use the images in\cite{masland01}.

\bibliographystyle{elsart-num-sort.bst}
\bibliography{bigbib}

\begin{thebibliography}{10}
\expandafter\ifx\csname url\endcsname\relax
  \def\url#1{\texttt{#1}}\fi
\expandafter\ifx\csname urlprefix\endcsname\relax\def\urlprefix{URL }\fi

\bibitem{arulampalam02tutorial}
S.~Arulampalam, S.~Maskell, N.~Gordon, T.~Clapp, A tutorial on particle filters
  for on-line non-linear/non-gaussian bayesian tracking, IEEE Transactions on
  Signal Processing 50~(2) (2002) 174--188.

\bibitem{bresenham1965}
J.~Bresenham, Algorithm for computer control of a digital plotter, IBM Systems
  Journal 4~(1) (1965) 25--30.

\bibitem{caserta90}
F.~Caserta, H.~Stanley, W.~Eldred, G.~Daccord, R.~Haussman, J.~Nittmann,
  Physical mechanisms underlying neurite outgrowth: A quantitative analysis of
  neuronal shape, Physical Review Letters 64~(1) (1990) 95--98.

\bibitem{castleman79}
K.~R. Castleman, Digital Image Processing, Prentice-Hall, Englewood Cliffs, NJ,
  1979.

\bibitem{cesarcost:1998}
R.~M. Cesar-Jr., L.~da~F.~Costa, Neural cell classification using wavelets and
  multiscale curvature, Biological Cybernetics 79~(4) (1998) 347--360.

\bibitem{cesar99}
R.~M. Cesar-Jr., L.~da~F.~Costa, Dendrogram generation for neural shape
  analysis, The Journal of Neuroscience Methods 93 (1999) 121--131.

\bibitem{surv_condmat}
L.~da~F.~Costa, Morphological complex networks: Can individual morphology
  determine the general connectivity and dynamics of networks?, in: COSIN 2005
  Final meeting, Conference on Complex Networks, Salou, Spain, 2005,
  \url{http://xxx.lanl.gov/abs/q-bio.MN/0503041}.

\bibitem{costabook01}
L.~da~F.~Costa, R.~M. Cesar-Jr., Shape Analysis and Classification: Theory and
  Practice, CRC Press, 2001.

\bibitem{costa2002}
L.~da~F.~Costa, E.~Manoel, F.~Faucereau, J.~Chelly, J.~van Pelt, G.~Ramakers, A
  shape analysis framework for neuromorphometry, Network: Computation in Neural
  Systems 13 (August 2002) 283--310(28),
  \url{http://www.ingentaconnect.com/content/apl/network/2002/00000013/0000000%
3/art00303}.

\bibitem{morphlotufo}
E.~R. Dougherty, R.~A. Lotufo, Hands-On Morphological Image Processing,
  SPIE-International Society for Optical Engine, 2003.

\bibitem{gabriel03}
P.~F. Gabriel, J.~G. Verly, J.~H. Piater, A.~Genon, The state of the art in
  multiple object tracking under occlusion in video sequences, in: In Advanced
  Concepts for Intelligent Vision Systems (ACIVS), 2003, 2003.

\bibitem{herman1998}
G.~T. Herman, Geometry of Digital Spaces, Birkhauser Boston, 1998.

\bibitem{mmorph:2006}
S.~{I}nformation {S}ystems, {SDC} {M}orphology {T}oolbox for {M}atlab,
  \url{http://www.mmorph.com/}, version 1.5 (October 2006).

\bibitem{jelinekfernandez98}
H.~Jelinek, E.~Fernandez, Neurons and fractals: How reliable and useful are
  calculations of fractal dimensions?, Journal of Neuroscience Methods 81~(1-2)
  (1998) 9--18.

\bibitem{leandrobranching}
J.~J.~G. Leandro, R.~M. Cesar-Jr., L.~da~F.~Costa, Determining the branching
  structure of 3{D} structures from respective 2{D} projections, in: Proc. 19th
  SIBGRAPI - Brazilian Symposium on Computer Graphics and Image Processing,
  IEEE Computer Society Press, 2006.

\bibitem{leandro:2008}
J.~J.~G. Leandro, R.~M. {Cesar Jr}, L.~da~F.~Costa, Automatic contour
  extraction from 2d neuron images, \url{http://aps.arxiv.org/abs/0804.3234}
  (2008).

\bibitem{leandrobioimage:2008}
J.~J.~G. Leandro, R.~M. Cesar-Jr., L.~da~F.~Costa, Automatic contour extraction
  of neurons in presence of overlap, in: Workshop on Bio-Image Informatics:
  Biological Imaging, Computer Vision and Data Mining, 2008, Santa Barbara, CA,
  USA, 2008, \url{http://www.ece.ucsb.edu/bioimage/workshop2008/index.html}.

\bibitem{masland01}
R.~H. Masland, The fundamental plan of the retina, Nature Neuroscience 4~(9)
  (2001) 877--886.

\bibitem{morigiwa89}
K.~Morigiwa, M.~Tauci, Y.~Fukuda, Fractal analysis of ganglion cell dendritic
  branching patterns of the rat and cat retinae, Neuroscience Research Suppl.
  10 (1989) S131--S140.

\bibitem{rocchi2007}
M.~B.~L. Rocchi, D.~Sisti, M.~C. Albertini, L.~Teodori, Current trends in shape
  and texture analysis in neurology: Aspects of the morphological substrate of
  volume and wiring transmission, Brain Research Reviews 55 (2007) 97--107.

\bibitem{Rothaus:2007}
K.~Rothaus, P.~Rhiem, X.~Jiang, Separation of the retinal vascular graph in
  arteries and veins, in: F.~Escolano, M.~Vento (eds.), GbRPR 2007, Graph-Based
  Representations in Pattern Recognition, 6th IAPR-TC-15 International
  Workshop, Alicante, Spain, Proceedings, vol. 4538 of Lecture Notes in
  Computer Science, Springer Verlag, 2007,
  \url{http://www.springerlink.com/content/d573048432h4k13x/}.

\bibitem{Rothaus:2008}
K.~Rothaus, P.~Rhiem, X.~Jiang, Separation of the retinal vascular graph based
  upon structural knowledge, doi:10.1016/j.imavis.2008.02.013, in press, Image
  and Vision Computing (2008).

\bibitem{sedgewick:1983}
R.~Sedgewick, Algorithms, Addison-Wesley Publishing Company, Reading, MA, USA,
  1983.

\bibitem{soille99}
P.~Soille, Morphological Image Analysis: Principles and Applications, Springer
  Verlag, 1999.

\end{thebibliography}
\end{document}